\documentclass[letterpaper, twoside]{article}
\usepackage{graphicx}
\usepackage{subcaption}
\usepackage[accepted]{aistats2026}

\usepackage{placeins}
\usepackage{amsmath}
\usepackage{amssymb}
\usepackage{enumitem}
\usepackage{multicol}
\usepackage{xcolor}
\usepackage{multirow}
\usepackage{adjustbox}
\usepackage{booktabs}

\usepackage{hyperref}
\hypersetup{
    colorlinks,
    linkcolor={blue!90!black},
    citecolor={magenta},
    urlcolor={blue!80!black}
}

\newcommand{\fplmm}{f^{(\text{PL})}(t)}
\newcommand{\fpl}{$\fplmm$}
\newcommand{\fsplmm}{f^{(\text{SPL})}(t)}
\newcommand{\fspl}{$\fsplmm$}
\newcommand{\fbwomm}{f^{(\text{BW})}_1(t)}
\newcommand{\fbwo}{$\fbwomm$}
\newcommand{\fbwtmm}{f^{(\text{BW})}_2(t)}
\newcommand{\fbwt}{$\fbwtmm$}
\newcommand{\fbwnmm}{f^{(\text{BW})}_n(t)}
\newcommand{\fbwn}{$\fbwnmm$}
\newcommand{\femm}{f^{(\text{E})}(t)}
\newcommand{\fe}{$\femm$}
\newcommand{\maskLmm}{\mathbf{M}^{(\text{R})}}
\newcommand{\maskL}{$\maskLmm$}
\newcommand{\maskCmm}{\mathbf{M}^{(\text{C})}}
\newcommand{\maskC}{$\maskCmm$}
\newcommand{\maskCLmm}{\mathbf{M}^{(\text{C,R})}}
\newcommand{\maskCL}{$\maskCLmm$}

\newcommand{\weightCLmm}{\mathbf{C}^{(\text{C,R})}}
\newcommand{\weightCL}{$\weightCLmm$}

\usepackage[margin=0.88in, footskip=3.30003pt]{geometry}
\usepackage[round]{natbib}

\begin{document}

%

\twocolumn[

\aistatstitle{Recency Biased Causal Attention for Time-series Forecasting}

\aistatsauthor{ Kareem Hegazy \And Michael W. Mahoney \And  N. Benjamin Erichson}

\aistatsaddress{ UC Berkeley and ICSI \And  UC Berkeley and ICSI and LBNL \And ICSI and LBNL } ]

\begin{abstract}
    Recency bias is a useful inductive prior for sequential modeling: it emphasizes nearby observations and can still allow longer-range dependencies. Standard Transformer attention lacks this property, relying on all-to-all interactions that overlook the causal and often local structure of temporal data. We propose a simple mechanism to introduce recency bias by reweighting attention scores with a smooth heavy-tailed decay. This adjustment strengthens local temporal dependencies without sacrificing the flexibility to capture broader and data-specific correlations. We show that recency-biased attention consistently improves sequential modeling, aligning Transformer more closely with the read–ignore–write operations of RNNs. Finally, we demonstrate that our approach achieves competitive and often superior performance on challenging time-series forecasting benchmarks.
\end{abstract}

\section{INTRODUCTION}

Recency bias is a useful inductive prior for sequential modeling: it emphasizes nearby information, while still allowing longer-range dependencies. Recurrent neural networks (RNNs) naturally embody this bias through their local and causal structure, which enables them to excel at tasks that require selective reading, ignoring, or writing of information. For example, in the flip-flop task~\citep{liu2023exposing}, RNNs easily learn to store and discard information as needed, whereas Transformers often fail despite their greater expressivity. This limitation highlights a fundamental gap: while RNNs leverage temporal locality, standard Transformer attention does not.

Transformers have nevertheless revolutionized sequential modeling, achieving state-of-the-art results across language and time-series benchmarks. At the core of their success lies the multihead attention (MHA) mechanism, which integrates information from all positions through pairwise similarity. This all-to-all formulation is powerful for natural language processing (NLP), where dependencies between words can span long distances and do not depend strongly on order. Yet the same property can be a liability for time-series forecasting and structured sequential tasks, where local and causal structure is essential, and less relevant distant interactions can obscure the signal.

Recent work has shown that introducing recency biases can improve Transformer performance on weakly local and even noncausal language tasks~\citep{press2022train,raffel2023exploringlimitstransferlearning,liu2025comprehensivesurveylongcontext}. These results suggest that models benefit from temporal locality even when the underlying data does not demand it. Surprisingly, however, the role of recency bias in time-series forecasting remains underexplored, despite the fact that time-series data is causal and highly local.

In time-series forecasting, each embedding corresponds to a causal temporal measurement, and the strength of dependencies typically decays with temporal separation. Time-series forecasting problems are inherently causal; the input and target sequences do not overlap.
However, attention methods in most state-of-the-art models~\citep{nie.patchtst.2023a, zhou.fedformer.2022, zhou2023OFA, Zhou.informer.2021, wu2021autoformer} do \textbf{not} apply a causal mask.
Thus, the temporal embeddings are non-causally updated with other input sequence embeddings from later times. This mismatch between the all-to-all nature of Transformer attention and the local-causal structure of time-series data raises a central question: how well suited is standard attention for time-series forecasting?

We develop the Recency Biased Causal Attention (RBCA) framework to generally study the effects of causal and recency biases on Transformers as they are applied to local and causal datasets.
We investigate the flexibility of recency biases to simultaneously perturb the attention correlation structure while sufficiently capturing pairwise correlations unique to each dataset.
To do so, we leverage insights from time-series forecasting problems to broadly improve upon Transformer's fundamental capabilities.
Specifically, we note that attention weights are a normalized similarity metric similar to correlations.
We are further motivated by the observation that many physical systems exhibit heavy-tailed autocorrelations, e.g., the pairwise correlation strength may decay as a power-law as the time delay grows \citep{clauset.powerlaw.2009}.

We develop power-law recency biases to impose a natural correlation structure on the attention weight distribution and systematically compare their effects across tasks. Our study evaluates recency biases in a principled way, highlighting both their strengths and their nuanced differences across causal and locally structured domains in NLP and time-series forecasting. We begin by testing whether recency biases equip Transformers with the basic read, ignore, and write capabilities characteristic of RNNs. Next, we apply RBCA to a standard encoder-decoder Transformer and assess its performance on causal and locally correlated time-series datasets. Finally, we investigate the effects of RBCA on modern time-series forecasting architectures by developing \emph{Powerformer}, a PatchTST~\citep{nie.patchtst.2023a} variant with RBCA.

Our main contributions are the following.

\begin{itemize}[leftmargin=*] 

\item We propose Recency-Biased Causal Attention (RBCA), a general framework for introducing recency bias into Transformers. With RBCA, we develop power-law-inspired recency biases that induce a heavy-tailed attention distribution while allowing attention to learn the unique data-dependent correlation structure.

\item We evaluate recency biases on the flip-flop task to test core sequential capabilities such as read, write, and ignore. We find that NLP-derived recency biases degrade performance, while our power-law biases enable Transformers to achieve better accuracy.

\item We assess RBCA in time-series forecasting. First, we apply it to a standard Transformer; and then we introduce \emph{Powerformer}, a simple encoder-only variant that amplifies recency effects. Our power-law biases consistently improve forecasting accuracy across benchmarks. Notably, \emph{Powerformer} further strengthens the recency bias despite its regularizing effect, thus highlighting the fundamental importance of recency biases for time-series prediction.
\end{itemize}


\section{RELATED WORK}

The success of Transformers~\citep {vaswani.transformer.2017} and LLMs inspired Transformer-based time-series models.
For example, Time-GPT \citep{garza2024timegpt1} and One-Fits-All \citep{zhou2023OFA}, which fine-tunes a pretrained GPT-2~\citep{radford2019language} with temporal embedding.
Many other models aimed to alter the attention mechanism or the input embeddings.

Many early time-series Transformer models altered the attention mechanism to improve alignment with time-series properties.
Fourier representations separate fast and slow varying dynamics, simplifying attention representations \citep{zhou.fedformer.2022, woo2022etsformerexponentialsmoothingtransformers}.
AutoFormer \citep{wu2021autoformer} replaces the similarity metric with autocorrelations.
FlowFormer \citep{wu.flowformer.2022} achieves linear complexity attention without a recency bias.
Informer \citep{Zhou.informer.2021} selects high attention weight values, increasing prediction capacity.
iTransformer \citep{liu2024itransformer} inverts the temporal and embedding dimensions during MHA.

Some works alter the input embedding to enrich the input or better align it with NLP tasks.
PatchTST patches the input sequence, outperforming more complicated methods \citep{nie.patchtst.2023a}.
Chronos \citep{ansari2024chronoslearninglanguagetime} tokenizes the continuous input and trains on large datasets.
TOTEM \citep{talukder2024totem} also discretizes the input time-series, but uses VQVAE \citep{vandenoord.vqvae.2017}.

To impose a locality bias, recency biases have been used in both NLP~\citep{liu2025comprehensivesurveylongcontext} and time-series tasks.
For NLP, ALiBi~\citep{press2022train} employs smooth exponential decay, while block~\citep{lu2025mobamixtureblockattention} and sliding-window~\citep{beltagy2020longformerlongdocumenttransformer} attentions employ a cutoff scheme.
Previous recency-biased time-series models include 
Reformer \citep{Kitaev2020Reformer}, which uses a locality-sensitive hashing mechanism, and ETSformer~\citep{woo2022etsformerexponentialsmoothingtransformers}, which employs a non-learnable exponential decay in one of their attention mechanisms.
Earlier time-series CNN-based works also employed temporal dilations to bias towards early data~\citep{li.enhancing_locality.2019, vandenoord.wavenet.2016, franceschi.unsupervised.2019}.

\begin{figure*}[!th]
    \centering
    \includegraphics[width=0.9\textwidth]{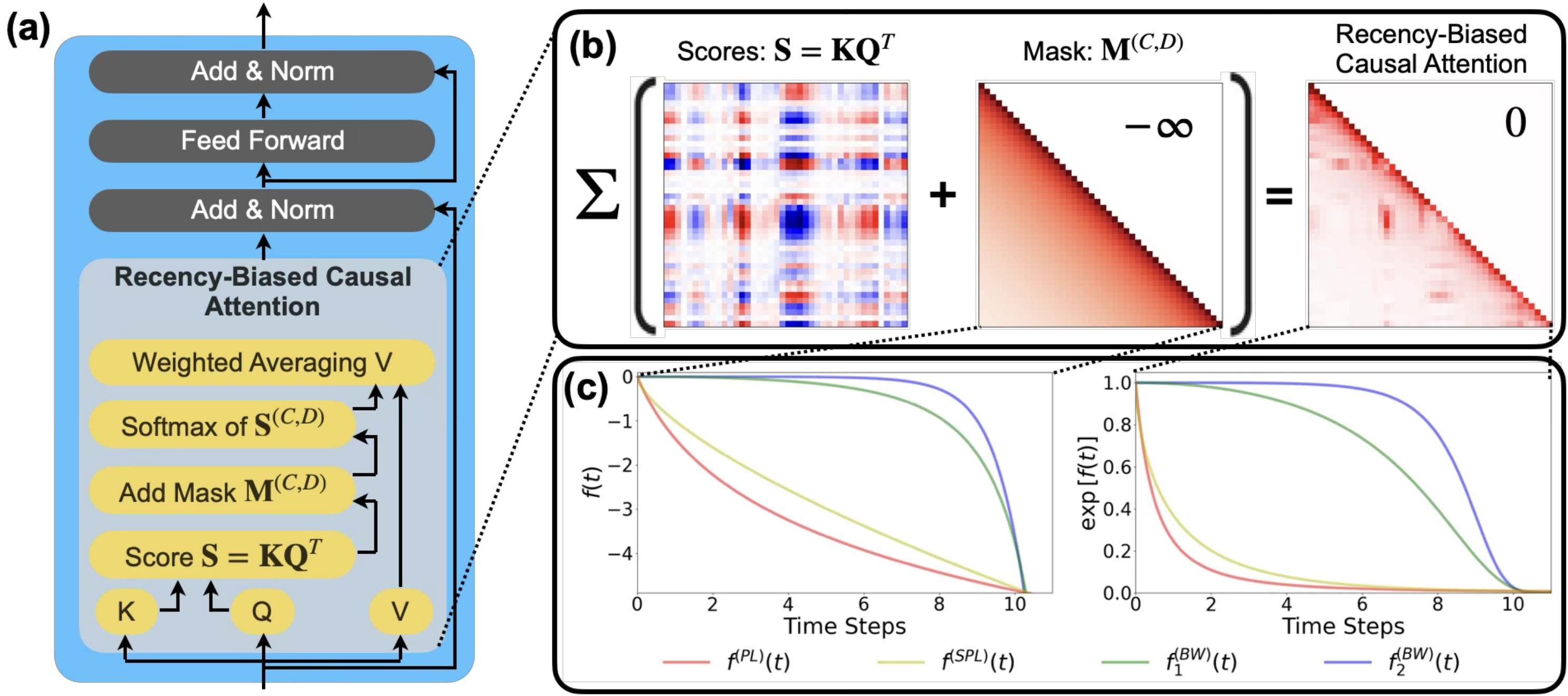}
    \caption{Illustration of Recency Biased Causal Attention (RBCA) and how it fits into the Transformer layer and adapts the attention. Panel (a) shows how RBCA replaces MHA in the standard Transformer block. Panel (b) shows a local-causal mask and subsequent weights, where $\Sigma$ corresponds to the softmax function, red and blue indicate positive and negative values, respectively, and white is labeled when ambiguous. Panel (c) demonstrates the 4 different masking functions we apply to the score (left) and their effects on the weights (right).}
    \label{fig:architecture}
\end{figure*}

\section{METHOD}

\begin{figure*}[!t]
\centering
\includegraphics[width=0.9\linewidth]{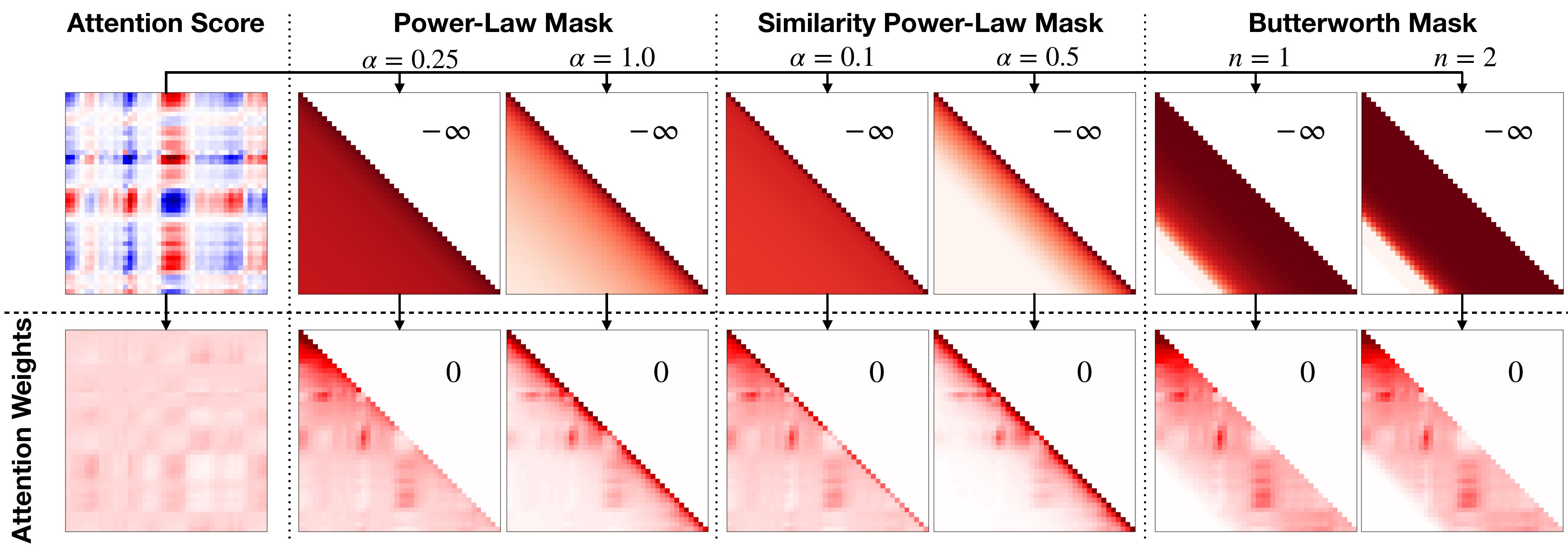}
\caption{A comparison between standard non-causal and nonlocal attention (bottom left) and examples of RBCA with varying masks (top row) applied to the same attention score (top left). In the top row, red is maximized at 0 and minimized (white) at $-\infty$. In the bottom row, red is maximized at 1 and minimized (white) at 0. For the attention score red and blue indicate positive and negative.}
\label{fig:maskPL}
\end{figure*}

We propose Recency-Biased Causal Attention (RBCA) as a framework for recency biases.
In this section, we review standard Multihead Attention (MHA) and demonstrate how we add causal and recency biases to it.
Figures~\ref{fig:architecture} and \ref{fig:maskPL} illustrate RBCA and its effect on attention weights.

\subsection{Standard Multihead Attention}

Transformers typically leverage MHA to compute new embeddings from weighted sums of input embeddings. Let $\mathbf{X} \in \mathbb{R}^{T \times D}$ be a sequence of $T$ input vectors, each of dimension $D$. For all $H$ attention heads, indexed by $h$, MHA first projects $\mathbf{X}$ into \emph{query}, \emph{key}, and \emph{value} matrices, respectively denoted by $\mathbf{Q}_h$, $\mathbf{K}_h$, and $\mathbf{V}_h$: $\mathbf{Q}_h = \mathbf{X}\,\mathbf{W}_h^{(\text{Q})}$, $\mathbf{K}_h = \mathbf{X}\,\mathbf{W}_h^{(\text{K})}$, and $\mathbf{V}_h = \mathbf{X}\,\mathbf{W}_h^{(\text{V})}$.
Each projection matrix $\mathbf{W}_h^{(\cdot)} \in \mathbb{R}^{D \times D_k}$ is learnable, with head size $D_k$. The parameters $\mathbf{Q}_h$, $\mathbf{K}_h$, and $\mathbf{V}_h$ thus each have the shape $T \times D_k$.

The \emph{unnormalized attention similarity scores} measure how each query interacts with all keys
\begin{equation}
	\label{eq:ShCh}
	\mathbf{S}_h = \frac{\mathbf{K}_h \,\mathbf{Q}_h^\top}{\sqrt{D_k}},
\end{equation}
where the factor of $1/\sqrt{D_k}$ helps stabilize the gradients by normalizing the dot products according to the head dimension.
In turn, the \emph{normalized attention weights} $\mathbf{C}_h = \text{Softmax}\bigl(\mathbf{S}_h\bigr)$ has each row summing to 1. Finally, each head's output $\tilde{\mathbf{X}}_h = \mathbf{C}_h \,\mathbf{V}_h$ is computed as a weighted sum of the values, concatenated with the other heads, and transformed by a linear projection
\begin{equation}
	\label{eq:MHAOutput}
 \tilde{\mathbf{X}} 
	\;=\;
	[\tilde{\mathbf{X}}_1,\dots,\tilde{\mathbf{X}}_H]
	\,\mathbf{W}^{(\text{A})} ,
\end{equation}
where $\mathbf{W}^{(\text{A})} \in \mathbb{R}^{(H\,D_k)\times D}$ is a learnable matrix. The result $\tilde{\mathbf{X}}$ has the same dimensionality as the input $\mathbf{X}$ (i.e., $T \times D$), allowing it to be passed into subsequent Transformer layers or decoded.

\paragraph{Enforcing Causality}
%
The standard causal mask prohibits future embeddings from updating past ones:
\begin{align}
&\mathbf{M}^{(\text{C})}_{t,t'} 
	= \begin{cases}
		-\infty & t' > t, \\
		0 & t' \leq t,
	\end{cases}
	\label{eq:mask_causal} \\
\mathbf{S}_h^{(\text{C})}
= &\mathbf{S}_h + \mathbf{M}^{(\text{C})},
\quad
\mathbf{C}_h^{(\text{C})}
= \text{Softmax}\bigl(\mathbf{S}_h^{(\text{C})}\bigr). \nonumber
\end{align}
The attention weights $\mathbf{C}^{(C)}_{t,t'}=0$ for all $t'>t$, ensuring a causal structure for the time-series data.

\subsection{Recency-Biased Causal Attention}
\label{sec:wcma}
The recency bias is a mask $\mathbf{M}^{(\text{L})}$ defined by some nonpositive function $f(\Delta t)\leq0$ that is dependent on the relative time difference ($\Delta t$) between embeddings:
\begin{equation}
	\mathbf{M}^{(\text{L})}_{i,j} 
	= \begin{cases}
		0 & j > i, \\
		f(t_i - t_j) & j \leq i .
	\end{cases}
	\label{eq:mask_local}
\end{equation}
RBCA combines the recency bias and causal mask:
\begin{align}
    \mathbf{M}^{(\text{C,L})} = \mathbf{M}^{(\text{C})}{}& + \mathbf{M}^{(\text{L})}, \quad
    \mathbf{S}_h^{(\text{C,L})} = \mathbf{S}_h + \mathbf{M}^{(\text{C,L})}
    \label{eq:attnSCD} \\
    \mathbf{C}_h^{(\text{C,L})}
    &= \text{Softmax}\bigl(\mathbf{S}_h^{(\text{C,L})}\bigr)\\
    C_{h,i,j}^{(\text{C,L})} &\propto 
    \begin{cases}
            0 & j > i, \\
    	\exp\bigl[f(t_i - t_j)\bigr] & j \le i.
    \end{cases}
    \label{eq:weight_ME}
\end{align}

By combining causal masking and smoothly decaying weights, RBCA captures temporal ordering and localized dependencies inherent to real-world processes.

\subsection{Recency Bias Functions}
%
Generally, previous NLP works consider sliding windows~\citep{beltagy2020longformerlongdocumenttransformer} and exponentially decaying recency biases (ALiBi~\citep{press2022train}).
For RBCA, these are given by
\begin{equation*}
    \fbwnmm = \left(1 + \left( \frac{z(t)}{t_c}\right)^{2n} \right)^{-1/2}, \quad \femm = -\alpha t ,
\end{equation*}
respectively, where $\alpha$ is the decay time constant, and \fbwn{} is the Butterworth filter that mimics a smoothed step function (Figs.~\ref{fig:architecture}c and \ref{fig:maskPL}).
The Butterworth filter (Section~\ref{sec:BW}) depends on the cutoff time $t_c$, the decay order $n$, and the digital filter gain $z(t)$.

Here, we leverage the fact that pairwise correlations in many physical systems follow a power-law distribution~\citep{clauset.powerlaw.2009}.
We note that attention weights are analogous to pairwise correlations, as they measure a pairwise similarity.
Consequently, we make two power-law recency biases
\begin{equation*}
    \fplmm = -\alpha\,\log(t), \quad \fsplmm = -(t)^{-\alpha} .
\end{equation*}
\fpl{} induces a power-law envelope onto the attention weights, and \fspl{} adds a power-law bias to the attention score that results in an exponential of a power-law after the softmax.

\subsection{Regularization and Reduced Complexity}
The recency bias is a regularizer on the input data, as it limits MHA's ability to attend to inputs far away.
Intuitively, the recency bias acts as a hard-constraint that forces the method to attend to recent measurements.
Some recency biases, depending on the decay rate $\alpha$, sufficiently attenuate the weights to 0 at some time $\tau$.
Figures~\ref{fig:architecture}c and \ref{fig:maskPL} show such instances.
In these cases, the attention complexity can be reduced from quadratic in input length $T$ to linear: O($\tau T$).


\subsection{Interpreting Power-Law Biases}
Power-law recency biases strike a balance between sampling sinusoidal and slowly developing signals over long periods of time with sampling nearby quickly varying signals.
To develop this intuition, we consider two limits of signal types: periodic systems (which benefit from long-time windows) and systems with fast exponentially decaying pairwise correlations (which benefit from a quickly decaying recency bias).
Sinusoidal signals benefit from long and equally weighted windows, while short exponentially decaying signals benefit from swiftly decaying biases.
The power-law biases are a combination of the benefits from the exponential bias and the long-time window.
At recent times, the mask decays the fastest: analogous to the exponential decay that amplifies the most recent information.
At longer times, the heavy tail allows the model to sample many more periods of an oscillation than other biases (e.g. exponential).
The heavy tail filters for long-term signals since higher-frequency oscillations and fast variations will be averaged out, but long-term oscillations and slowly developing signals on the order of the tail's length will remain.
By attending to attenuated distant tokens, the power law enables RBCA to capture (or remember) long-term signals, providing a more expressive class of inductive biases that outperform exponential decay in practice.

\section{EXPERIMENTS}
\label{sec:experiments}

We investigate the general effects of recency biases (RBCA) on Transformers for causal and local tasks.
This includes both NLP and time-series tasks.
First, we test if the recency bias endows Transformer with capabilities similar to RNN's read, ignore, and write.
We see that some biases do give Transformer these capabilities.
Second, we build upon this insight by applying RBCA to the original encoder-decoder Transformer and testing on standard time-series tasks. Lastly, we continue investigating the benefits of RBCA by applying it to a simple state-of-the-art time-series model (PatchTST~\citep{nie.patchtst.2023a}) and find that it further improves performance and outperforms other models that claim to beat PatchTST.

\subsection{Addressing Transformer Glitching}

Prior work \citep{liu2023exposing} showed that Transformers struggle to reliably perform simple, yet important, RNN tasks together: write information to a state, ignore input information, and read from the stored state.
These tasks are important for filtering and recalling important information in long sequences.
RNNs can easily perform this task, likely due to their causal structure and implicit recency bias.
Here, we investigate whether the causal and recency biases from RBCA endow Transformer with similar capabilities.

\paragraph{Evaluation} To test these capabilities, \citep{liu2023exposing} developed the flip-flop test: a string of alternating characters $c \in \{i, w, r\}$ and numbers $n \in \{0,1\}$.
Each character instructs the model to ignore ($i$) the following number, write ($w$) the following number to memory, or read ($r$) the latest number that was written.
For the example $w0i1i0w1i0r$, the model reads $1$ since it was the last written number.
The out-of-distribution (OOD) test contains many ignores ($i$) after the last write ($w$) and tests Transformer's ability to generalize its read, ignore, and write capabilities to longer sequences.
We train and test a 2-layer encoder-only Transformer with and without RBCA.
Each attention head has a different $\alpha$.
Section~\ref{sm:flipflop_exp} provides details on the model, evaluation, and results.

\begin{figure*}[!ht]
    \centering
    \includegraphics[width=0.68\textwidth]{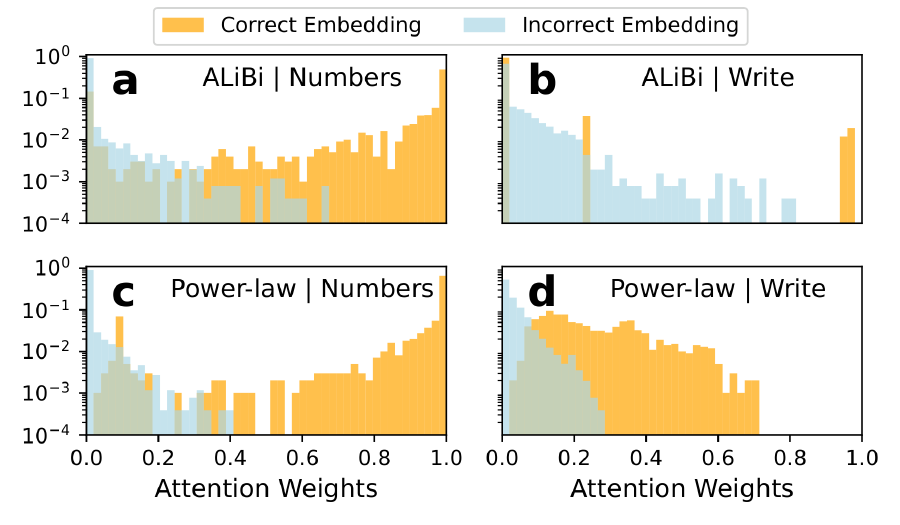}
    \caption{The attention weights corresponding to the flip flop `read' command and the correct/incorrect `write' command and number. We show these distributions for the ALiBi (\fe) and the \fpl{} biases.}
    \label{fig:flipflop_weights}
\end{figure*}

\paragraph{Attention Analysis} We first investigate the effects of RBCA on the attention weight distributions of ALiBi (one of the most prominent NLP recency biases) and our physically-inspired power-law bias: \fe{} and \fpl{} respectively.
Figure~\ref{fig:flipflop_weights} shows these attention weight distributions for the best attention heads, full results are in Section~\ref{sm:flipflop_exp}.
Most importantly, the attention weight between the `read' and correct number pair needs to be high, while the weight between the `read' and the incorrect numbers should be close to 0.
For ALiBi (\fe), the amount of weights equal to 0 and 1 for the correct number pair are similar (Fig.~\ref{fig:flipflop_weights}a).
Conversely, for \fpl, all weights corresponding to the correct number are nonzero, with almost 90\% equal to 1 (Fig.~\ref{fig:flipflop_weights}c).
When identifying the correct `write' command, almost all the weights corresponding to the correct `write' are 0 for ALiBi (Fig.~\ref{fig:flipflop_weights}b), indicating that it is generally unable to find it.
For \fpl, we see a distinct shift between the correct and incorrect `write', with all weights greater than 0 for the correct `read' and `write' pair: (Fig.~\ref{fig:flipflop_weights}d).
After observing these fundamental changes to the attention weight structure, we now explore these biases' performances. 


\paragraph{Results} When comparing the RBCA accuracy on the OOD test set against standard MHA, both power-law biases (\fpl{} and \fspl) consistently outperform, while both NLP biases (\fe{} and \fbwn) underperform.
As noted by \citep{press2022train}, the OOD accuracy varies during training, but \fpl{} and \fspl{} consistently hover around 100\%, half the time being exactly 100\%.
Both NLP-inspired recency biases result in \textbf{worse} accuracy compared to vanilla MHA.
MHA hovers around 85\%, ALiBi (\fe) hovers around 75\%, and sliding window (\fbwn) performs very poorly.
Section~\ref{sm:flipflop_exp} provides full OOD test results.
These results indicate that recency biases significantly affect sequential read, write, and ignore capabilities.
The correct recency bias can be sufficient to endow the Transformer with these capabilities.

\begin{figure*}[t]
        \centering
        \includegraphics[width=0.98\textwidth]{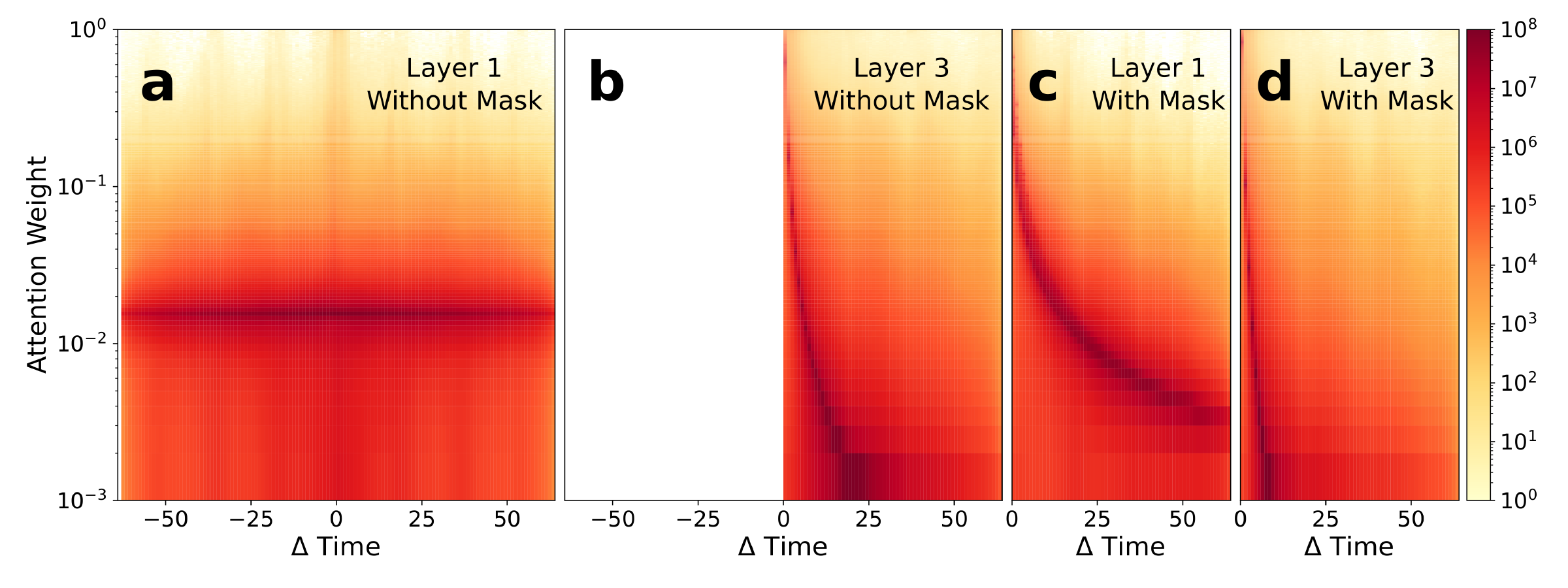}
        \caption{After training \emph{Powerformer}, we show the attention weights without the mask $\mathbf{M}^{(\text{C,L})}$ (a and b) and with the mask (c and d) for the first (a and c) and last (b and d) encoder layers. The linear trend illustrates the power law. These results are from the Weather dataset with \fpl{}, $\alpha=1.0$, and input/prediction lengths 512/96.}
        \label{fig:powerformer_attn_w_wo_mask}
\end{figure*}

\subsection{\label{sec:transformer}RBCA in Vanilla Transformer}

Given the success of power-law-biased RBCA on flip-flop, we further evaluate RBCA's capabilities on more challenging causal and local tasks, namely time-series forecasting.
Here, we take a principled approach by first testing the effects of RBCA on a standard encoder-decoder Transformer.
This is important, because adding more features to the model will convolute RBCA's effects and potentially confuse the results.

\paragraph{Datasets} Our time-series tasks use 7 standard benchmark datasets \citep{Zhou.informer.2021, wu2021autoformer}: ETT (4 subsets), Weather, Electricity, and Traffic. The ETT subsets measure the electricity transformer oil temperature at 2 different time scales (every 15 minutes and hourly) for two regions.
These are particularly interesting for testing the RBCA's sensitivity to sampling changes and distribution shifts.
Section~\ref{sec:datasets} provides a detailed description of these datasets, available in \citep{wu2021autoformer}.
We note that Weather, Electricity, and Traffic have much larger timescales, spanning from one to multiple years (Table~\ref{tab:data_size}).
These 7 datasets are large enough to capture slowly decaying temporal correlations (Appendix~\ref{sec:datasets}).

\paragraph{Evaluation}
Following previous works \citep{nie.patchtst.2023a, wu2021autoformer, zhou.fedformer.2022, woo2022etsformerexponentialsmoothingtransformers, liu2024itransformer, talukder2024totem, Kitaev2020Reformer, Zhou.informer.2021}, we train for multiple prediction lengths (96, 192, 336, and 720) and input lengths (336 and 512).
Each model is trained three times for each setting with different seeds.
For each dataset, we treat the recency bias, decay rate, and input sequence length as hyperparameters.
We select one input sequence for the entire dataset, and a recency bias and decay rate for each prediction length.
Sections~\ref{sm:exp_details} and \ref{sm:hyperparams} provide details on our experimental process and hyperparameters, respectively.
All benchmarks are trained with varying input sequences, and the optimal model is selected in the same way.

For each dataset, we evaluate the \fpl{} and \fspl{} biases.
We evaluate \fbwn{} on the ETT and weather datasets only, due to its poor performance on these datasets and the Transformer glitching benchmark.
We do not evaluate \fe{} due to its similarity to \fspl{} and inferior performance on the Transformer glitching benchmark.

\paragraph{Results} In the vast majority of cases, RBCA outperforms standard MHA, especially for datasets with slowly decaying long-range dependencies.
When RBCA outperforms MHA, it does so by much larger margins than when it underperforms.
For example, on the ETTh2 dataset RBCA and MHA outperform each other the same number of times, however, RBCA improves MSE and MAE by 16\% and underperforms by 2\%.
Table~\ref{tab:transformer_powerLaw_full} shows the full MSE and MAE results.

\subsection{\emph{Powerformer}}

To further investigate RBCA's effects on time-series forecasting, we add it to modern time-series architectures to determine how RBCA may affect state-of-the-art models.
Similar to our previous tests, our goal is to understand the effects of recency biases on attention.
For this reason, we develop a \textbf{simple} time-series model, allowing us to isolate the recency bias effects.
Adding many features convolutes the recency bias' effects and confuses the source of the outcome.

\paragraph{Model} We introduce \emph{Powerformer}, a time-series Transformer variant built atop PatchTST~\citep{nie.patchtst.2023a} by replacing MHA with RBCA.
\emph{Powerformer} integrates the encoder-only architecture, RBCA, and PatchTST's~\citep{nie.patchtst.2023a} successful patching framework.
PatchTST follows common time-series architecture designs: an encoder-only Transformer with a linear readout head.
We select PatchTST because it does not alter the Transformer, but instead only makes changes to the input embeddings by patching.
The lack of alterations to the attention mechanism in PatchTST will isolate the effects of RBCA on both attention weight distributions and forecasting performance.
While there have been more advances in time-series forecasting since PatchTST, such advancements make larger changes to the Transformer and attention architecture.
We believe that many of these changes will potentially confuse the results by making it challenging to determine where the improvement came from. 
Appendix~\ref{sm:powerformer_architecture} provides further details on \emph{Powerformer}'s architecture.

\paragraph{Evaluation} We evaluate \emph{Powerformer} on the same datasets and in the same manner as done in Section~\ref{sec:transformer}.
Sections~\ref{sm:exp_details} and \ref{sm:hyperparams} provide further details on our experimental process and hyperparameters, respectively.
We benchmark \emph{Powerformer} against models that focus on modifying the attention mechanism and models that use recency biases.
These models take a similar approach to \emph{Powerformer} and are thus a fair comparison.
We do not compare against all state-of-the-art models, as they contain techniques \emph{Powerformer} purposely lacks, but can be added to \emph{Powerformer}, or RBCA could be added to other models.
Either expanding \emph{Powerformer} or adding RBCA to other models is an exciting direction for future work.

\paragraph{Attention Analysis} Interestingly, \emph{Powerformer} strengthens our causal and recency biases beyond \maskCL, which is a key insight into the importance of the causal and recency biases.
At first glance, this is counterintuitive because the causal and recency biases act as regularizers that remove (and diminish contributions from) parts of the input sequence.
The removed data could otherwise be used to help the model memorize the output and overfit.
Moreover, since the recency bias mask (\maskL) is constant, \emph{Powerformer} has to capability to mitigate it by learning to counter the mask's regularization effects.

Rather than learning to remove the recency bias regularization, \emph{Powerformer} strengthens the recency bias.
To demonstrate this behavior, we look at the attention weight distribution with and without \maskCL on \emph{Powerformer} trained with \fpl{} ($\alpha=1$) and \maskCL.
Figure~\ref{fig:powerformer_attn_w_wo_mask} shows the attention weight distribution as a function of time delay between embeddings.
The first encoder layer without the mask (Fig.~\ref{fig:powerformer_attn_w_wo_mask}a) is relatively invariant to causality and locality (unbiased).
When \maskCL{} is applied to the first layer (Fig.~\ref{fig:powerformer_attn_w_wo_mask}c) the non-causal weights vanish and the power-law structure emerges.
This behavior is expected, unlike the behavior in the last layer.
In the last layer, the attention weight distribution \textbf{without} \maskCL{} (Fig.~\ref{fig:powerformer_attn_w_wo_mask}b) is entirely causal with a strong recency bias.
That is, even without applying the causal and recency bias, \emph{Powerformer} is enforcing causality and the power-law recency bias, rather than trying to remove this regularization.
Therefore, once \maskCL{} is applied to the last layer (Fig.~\ref{fig:powerformer_attn_w_wo_mask}d), the recency bias is further increased, as seen by the steeper slope.
Therefore, \emph{Powerformer} learns embeddings that further strengthen the effects of the recency bias, which is strong evidence that these biases are crucial in time-series forecasting.

\begin{table*}[!t]
	\centering
        \caption{\emph{Powerformer's} performance on standard publicly-available time-series datasets compared to similar state-of-the-art Transformer models. The best and second best results are bolded and underlined, respectively. 
    }
    \label{tab:performance}
     \resizebox{0.95\textwidth}{!}{
	\begin{tabular}{c|c|cc|cc|cc|cc|cc|cc}
	\toprule
	\multicolumn{2}{c|}{} & \multicolumn{2}{c|}{Powerformer} & \multicolumn{2}{c|}{PatchTST} & \multicolumn{2}{c|}{iTransformer} & \multicolumn{2}{c|}{One-Fits-All} & \multicolumn{2}{c|}{FEDformer} & \multicolumn{2}{c}{ETSformer} \\
	\midrule
	\multicolumn{2}{c|}{Metric} & MSE & MAE & MSE & MAE & MSE & MAE & MSE & MAE & MSE & MAE & MSE & MAE \\
	\midrule
	\multirow{4}{*}{\rotatebox[origin=c]{90}{\text{ETTh1}}}
		& 96 &  \textbf{0.361} & \textbf{0.390} &  \underline{0.370} & 0.400 &  0.401 & 0.417 &  0.376 & \underline{0.397} &  0.376 & 0.415 &  0.494 & 0.479 \\
		& 192 &  \textbf{0.395} & \textbf{0.410} &  \underline{0.413} & 0.429 &  0.440 & 0.445 &  0.416 & \underline{0.418} &  0.423 & 0.446 &  0.538 & 0.504 \\
		& 336 &  \textbf{0.406} & \textbf{0.420} &  \underline{0.422} & 0.440 &  0.455 & 0.454 &  0.442 & \underline{0.433} &  0.444 & 0.462 &  0.574 & 0.521 \\
		& 720 &  \textbf{0.434} & \textbf{0.455} &  \underline{0.447} & 0.468 &  0.520 & 0.511 &  0.477 & \underline{0.456} &  0.469 & 0.492 &  0.562 & 0.535 \\
	\midrule
	\multirow{4}{*}{\rotatebox[origin=c]{90}{\text{ETTh2}}}
		& 96 &  \textbf{0.269} & \textbf{0.334} &  \underline{0.274} & \underline{0.336} &  0.302 & 0.358 &  0.285 & 0.342 &  0.332 & 0.374 &  0.340 & 0.391 \\
		& 192 &  \textbf{0.330} & \textbf{0.375} &  \underline{0.339} & \underline{0.379} &  0.369 & 0.400 &  0.354 & 0.389 &  0.407 & 0.446 &  0.430 & 0.439 \\
		& 336 &  \textbf{0.323} & \textbf{0.380} &  \underline{0.331} & \textbf{0.380} &  0.407 & 0.430 &  0.373 & \underline{0.407} &  0.400 & 0.447 &  0.485 & 0.479 \\
		& 720 &  \textbf{0.370} & \textbf{0.417} &  \underline{0.379} & \underline{0.422} &  0.436 & 0.462 &  0.406 & 0.441 &  0.412 & 0.469 &  0.500 & 0.497 \\
	\midrule
	\multirow{4}{*}{\rotatebox[origin=c]{90}{\text{ETTm1}}}
		& 96 &  \textbf{0.285} & \textbf{0.342} &  \underline{0.290} & \textbf{0.342} &  0.301 & 0.355 &  0.292 & \underline{0.346} &  0.326 & 0.390 &  0.375 & 0.398 \\
		& 192 &  \textbf{0.328} & \textbf{0.368} &  \underline{0.332} & \underline{0.369} &  0.342 & 0.380 &  \underline{0.332} & 0.372 &  0.365 & 0.415 &  0.408 & 0.410 \\
		& 336 &  \textbf{0.357} & \textbf{0.385} &  \underline{0.366} & \underline{0.392} &  0.376 & 0.401 &  \underline{0.366} & 0.394 &  0.392 & 0.425 &  0.435 & 0.428 \\
		& 720 &  \textbf{0.410} & \textbf{0.414} &  0.420 & 0.424 &  0.440 & 0.438 &  \underline{0.417} & \underline{0.421} &  0.446 & 0.458 &  0.499 & 0.462 \\
	\midrule
	\multirow{4}{*}{\rotatebox[origin=c]{90}{\text{ETTm2}}}
		& 96 &  \textbf{0.163} & \textbf{0.251} &  \underline{0.165} & \underline{0.255} &  0.179 & 0.271 &  0.173 & 0.262 &  0.180 & 0.271 &  0.189 & 0.280 \\
		& 192 &  \textbf{0.218} & \textbf{0.290} &  \underline{0.220} & \underline{0.292} &  0.236 & 0.310 &  0.229 & 0.301 &  0.252 & 0.318 &  0.253 & 0.319 \\
		& 336 &  \textbf{0.273} & \textbf{0.326} &  \underline{0.278} & \underline{0.329} &  0.284 & 0.342 &  0.286 & 0.341 &  0.324 & 0.364 &  0.314 & 0.357 \\
		& 720 &  \textbf{0.358} & \textbf{0.380} &  \underline{0.367} & \underline{0.385} &  0.370 & 0.394 &  0.378 & 0.401 &  0.410 & 0.420 &  0.414 & 0.413 \\
	\midrule
	\multirow{4}{*}{\rotatebox[origin=c]{90}{\text{Weather}}}
		& 96 &  \textbf{0.147} & \textbf{0.197} &  \underline{0.149} & \underline{0.198} &  0.163 & 0.211 &  0.162 & 0.212 &  0.238 & 0.314 &  0.197 & 0.281 \\
		& 192 &  \textbf{0.191} & \textbf{0.239} &  \underline{0.194} & \underline{0.241} &  0.204 & 0.253 &  0.204 & 0.248 &  0.275 & 0.329 &  0.237 & 0.312 \\
		& 336 &  \textbf{0.243} & \textbf{0.279} &  \underline{0.245} & \underline{0.282} &  0.256 & 0.291 &  0.254 & 0.286 &  0.339 & 0.377 &  0.298 & 0.353 \\
		& 720 &  \textbf{0.310} & \textbf{0.329} &  \underline{0.314} & \underline{0.334} &  0.326 & 0.337 &  0.326 & 0.337 &  0.389 & 0.409 &  0.352 & 0.388 \\
	\midrule
	\multirow{4}{*}{\rotatebox[origin=c]{90}{\text{Electricity}}}
		& 96 &  \textbf{0.129} & \underline{0.223} &  \textbf{0.129} & \textbf{0.222} &  \underline{0.131} & 0.226 &  0.139 & 0.238 &  0.186 & 0.302 &  0.187 & 0.304 \\
		& 192 &  \textbf{0.145} & \textbf{0.240} &  \underline{0.147} & \textbf{0.240} &  0.152 & \underline{0.247} &  0.153 & 0.251 &  0.197 & 0.311 &  0.199 & 0.315 \\
		& 336 &  \textbf{0.162} & \textbf{0.257} &  \underline{0.163} & \underline{0.259} &  0.168 & 0.263 &  0.169 & 0.266 &  0.213 & 0.328 &  0.212 & 0.329 \\
		& 720 &  0.198 & \underline{0.289} &  \underline{0.197} & 0.290 &  \textbf{0.191} & \textbf{0.284} &  0.206 & 0.297 &  0.233 & 0.344 &  0.233 & 0.345 \\
	\midrule
	\multirow{4}{*}{\rotatebox[origin=c]{90}{\text{Traffic}}}
		& 96 &  0.365 & \underline{0.252} &  \underline{0.360} & \textbf{0.249} &  \textbf{0.352} & 0.258 &  0.388 & 0.282 & 0.576 & 0.359 &  0.607 & 0.392 \\
		& 192 &  0.382 & \underline{0.258} &  \underline{0.379} & \textbf{0.256} &  \textbf{0.368} & 0.267 &  0.407 & 0.290 & 0.610 & 0.380 &  0.621 & 0.399 \\
		& 336 &  \underline{0.391} & \underline{0.265} &  0.392 & \textbf{0.264} &  \textbf{0.384} & 0.273 &  0.412 & 0.294 &  0.608 & 0.375 &  0.622 & 0.396 \\
		& 720 &  \underline{0.430} & \textbf{0.286} &  0.432 & \textbf{0.286} &  \textbf{0.417} & \underline{0.290} &  0.450 & 0.312 &  0.621 & 0.375 &  0.632 & 0.396 \\
        \midrule
        \multicolumn{2}{c|}{$1^\text{st} / 2^\text{nd}$} & \multicolumn{2}{c|}{\textbf{46 / 7}} & \multicolumn{2}{c|}{\underline{9 / 38}} & \multicolumn{2}{c|}{6 / 3} & \multicolumn{2}{c|}{0 / 10} & \multicolumn{2}{c|}{0 / 0} & \multicolumn{2}{c}{0 / 0} \\
	\bottomrule
	\end{tabular}}%
\end{table*}

\paragraph{Results} Compared to our benchmarks (Table~\ref{tab:performance}), \emph{Powerformer} achieves the best performance against more complicated (FEDformer~\citep{zhou.fedformer.2022}, ETSformer~\citep{woo2022etsformerexponentialsmoothingtransformers}, One-Fits-All~\citep{zhou2023OFA}) models.
\emph{Powerformer} achieves the best performance in 46 forecasting tasks compared to the next best model (PatchTST) which achieves 9. 
We note that \emph{Powerformer} outperforms ETSformer, which employs an exponential temporal decay without attention.
This result reinforces the importance of correctly balancing RBCA's recency bias with attentions ability to learn data-dependent correlations.
Notably, \emph{Powerformer} outperforms One-Fits-All~\citep{zhou2023OFA}, which is a pre-trained GPT-2 model more than twice \emph{Powerformer's} size.

Qualitatively, \emph{Powerformer} outperforms the next best model (PatchTST) by capturing high-frequency variability.
\emph{Powerformer} better models sharper changes in oscillations and deviations from a generic sinusoidal signal, as demonstrated in Appendix~\ref{sm:visualizations}.
This behavior can be described in the context of the power-law's heavy tail, which induces a multi-scale attention structure.
During attention, the embeddings in the long tail will be summed together with relatively similar weights since the heavy tail and the unbiased attention weights (Fig.~\ref{fig:maskPL}) are relatively constant.
This summation removes short-lived input sequence frequency components with periods less than the tail's length.
This is because the summation covers multiple, if not many, periods of these frequency components, causing them to cancel out. 
However, long-term signals with periods greater than the tail length will be captured.
Therefore, the tail highlights longer-term dynamics in the input sequence.
Conversely, the quickly varying power-law near 0 amplifies a shorter context window.
This shorter window highlights local higher-frequency components by reducing the period at which the signal is washed out in the summation.
Again, striking a balance between attending to long-term sinusoidal signals and quickly decaying exponential pair-wise distributions.
Therefore, the power-law recency bias endows \emph{Powerformer} with an implicit temporal multiscale structure: the tail focuses on low-frequency signal while high-frequency components come from the most recent signal.

\emph{Powerformer} underperforms for the Traffic dataset and outperforms by a smaller margin in the Electricity dataset due to these datasets' highly periodic nature.
As highlighted, \emph{Powerformer} can capture short and long-term sinusoidal signals, but long and constant attention windows -- used in all the other benchmarks -- are better suited to highly-periodic signals.
In cases where the data is highly periodic, like Traffic (Fig.~\ref{fig:Traffic_correlations}) and Electricity (Figs.~\ref{fig:Electricity_correlations} and \ref{fig:electricity_forecast}), these systems can often be effectively modeled by simpler Fourier or statistical methods without requiring the much larger training and evaluation overhead of Transformers.
\emph{Powerformer} is built as a general-purpose time-series method to handle challenging non-stationary dynamics.

\emph{Powerformer} is sensitive to the recency bias and aligns $f(t)$ with the dataset's specific pairwise correlation.
To test this, we evaluated the first 5 datasets on \fbwn, which induces a flat pairwise correlation structure.
The \fpl{} and \fspl{} biases consistently outperform \fbwn, see Tables~\ref{tab:powerformer_powerLaw_results}, \ref{tab:powerformer_simPowerLaw_results}, \ref{tab:powerformer_butter1_results}, and \ref{tab:powerformer_butter2_results} for the full evaluation results of \fpl, \fspl, \fbwo, and \fbwt{}, respectively.
The superior performance of \fpl{} and \fspl{} over \fbwn{} demonstrates \emph{Powerformer's} sensitivity to the dataset's correlation structure.

In addition to distinguishing between biases that naturally align with the data distribution, \emph{Powerformer} is also sensitive to the dataset's correlation time scale.
The ETTh and ETTm datasets test \emph{Powerformer's} ability to adapt the timescale $\alpha$ of \maskL{} for datasets sampled from the same distribution but with different temporal resolution.
Since ETTh and ETTm are sampled on the hour and minute timescales, respectively, ETTm requires a slowly decaying mask to access the same information as a quickly decaying mask for ETTh.
As expected, \emph{Powerformer} with \fspl{} (Table~\ref{tab:powerformer_simPowerLaw_results}) favors fast decaying masks on ETTh and slowly decaying masks on ETTm.
This illustrates \emph{Powerformer's} sensitivity to adjust its timescale to cover the same information content in each dataset.

\subsection{Ablation}
We ablate \emph{Powerformer} to determine the effects of the causal mask (\maskC) and the addition of the recency bias (\maskL).
Table~\ref{tab:ablation} compares the performance of \emph{Powerformer} with MHA (without \maskCL), with only the causal mask (\maskC), and with RBCA (\maskCL).
We first observe that causal-only attention (\maskC) outperforms MHA, validating the importance of our causal inductive bias.
In all but one case, RBCA ties or outperforms causal-only attention.

To learn the temporal information timescale, we make $\alpha$ learnable instead of treating it as a hyperparameter.
Section~\ref{sm:learnable_decay_times} outlines the experimental details and full results.
During training, $\alpha$ drops consistently and monotonically.
This is expected as a small magnitude $\alpha$ facilitates overfitting by providing the model with extraneous (long-term) information.
To keep $\alpha$ from converging to 0, we initialize $\alpha$ with the hyperparameter tuning values and quickly decay the learning rate.
The learned $\alpha$ values decrease in magnitude, but we observe very little difference between the learnable and constant $\alpha$ (Table~\ref{tab:powerformer_learnable_time}).
This indicates that RBCA regularizes the model, removing superfluous information to avoid overfitting and improve generalization.

\section{CONCLUSION}
We develop RBCA to test the importance of causal and recency biases in Transformers, as they are applied to causal and locally correlated NLP and time-series tasks.
Some recency biases (\fpl) are sufficient requirements to imbue Transformer with advantageous RNN capabilities: read and write to memory, and ignore input information.
However, common NLP recency biases (\fe{} and \fbwn) diminish these capabilities.
This indicates that locality biases are important for Transformers to gain important capabilities, but the bias' implementation matters.

We develop \emph{Powerformer} to impose and investigate the importance of causal and recency biases for Transformer-based time-series models.
\emph{Powerformer}, with its simple architecture, outperforms larger and more complex models.
Notably, \emph{Powerformer} strengthens the causal and recency biases by enforcing a power-law decay before the \maskCL{} is applied.
This is strong evidence that the causal and recency biases are important for time-series problems, as \emph{Powerformer} is essentially removing information at training time that could be used to overfit.

\section*{Acknowledgements}

The authors thank Dongwei Lyu for her help in running the flip flop experiments to test the RBCA Transformer glitching. NBE would like to acknowledge the Laboratory Directed Research and Development Program of Lawrence Berkeley National Laboratory under U.S. Department of Energy Contract No. DE-AC02-05CH11231, and the Scientific Discovery through Advanced Computing (SciDAC) program, under Contract Number DE-AC02-05CH11231 at Berkeley Lab, and NERSC (DE-AC02-05CH11231) for providing compute resources.
\newpage

\bibliography{references} 

\section*{Checklist}

\begin{enumerate}

  \item For all models and algorithms presented, check if you include:
  \begin{enumerate}
    \item A clear description of the mathematical setting, assumptions, algorithm, and/or model. [Yes]
    \item An analysis of the properties and complexity (time, space, sample size) of any algorithm. [Yes]
    \item (Optional) Anonymized source code, with specification of all dependencies, including external libraries. [Yes]
  \end{enumerate}

  \item For any theoretical claim, check if you include:
  \begin{enumerate}
    \item Statements of the full set of assumptions of all theoretical results. [Yes]
    \item Complete proofs of all theoretical results. [Yes]
    \item Clear explanations of any assumptions. [Yes]     
  \end{enumerate}

  \item For all figures and tables that present empirical results, check if you include:
  \begin{enumerate}
    \item The code, data, and instructions needed to reproduce the main experimental results (either in the supplemental material or as a URL). [Yes]
    \item All the training details (e.g., data splits, hyperparameters, how they were chosen). [Yes]
    \item A clear definition of the specific measure or statistics and error bars (e.g., with respect to the random seed after running experiments multiple times). [Yes]
    \item A description of the computing infrastructure used. (e.g., type of GPUs, internal cluster, or cloud provider). [Yes]
  \end{enumerate}

  \item If you are using existing assets (e.g., code, data, models) or curating/releasing new assets, check if you include:
  \begin{enumerate}
    \item Citations of the creator If your work uses existing assets. [Yes]
    \item The license information of the assets, if applicable. [Yes]
    \item New assets either in the supplemental material or as a URL, if applicable. [Yes]
    \item Information about consent from data providers/curators. [Yes]
    \item Discussion of sensible content if applicable, e.g., personally identifiable information or offensive content. [Not Applicable]
  \end{enumerate}

  \item If you used crowdsourcing or conducted research with human subjects, check if you include:
  \begin{enumerate}
    \item The full text of instructions given to participants and screenshots. [Not Applicable]
    \item Descriptions of potential participant risks, with links to Institutional Review Board (IRB) approvals if applicable. [Not Applicable]
    \item The estimated hourly wage paid to participants and the total amount spent on participant compensation. [Not Applicable]
  \end{enumerate}

\end{enumerate}

\clearpage
\thispagestyle{empty}

\setcounter{equation}{0}
\setcounter{figure}{0}
\setcounter{table}{0}
\setcounter{page}{1}
\setcounter{section}{0}
\makeatletter
\renewcommand \thesection{S\@arabic\c@section}
\renewcommand{\theequation}{S\arabic{equation}}
\renewcommand{\thefigure}{S\arabic{figure}}
\renewcommand{\thetable}{S\arabic{table}}
\onecolumn
\aistatstitle{Supplementary Materials}

\section{\label{sec:filters}LOCALITY INDUCING DECAY FUNCTIONS}

\subsection{Power-Law Functions}

The two power-law-based masks \fpl{} and \fspl{} apply a power-law decay to the attention weights and scores, respectively.
Here, we provide intuition for each mask type a well as illustrate the masks used in our evaluation.

\begin{figure}[!ht]
\begin{minipage}{0.495\textwidth}
    \centering
    \includegraphics[width=\linewidth]{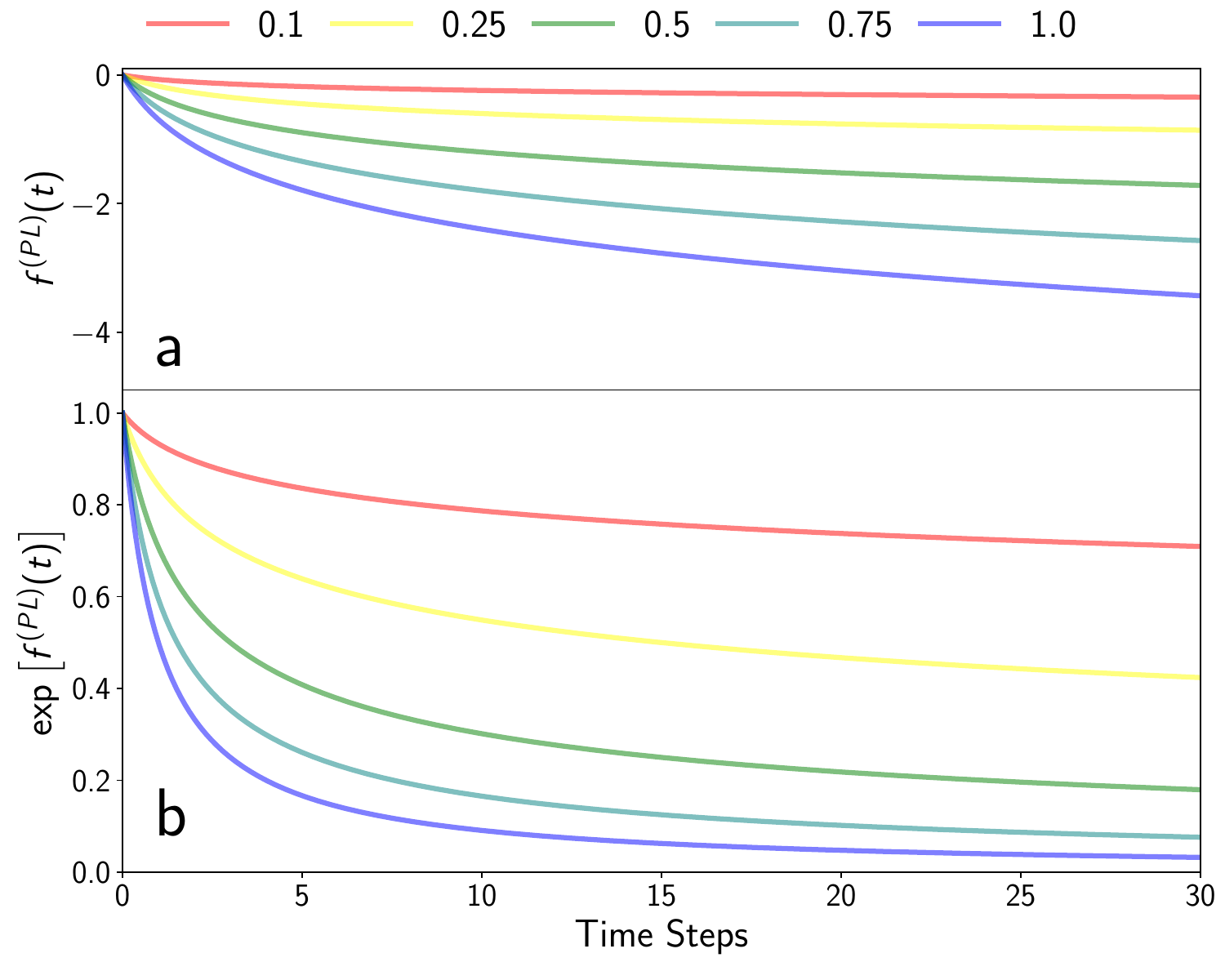}
    \caption{The power-law masks (\fpl) used when evaluating RBCA on Transformer and when evaluating \emph{Powerformer}. Panel (a) shows the additive mask, (b) shows the multiplicative effects on the attention weights.}
    \label{fig:powerlaw_masks}
\end{minipage}
\hfill
\begin{minipage}{0.495\textwidth}
    \centering
    \includegraphics[width=\linewidth]{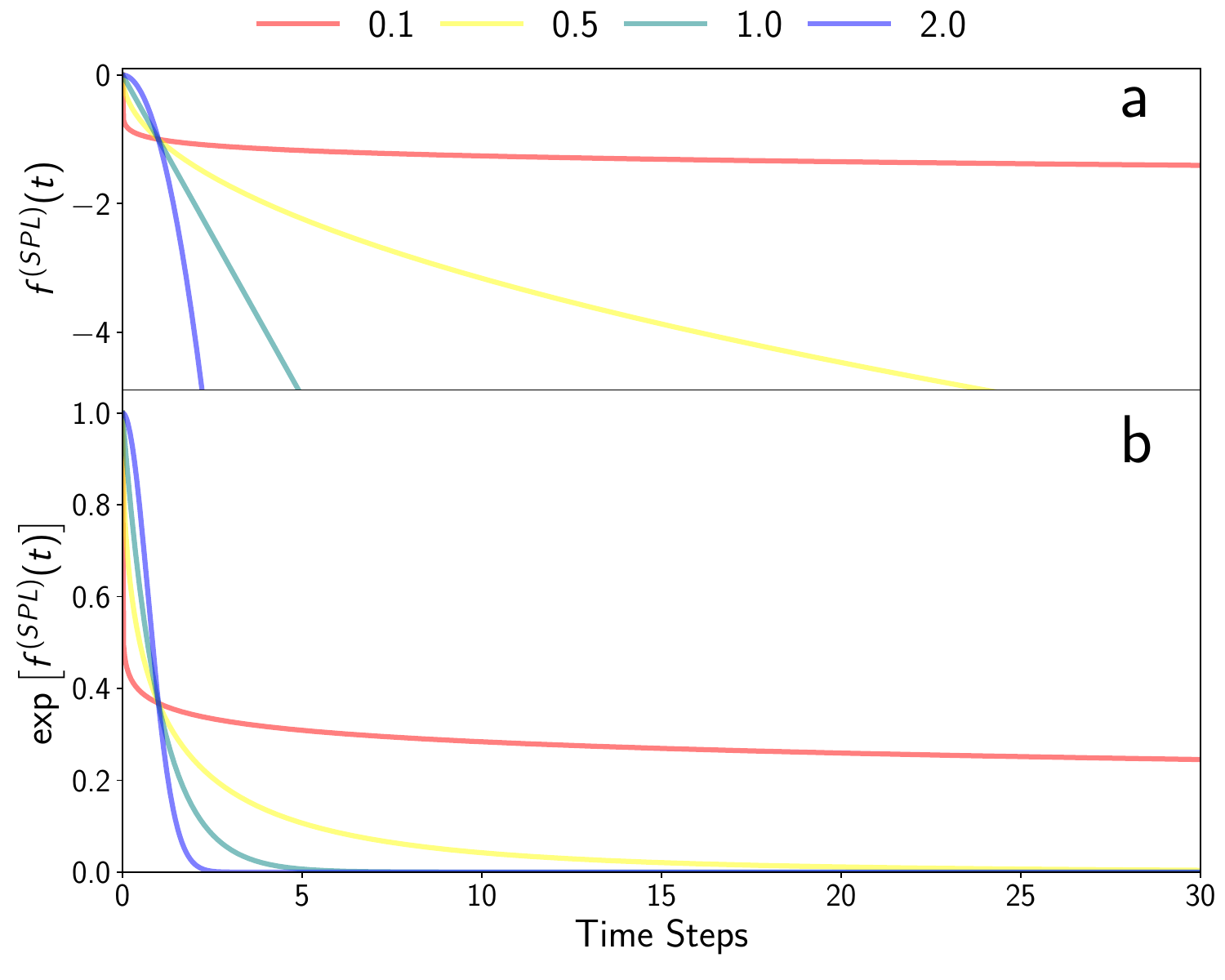}
    \caption{The similarity power-law masks (\fspl) used when evaluating RBCA on Transformer and when evaluating \emph{Powerformer}. Panel (a) shows the additive mask, (b) shows the multiplicative effects on the attention weights.}
    \label{fig:simpowerlaw_masks}
\end{minipage}
\end{figure}

We consider inducing a multiplicative power-law decay to the attention weights (\fpl) to be the most natural way of inducing locality:
\begin{equation}
    \fplmm = -\alpha\,\log(t) .
\end{equation}
The attention weights are analogous to correlation coefficients.
Therefore, since power-law distributions show up in correlation structures, we would like the attention weights to follow a similar power-law distribution.
When training Transformer and \emph{Powerformer} with \fpl{} we used decay times of $\alpha \in [0.1, 0.25, 0.5, 0.75, 1.0]$.
Figure~\ref{fig:powerlaw_masks} shows the additive attention mask and the resulting multiplicative effects on the attention weights.

We additionally apply the power-law directly to the similarity score:
\begin{equation}
    \fsplmm = -(t)^{-\alpha} .
\end{equation}
During the attention calculation, we get an exponential of this power-law, which results in steep initial decay and a long tail.
When training Transformer and \emph{Powerformer} with \fspl{} we used decay times of $\alpha \in [0.1, 0.5, 1.0, 2.0]$.
Figure~\ref{fig:simpowerlaw_masks} shows the additive attention mask and the resulting multiplicative effects on the attention weights.

\begin{figure}[!ht]
    \centering
    \includegraphics[width=0.5\linewidth]{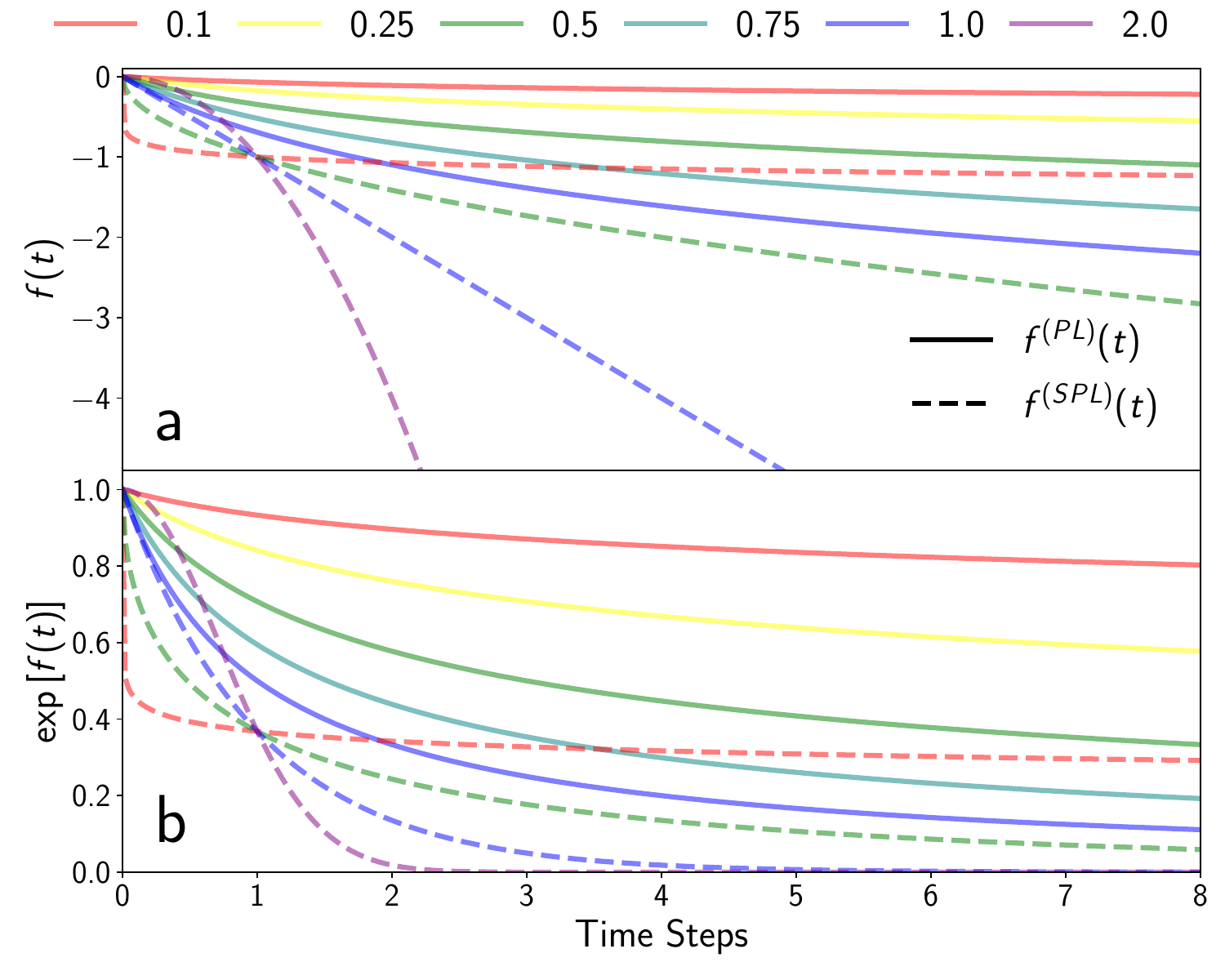}
    \caption{The \fpl{} and \fspl{} mask (a) and their effects on the weights (b) are shown to illustrate the coverage of decays used when assessing \emph{Powerformer}.}
    \label{fig:all_powerlaw_masks}
\end{figure}

Combining the \fpl{} and \fspl{} masks, we cover decays starting from a relatively flat mask (\fpl{} with $\alpha=0.5$) to a sharply decaying (\fspl{} with $\alpha = 2.0$).
To illustrate our coverage we plot \fpl{} and \fspl{} together in Fig.~\ref{fig:all_powerlaw_masks}.

\subsection{Butterworth Filter\label{sec:BW}}

\begin{figure}[!ht]
\centering
\includegraphics[width=0.5\linewidth]{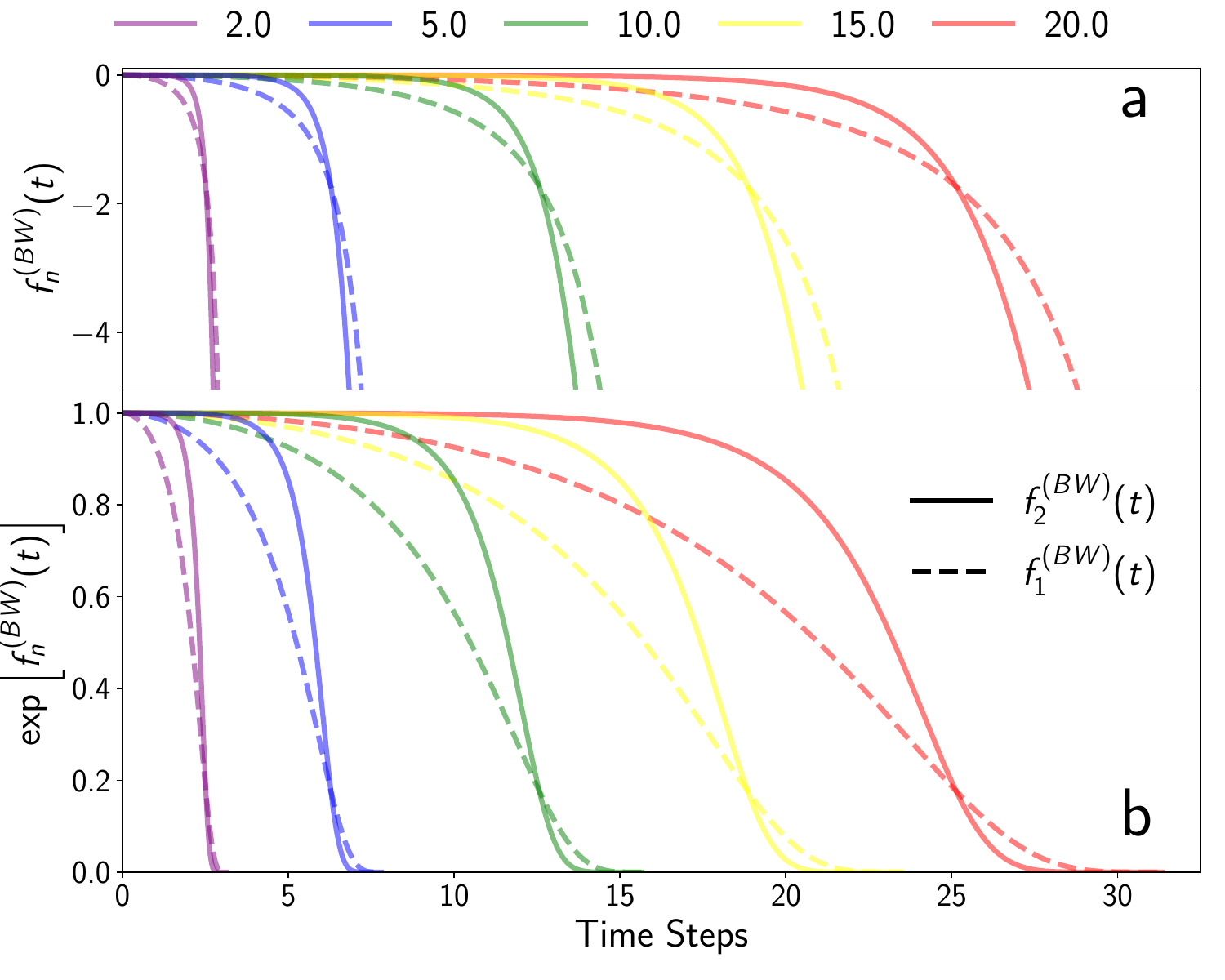}
\caption{We show the effects of the Butterworth filter masks $\left( \fbwnmm \right)$ for orders 1 (dashed lines) and 2 (solid lines), while varying the critical time. 
Panel (a) shows the effects on the attention scores and Panel (b) shows the corresponding effects on the attention weights after applying Softmax to \fbwn.}
\label{fig:maskBF}
\end{figure}

The Butterworth filter~\citep{butterworth.filter.1930} is often used in signal processing for low-, high-, and band-pass filters.
It is designed to be optimally flat within the passband and sometimes referred to as the ``maximally flat filter."
The motivation behind it's derivation is to have equal sensitivity to frequency modes within the passband.
We leveraged the Butterworth filter to equally weight temporal measurements within the lookback window.
The analog Butterworth filter gain is given by 
\begin{equation}
    f^{(\text{BW})}_n(z) = \frac{-1}{\sqrt{1 + \left( \frac{z}{t_\text{c}} \right)^{2n}}} ,
\end{equation}
where $t_c$ is the critical time that sets the width of the filter and the order $n$ determines how fast the gain decays after $t_c$.
In this work we use the digital Butterworth filter gain.
Figure~\ref{fig:maskBF} shows the \fbwo{} and \fbwt{} contributions on the attention score and weights for varying decay constants.

\subsubsection{Digital Filter Gain\label{sc:butterGain}}
To transform the analog Butterworth filter into a digital filter, the filter design must be discretized.
This discretization is often done using the bilinear transform
\begin{equation}
    z = \frac{2}{T}\frac{z-1}{z+1} ,
\end{equation}
where $T$ is the numerical integration step size used in the trapezoidal rule.

\subsubsection{Code to Calculate the Gain}
For reproducibility, we provide the code used to calculate the digital Butterworth filter gains.
We multiply the gain by 5 to scale with the attention key-query overlap values.

\begin{verbatim}
    import numpy as np
    import scipy as sp
    
    def butterworth_filter(scale, order, times):
        b, a = sp.signal.butter(order, 0.8, 'lowpass', analog=False)
        t, decay = sp.signal.freqz(b, a)
        t = scale*t/2
        dc = 5*np.log(np.abs(decay))
        decay_interp = sp.interpolate.interp1d(t, dc)
        return decay_interp(times)
\end{verbatim}

\section{DATASETS\label{sec:datasets}}
We evaluate \emph{Powerformer} on 7 real-world datasets that have become standard public benchmarks for time-series forecast tasks. The datasets can be found and downloaded in Ref.~\citep{wu2021autoformer}, or individually at references~below.

\begin{table}[!ht]
    \centering
    \caption{We provide the number of variables and the number of timesteps for each dataset.}
    \label{tab:data_size}
    \vskip 0.1in
    \begin{tabular}{c|cccccc}
        \toprule
        Datasets &  Illness & ETTh* & ETTm* & Weather & Electricity & Traffic \\
        \midrule
        Features &  7 & 7 & 7 & 21 & 321 & 862 \\
        Timesteps & 966 & 17420 & 69680 & 52696 & 26304 & 17544 \\
        \bottomrule
    \end{tabular}
\end{table}

\begin{figure*}[!htb]
    \centering
    \includegraphics[width=0.98\textwidth]{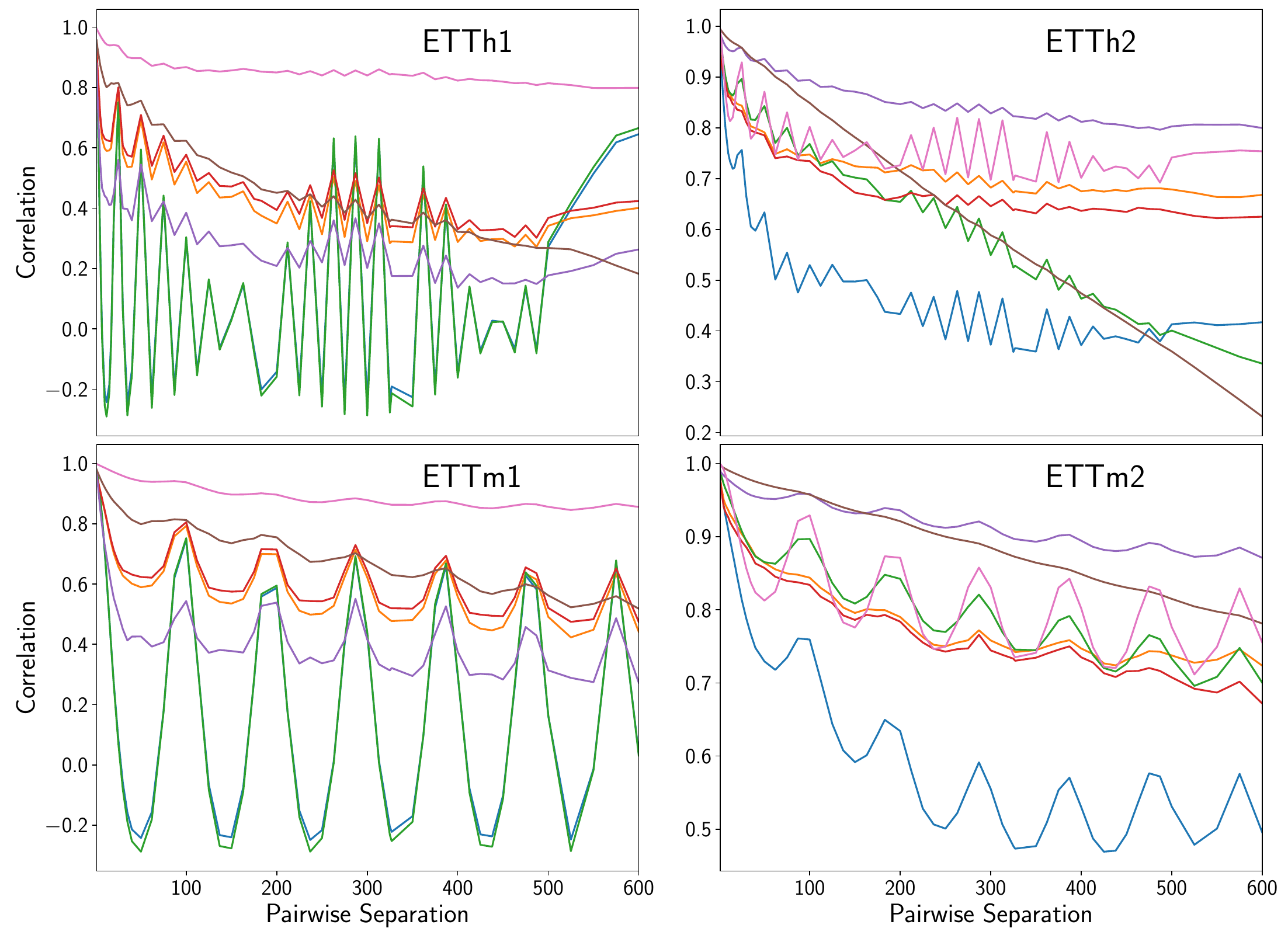}
    \caption{We show the ETT datasets' pairwise correlations as a pairwise separation function. Each colored line represents a different variable in the dataset.}
    \label{fig:ETT_correlations}
\end{figure*}

\hspace{1cm} \textbf{ETT}\footnote{https://github.com/zhouhaoyi/ETDataset}~\citep{Zhou.informer.2021} provides 7 measurements of the electrical transformer power load (including oil temperature) between July 2016 and July 2018. This is done over 2 regions in China at 2 different sampling rates (hourly and every 15 minutes), resulting in 4 separate datasets (ETTh1, ETTh2, ETTm1, ETTm2). We show the pairwise correlation dependence in Fig.~\ref{fig:ETT_correlations}.

\begin{figure}[!hb]
\begin{minipage}{.495\textwidth}
    \includegraphics[width=1.0\textwidth]{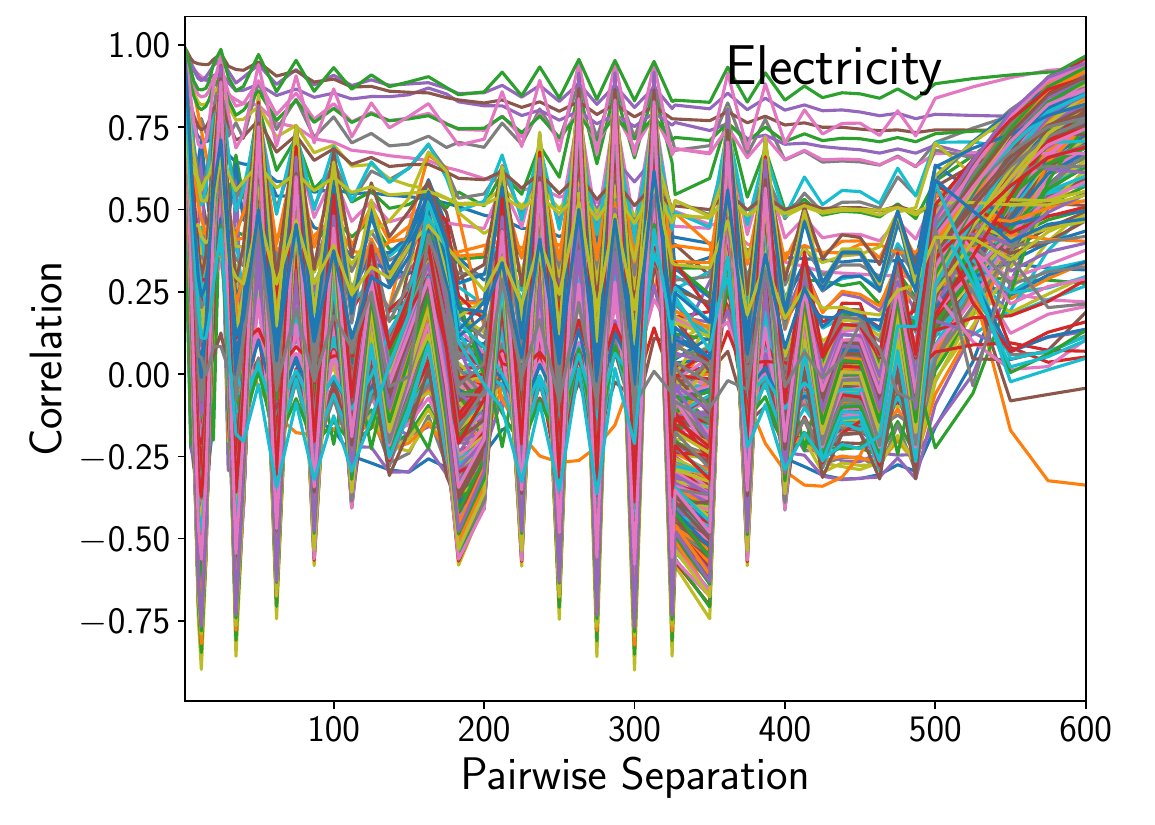}
    \caption{We show the Electricity dataset's pairwise correlations as a pairwise separation function. Each colored line represents a different variable in the dataset.}
    \label{fig:Electricity_correlations}
\end{minipage}
\hfill    
\begin{minipage}{.495\textwidth}
    \includegraphics[width=1.0\textwidth]{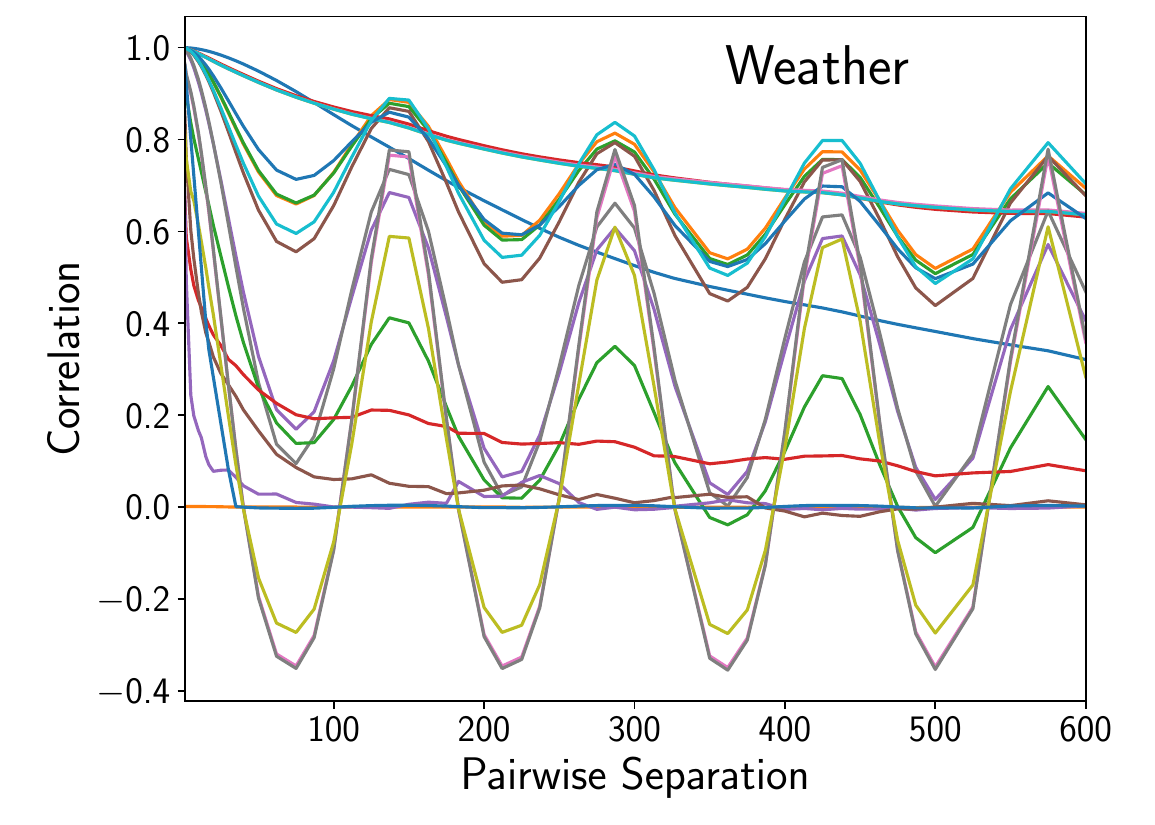}
    \caption{We show the Weather dataset's pairwise correlations as a pairwise separation function. Each colored line represents a different variable in the dataset.}
    \label{fig:Weather_correlations}
\end{minipage}
\end{figure}

\hspace{1cm} \textbf{Weather}\footnote{https://www.bg-ena.mpg.de/wetter/}~\citep{wu2021autoformer} provides 21 meteorological measurements (air temperature, air pressure, humidity, precipitation, etc.) collected in Germany over the whole of 2020. These measurements are recorded every 10 minutes. We show the pairwise correlation dependence in Fig.~\ref{fig:Weather_correlations}.

\hspace{1cm} \textbf{Electricity}\footnote{https://archive.ics.uci.edu/
    dataset/321/electricityloaddiagrams20112014}~\citep{wu2021autoformer} provides the electricity consumption (kWh) of 321 consumers from 2012 to 2014. These measurements are sampled every hour. We show the pairwise correlation dependence in Fig.~\ref{fig:Electricity_correlations}.
    
\begin{figure}[!htb]
    \centering
    \includegraphics[width=0.47\linewidth]{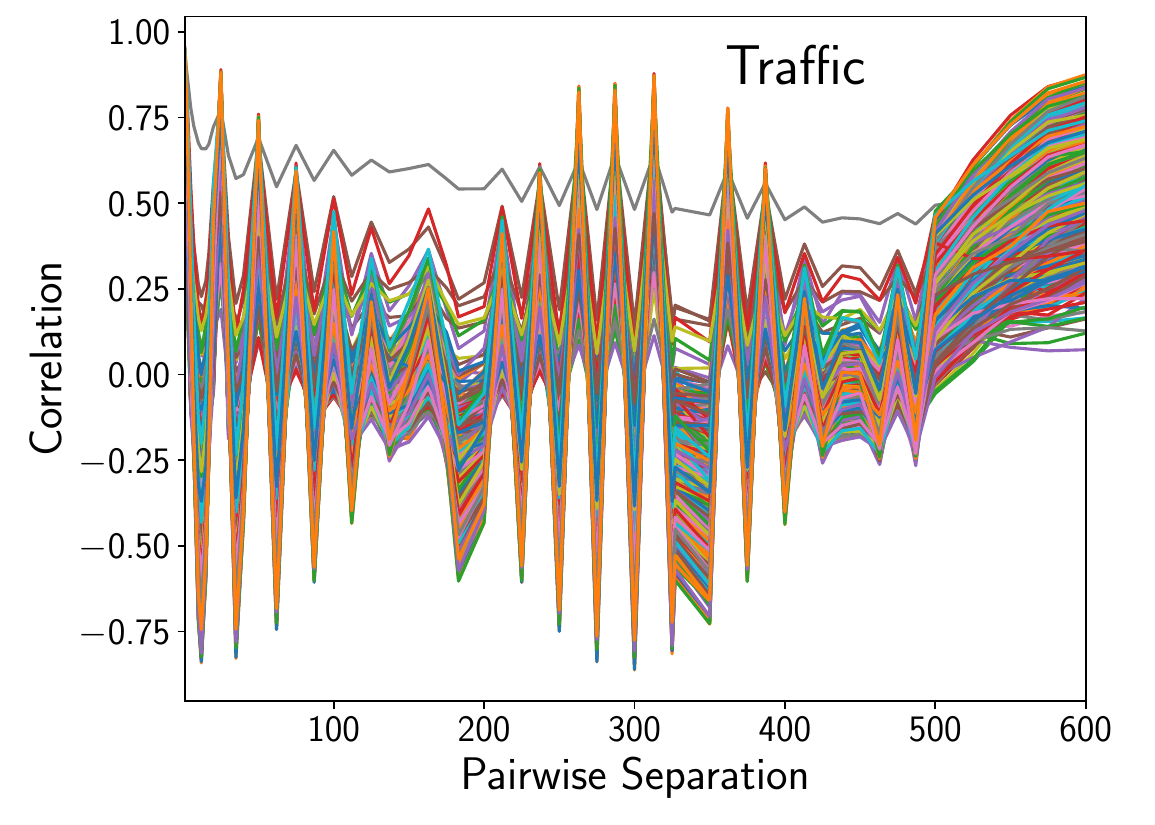}
    \caption{We show the Weather dataset's pairwise correlations as a pairwise separation function. Each colored line represents a different variable in the dataset.}
    \label{fig:Traffic_correlations}
\end{figure}

\hspace{1cm} \textbf{Traffic}\footnote{http://pems.dot.ca.gov}~\citep{wu2021autoformer} provides occupancy rates on San Francisco Bay Area freeways from 826 sensors. This data comes from the California Department of Transportation and is sampled hourly. We show the pairwise correlation dependence in Fig.~\ref{fig:Traffic_correlations}.

\FloatBarrier

\section{ARCHITECTURE DETAILS}
\subsection{Transformer}
\label{sm:transformer_architecture}
We use a vanilla encoder-decoder Transformer architecture for all Transformer experiments.
The decoder input is the last 48 time-steps in the sequence concatenated by an array of zeros with a length of the prediction window.
For our baseline experiment, the decoder self-attention has a causal mask, which is common in Transformer architectures, while the encoder self-attention and decoder cross-attention do not have masks.
Our RBCA variant of the Transformer applies RBCA to both the encoder and decoder self-attention, but not the decoder cross-attention.
We do not apply RBCA to the decoder cross-attention because this process is already causal as it forecasts future points based on the given sequence.
We present the Transformer hyperparameters used in this work in Table~\ref{tab:powerformer_params}.

\subsection{PatchTST}
\label{sm:patchtst_architecture}
Here, we review PatchTST's patching mechanism, as PatchTST is otherwise similar to a traditional Transformer with a linear readout head that simultaneously predicts all future points.
For a detailed description of PatchTST, see Ref.~\citep{nie.patchtst.2023a}.
Let us consider an input time-series $\mathbf{X}$ with dimensions $[B, T_\text{seq}, D]$, where $B$ is batch size, $T_\text{seq}$ is the time interval, and $D$ is the number of input dimensions.
The multivariate input is first normalized to zero mean and unit variance along the time axis and flattened into $D$ univariate inputs, resulting in a dimensionality of $[B\times D, T_\text{seq}]$.
PatchTST batches these time series via a convolution with $N$ filters, a stride $s$, and patch size $p$, resulting in a dimensionality of $[B \times D, P, N]$.
Here, $P = (T_\text{seq}-p)//s + 1$ is the number of patches, and $//$ is integer division.
At this point, the patched data is fed into the Transformer where a positional embedding is added and the patched embeddings are further contextualized by vanilla MHA~\citep{vaswani.transformer.2017}.
The linear output head takes the $[B \times D, P, N]$ output of the Transformer, flattens the last two dimensions, and applies a linear matrix multiplication (of shape $[P \times N, T_\text{pred}]$) to predict the output time series of length $T_\text{pred}$.
The linear head returns a matrix of shape $[B \times D, T_\text{pred}]$.
This matrix is reshaped into the original multivariate dimensionality $([B, T_\text{pred}, D])$ and renormalized.
We note that the patching mechanism is performed before applying the Transformer and is independent of it.

\subsection{Powerformer}
\label{sm:powerformer_architecture}
\emph{Powerformer} forecasts univariate time-series by combining RBCA, the commonly used encoder-only Transformer architecture, and patching~\citep{nie.patchtst.2023a}.
We selected PatchTST as the base of \emph{Powerformer} since PatchTST already includes decomposes multivariate time-series into univariate channels and performs the patching.
Additionally, similar to other modern time-series Transformer-based models~\citep{zhou2023OFA, liu2024itransformer, talukder2024totem}, PatchTST~\citep{nie.patchtst.2023a} uses an encoder-only architecture with a linear readout head.
\emph{Powerformer} uses a standard Transformer encoder architecture with the primary difference being the replacement of the vanilla MHA by RBCA.
We additionally remove the dropout applied to the attention weights because our implicit biases enforce a deterministic temporal dependency on pairwise correlations.
Dropout enforces redundancy between pairwise correlations and is therefore in conflict with our implicit biases.
The input data comes in with $D$ variates and shape $[B, D, T_\text{seq}]$, where $B$ is the batch size and $T_\text{seq}$ is the input sequence length.
The PatchTST patching mechanism (described in Appendix~\ref{sm:patchtst_architecture}) returns the patched and normalized univariate embeddings with shape $[B \times D, P, N]$, where $P$ is the number of patches and $N$ is the size of the embeddings.
These embeddings are encoded through multiple RBCA Transformer encoder layers (see Appendix~\ref{sm:hyperparams} for hyperparameter details).
The encoder output (of shape $[B \times D, P, N]$) is flattened along the last two dimensions ($[B \times D, P \times N]$) in preparation for the linear readout head.
The linear readout matrix has the shape $[P \times N, T_\text{pred}]$ and after acting on the flattened encoder output produces the univariate forecasts of shape $[B \times D, T_\text{pred}]$.
Finally, these univariate forecasts are renormalized and reshaped into the multivariate input structure $[B, D, T_\text{pred}]$ and returned to the user.
Consequently, \emph{Powerformer} is a simple Transformer-based model with causal and local masks added to the attention scores $\mathbf{S}$, and patching applied to the input data.
We present the \emph{Powerformer} hyperparameters used in this work in Table~\ref{tab:powerformer_params}.

\subsection{Hyperparameter Tuning and Selection}
\label{sm:hyperparams}
In addition to tuning standard Transformer hyperparameters, \emph{Powerformer} introduces two patching and two recency-bias hyperparameters.
The patching mechanism is determined by the patch length and the patch stride.
The patch length is the number of input tokens that are averaged together.
The patch stride is the stride of the patching mechanism that helps reduce the size of the input data.
For the recency-bias mask, one can tune the type of mask functional and it's decay time.
As previously described, and illustrated in Fig.~\ref{fig:maskPL}, we test four masks: power-law mask (\fpl), similarity power-law mask (\fspl), and two Butterworth masks for $n\in{1,2}$ (\fbwn).
For each mask, one may try different decay constants $\alpha$ that determine how fast the mask decays.
In this works we tried the following decay constants for each mask: for \fpl{} we evaluate $\alpha \in \{ 0.1, 0.25, 0.5, 0.75, 1.0 \}$; for \fspl{} we evaluate $\alpha \in \{ 0.1, 0.5, 1.0, 2.0 \}$; and for \fbwn{} we evaluate $\alpha \in \{ 2, 5, 10, 15, 20 \}$.
Although \emph{Powerformer} introduces two more hyperparameters over PatchTST, we observe in Tables~\ref{tab:powerformer_powerLaw_results}, \ref{tab:powerformer_simPowerLaw_results}, \ref{tab:powerformer_butter1_results}, and \ref{tab:powerformer_butter2_results} that many masks are not very sensitive to changes in $\alpha$ within a certain range.
For this reason, tuning $\alpha$ will likely be very fast as it does not require much fine-tuning after finding this large range.

We note that in addition to the hyperparameters previously discussed, we try two different input sequence lengths (336, 512).
There is no standard measurement rate for general time-series datasets, and in general, their dynamics are quite different and happen on different time scales.
Thus, the amount of variation within some number of time steps depends on both the dynamics being measured and the measurement rate.
The input sequence length must therefore change to encompass the same amount of information (or variation) for datasets with differing measurement rates and temporal variability.

To reproduce our results, we list our hyperparameters used on each dataset for \emph{Powerformer} (Table~\ref{tab:powerformer_params}) and Transformer (Table~\ref{tab:transformer_params}).
For some models, we apply weight decay to the encoder only, leaving the forecasting head without regularization.
For each dataset and model, we trained with the following 3 random seeds: 2021, 1776, and 1953.
\begin{table}[!ht]
    \centering
    \caption{We provide \emph{Powerformer} architecture and training parameters used for each dataset.}
    \label{tab:powerformer_params}
    \vskip 0.1in
    \resizebox{\textwidth}{!}{
    \begin{tabular}{c|c|c|c|c|c|c|c}
        \toprule
        \textbf{Parameters} & \textbf{ETTh1} & \textbf{ETTh2} & \textbf{ETTm1} & \textbf{ETTm2} & \textbf{Weather} & \textbf{Electricity} & \textbf{Traffic} \\
        \midrule
        Sequence Length     & 336 / 512   & 336 / 512   & 336 / 512   & 336 / 512   & 336 / 512 & 336 / 512   & 336 / 512  \\
        Patch Length        & 16    & 16    & 16    & 16    & 16    & 16    & 16    \\
        Patch Stride        & 8     & 8     & 8     & 8     & 8     & 8     & 8     \\
        Encoder Layers      & 3     & 3     & 3     & 3     & 3     & 3     & 3     \\
        Embedding Size      & 16    & 16    & 128   & 128   & 128   & 128   & 128   \\
        MHA Heads           & 4     & 4     & 16    & 16    & 16    & 16    & 16    \\
        MHA Feed Forward    & 128   & 128   & 256   & 256   & 256   & 256   & 256   \\
        Dropout \%          & 30    & 30    & 20    & 20    & 20    & 20    & 20    \\
        Linear Dropout \%   & 30    & 30    & 20    & 20    & 20    & 20    & 20    \\
        Train Epochs        & 100   & 100   & 100   & 100   & 100   & 100   & 100   \\
        Patience            & --    & --    & 20    & 20    & 20    & 10    & 10    \\
        Learning Rate       & $10^{-3}$&$10^{-4}$&$10^{-5}$&$10^{-4}$&$10^{-4}$&$10^{-4}$&$10^{-4}$ \\
        Weight Decay        & 1.0   & 1.0   & --    & 0.05  & --    & --    & --    \\
        Batch Size          & 128   & 128   & 128   & 128   & 128   & 32    & 24\\
        \bottomrule
    \end{tabular}}
\end{table}

\begin{table}[!ht]
    \centering
    \caption{We provide Transformer architecture and training parameters used for each dataset.}
    \label{tab:transformer_params}
    \vskip 0.1in
    \resizebox{\textwidth}{!}{
    \begin{tabular}{c|c|c|c|c|c|c|c}
        \toprule
        \textbf{Parameters} & \textbf{ETTh1} & \textbf{ETTh2} & \textbf{ETTm1} & \textbf{ETTm2} & \textbf{Weather} & \textbf{Electricity} & \textbf{Traffic} \\
        \midrule
        Sequence Length     & 256   & 256   & 256   & 256   & 256   & 256   & 256  \\
        Decoder Input Length        & 48    & 48    & 48    & 48    & 48    & 48    & 48 \\
        Encoder Layers      & 2     & 2     & 2     & 2     & 2     & 2     & 2     \\
        Decoder Layers      & 1     & 1     & 1     & 1     & 1     & 1     & 1     \\
        Embedding Size      & 512    & 512    & 512   & 512   & 512   & 512   & 512   \\
        MHA Heads           & 8     & 8     & 8    & 8    & 8    & 8    & 8    \\
        MHA Feed Forward    & 2048   & 2048   & 2048   & 2048   & 2048   & 2048   & 2048   \\
        Dropout \%          & 5    & 5    & 5    & 5    & 5    & 5    & 5    \\
        Linear Dropout \%   & 5    & 5    & 5    & 5    & 5    & 5    & 5    \\
        Train Epochs        & 100   & 100   & 100   & 100   & 100   & 100   & 100   \\
        Patience            & --    & --    & --    & --    & --    & --    & --    \\
        Learning Rate       & $10^{-4}$&$10^{-4}$&$10^{-4}$&$10^{-4}$&$10^{-4}$&$10^{-4}$&$10^{-4}$ \\
        Batch Size          & 128   & 128   & 128   & 128   & 128   & 128   & 128   \\
        \bottomrule
    \end{tabular}}
\end{table}
\FloatBarrier

\section{FORECASTING VISUALIZATION}
\label{sm:visualizations}

Here, we show \emph{Powerformer} forecasting visualizations and compare various decay time constants against ground truth and PatcTST~\citep{nie.patchtst.2023a}.
We also highlight plots where both \emph{Powerformer} and PatchTST struggle.
In these cases, \emph{Powerformer} generally outperforms by better capturing faster-moving features, non-periodic features, and is better able to alter periodic predictions.
We believe this is due to \emph{Powerformer's} ability to focus on and amplify local high-frequency features while dampening or removing long-time high-frequency components.
RBCA achieves this through the power-law masks \fpl{} and \fspl, which amplify local correlations and dampen long-time correlations.

\begin{figure}[!hb]
\centering
\begin{minipage}{.495\textwidth}
    \includegraphics[width=1.0\textwidth]{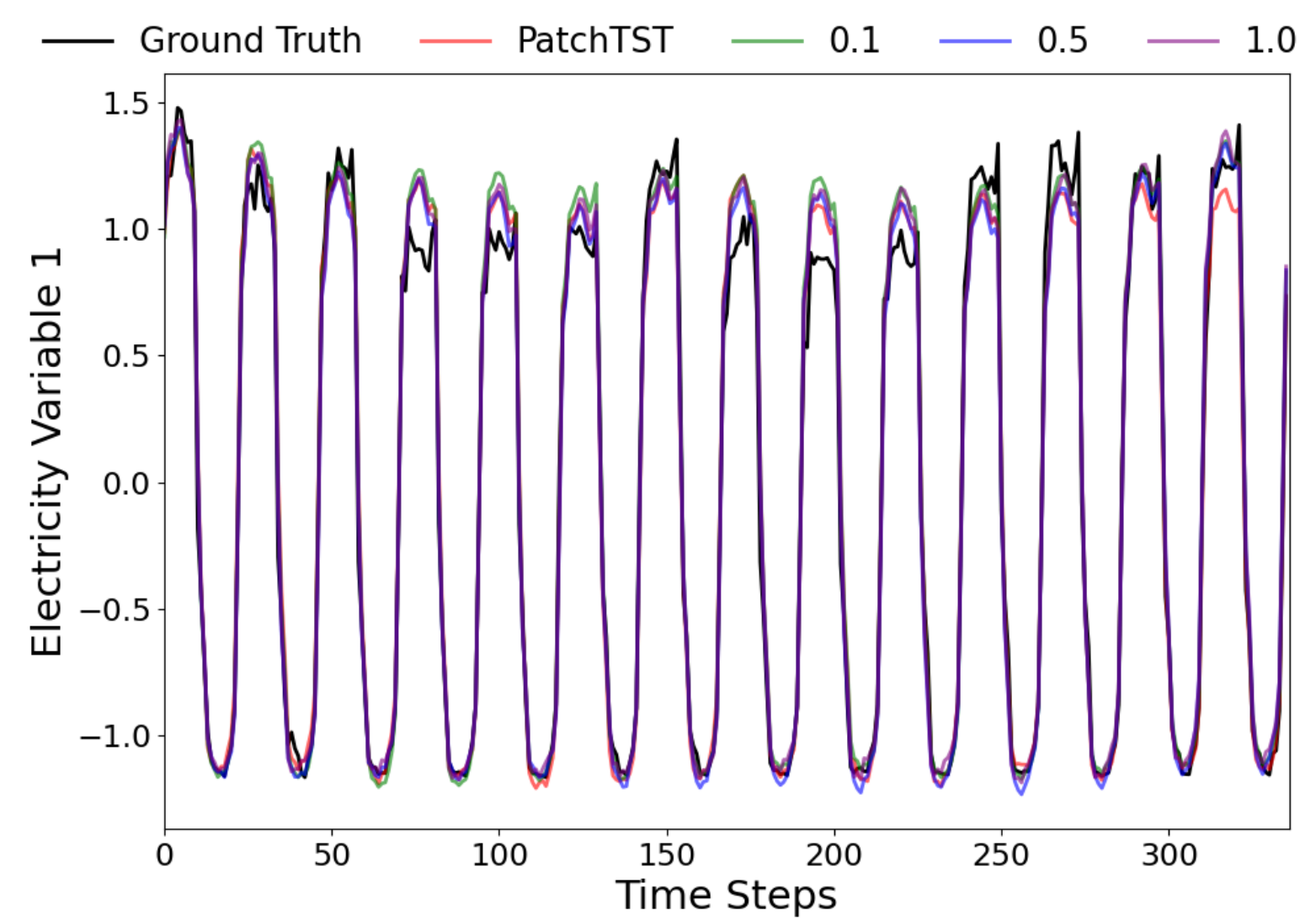}
\end{minipage}
\hfill    
\begin{minipage}{.495\textwidth}
    \includegraphics[width=1.0\textwidth]{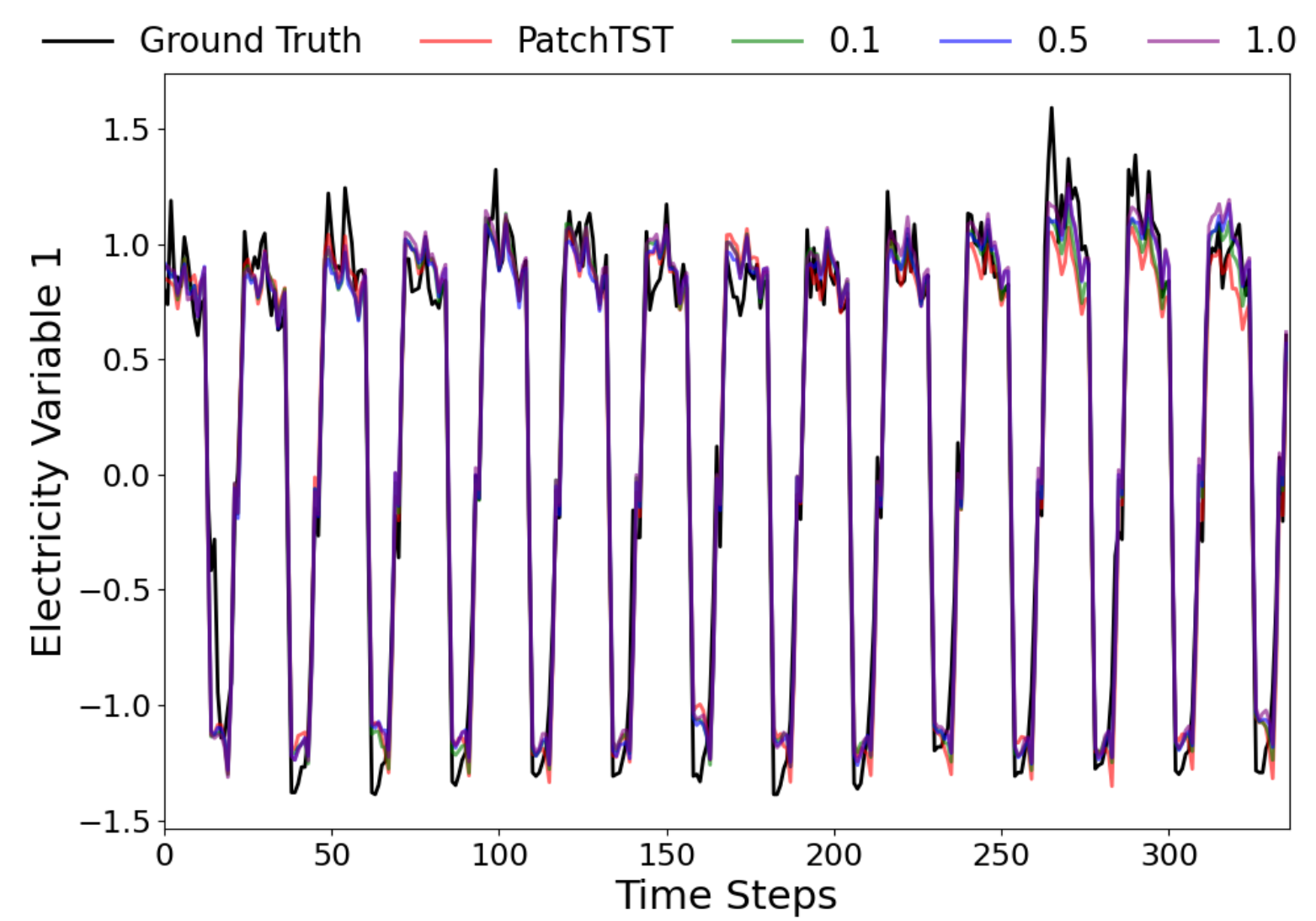}
\end{minipage}
\begin{minipage}{.495\textwidth}
    \includegraphics[width=1.0\textwidth]{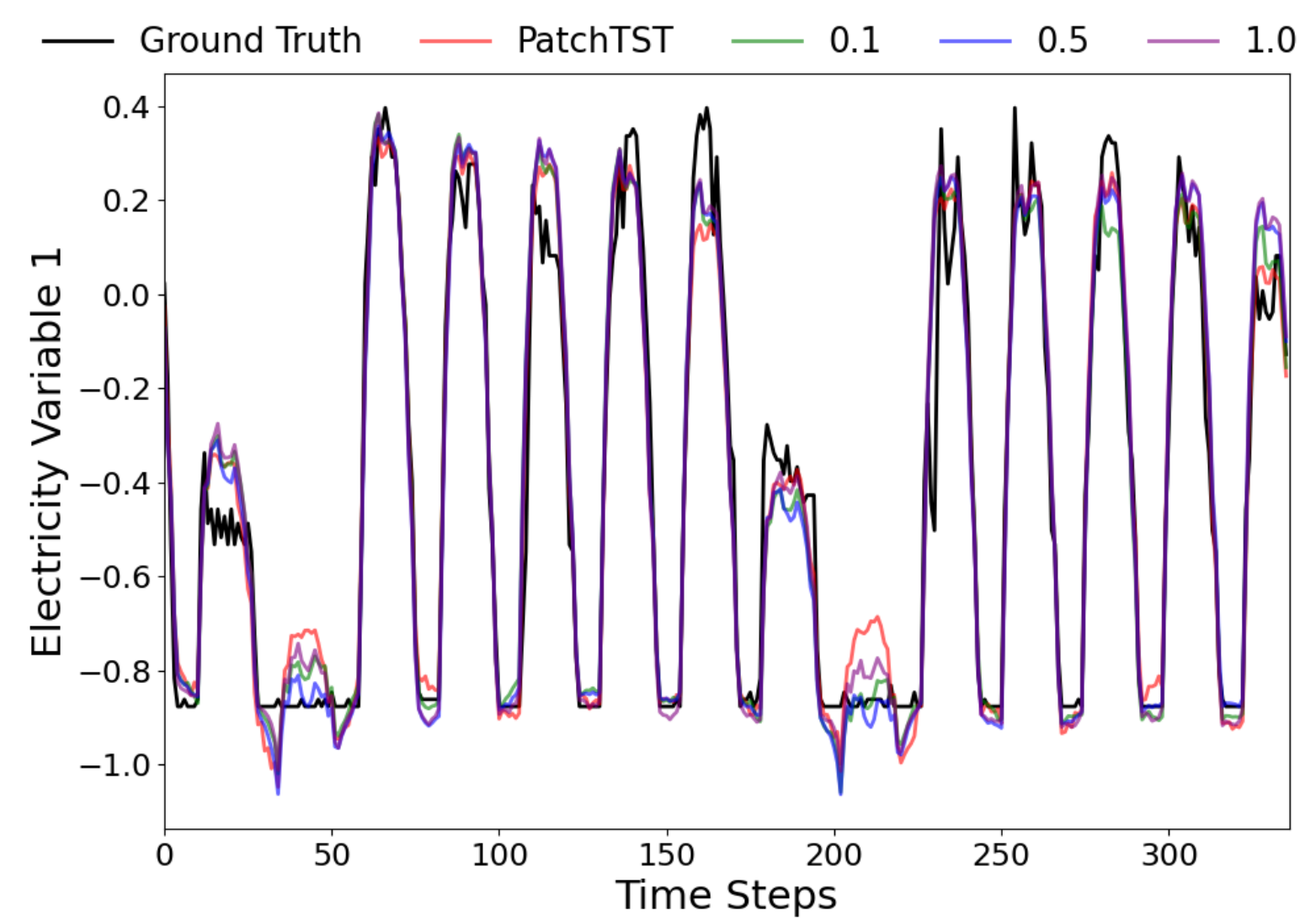}
\end{minipage}
\hfill    
\begin{minipage}{.495\textwidth}
    \includegraphics[width=1.0\textwidth]{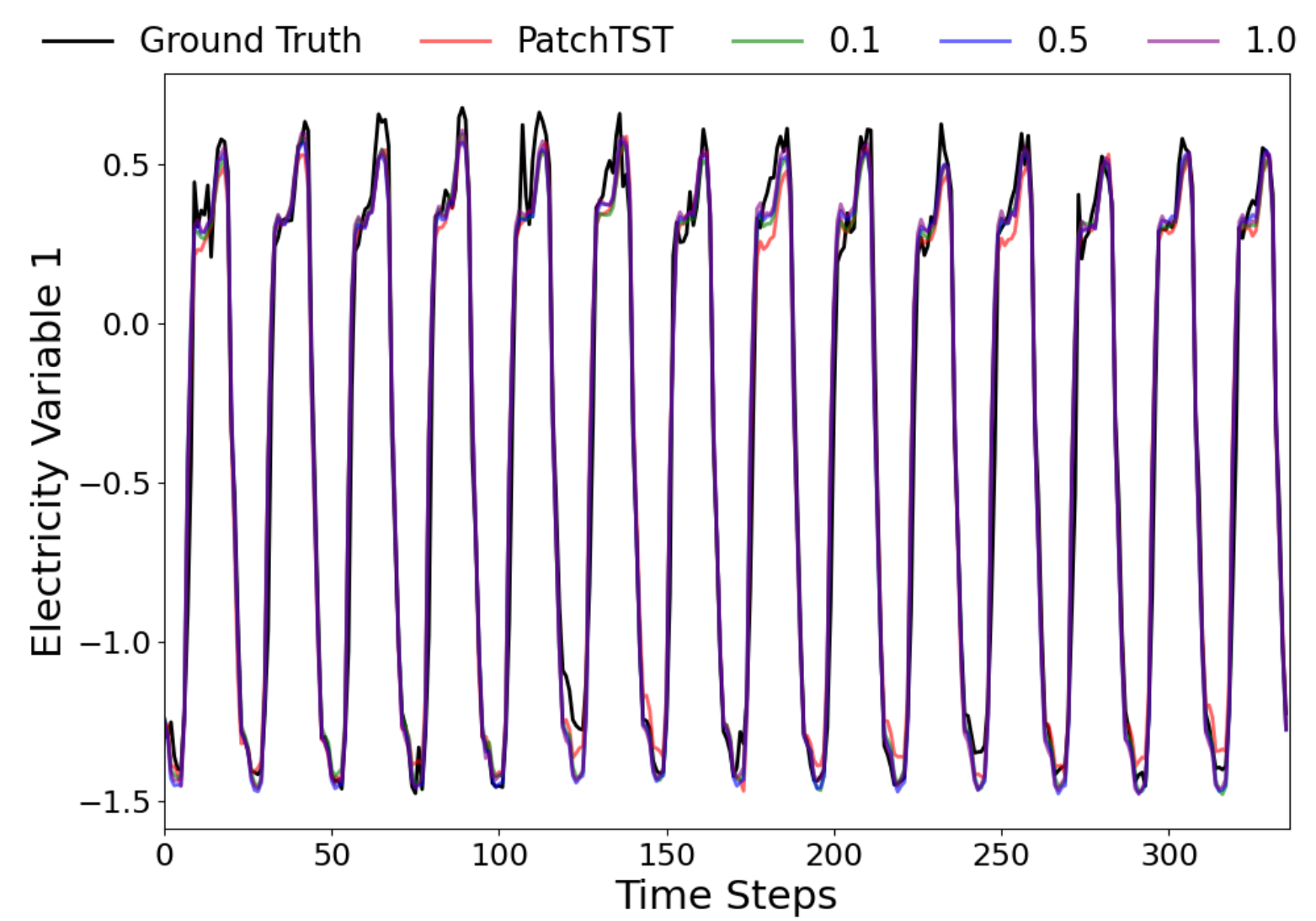}
\end{minipage}
\caption{\label{fig:electricity_forecast}We show forecasting results on the Electricity dataset for \emph{Powerformer} with \fpl{} and PatchTST. The colored lines represent PatchTST and different \fpl{} decay constants. For these forecasts, the sequence length is 512 and the prediction length is 336.}
\end{figure}

\begin{figure}[!hb]
\begin{minipage}{.495\textwidth}
    \includegraphics[width=1.0\textwidth]{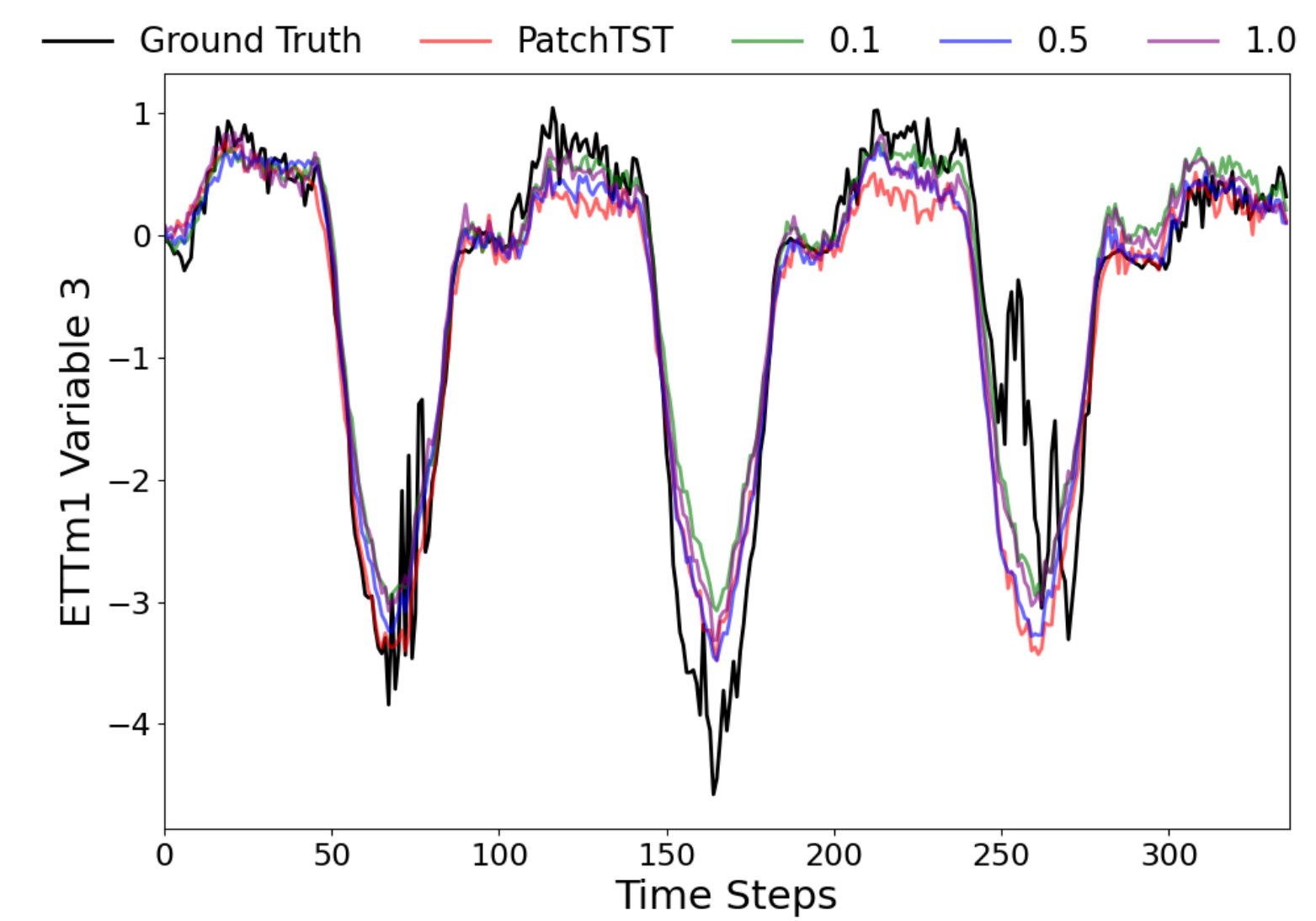}
\end{minipage}
\hfill    
\begin{minipage}{.495\textwidth}
    \includegraphics[width=1.0\textwidth]{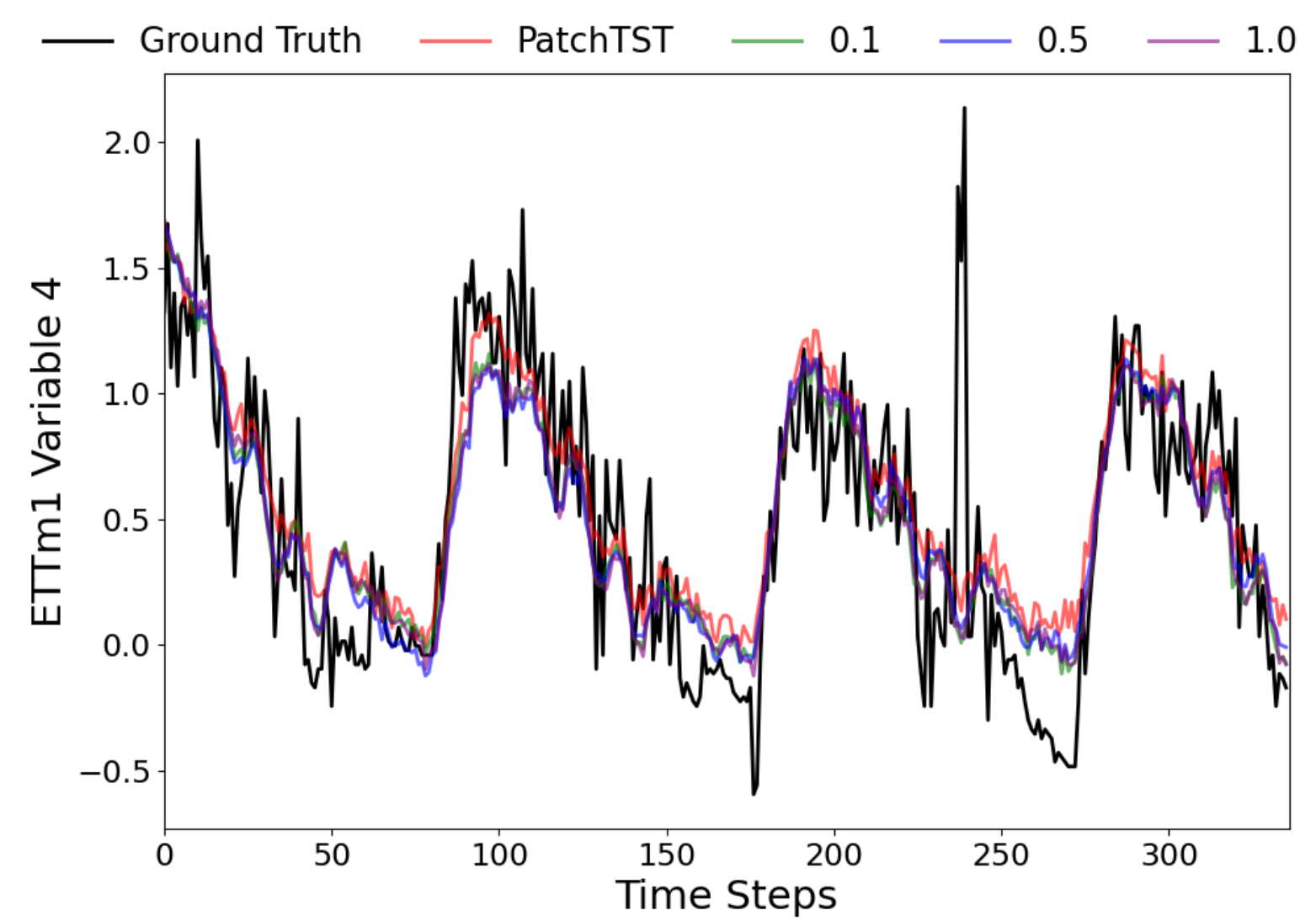}
\end{minipage}
\begin{minipage}{.495\textwidth}
    \includegraphics[width=1.0\textwidth]{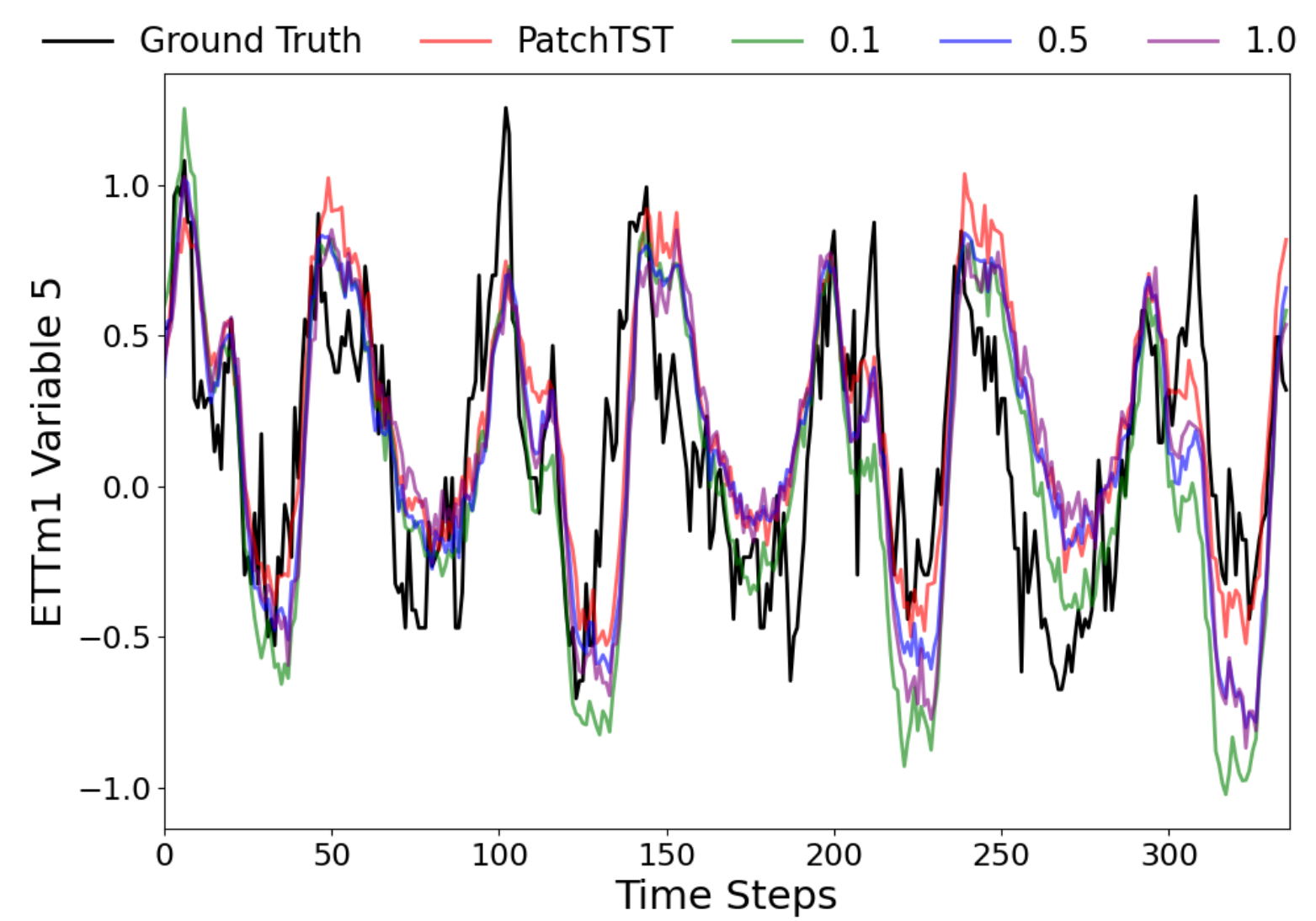}
\end{minipage}
\hfill 
\begin{minipage}{.495\textwidth}
    \includegraphics[width=1.0\textwidth]{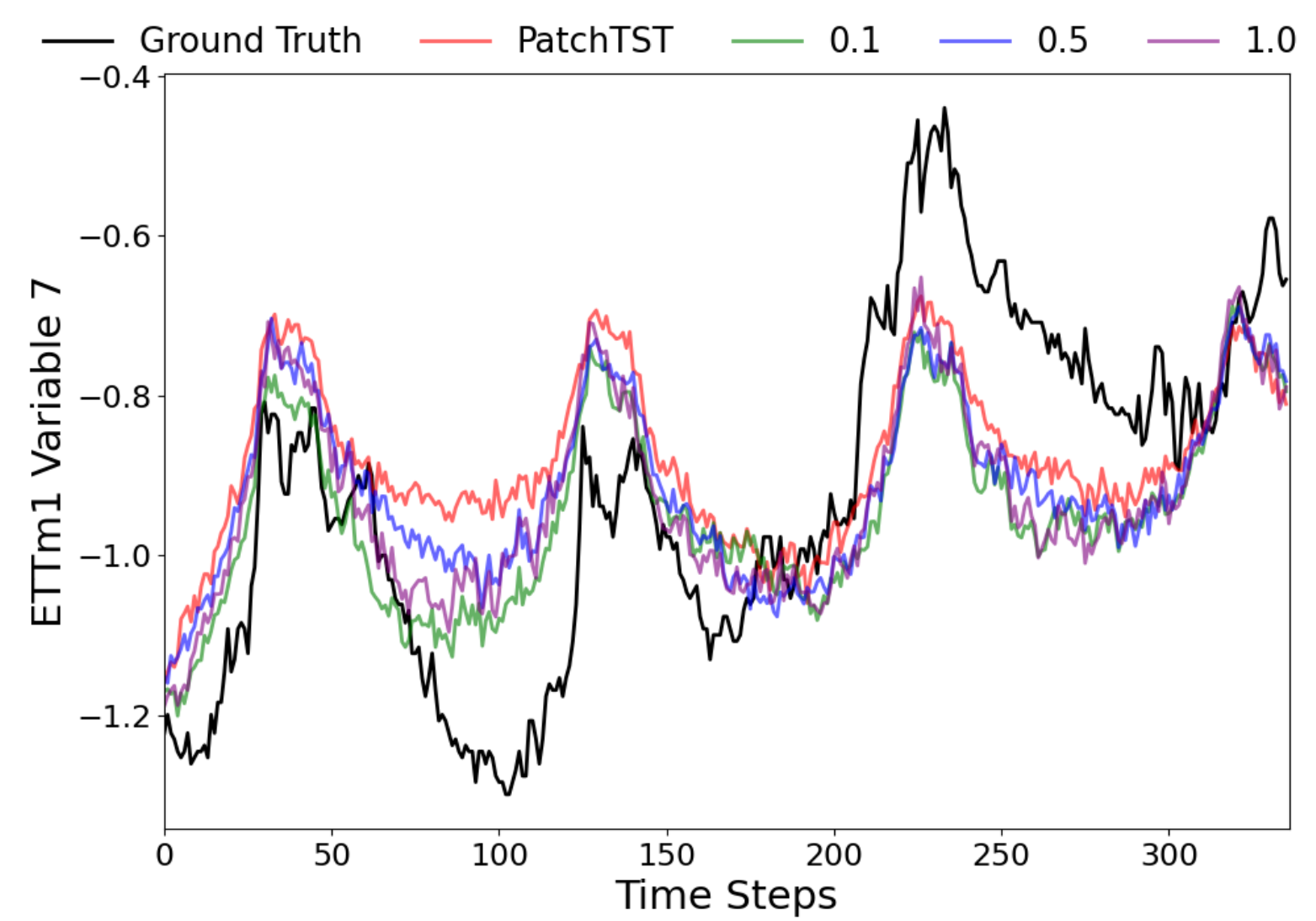}
\end{minipage}
\caption{We show forecasting results on the ETTm1 dataset for \emph{Powerformer} with \fpl{} and PatchTST. The colored lines represent PatchTST and different \fpl{} decay constants. For these forecasts, the sequence length is 512 and the prediction length is 336.}
\end{figure}

\begin{figure}[!hb]
\begin{minipage}{.495\textwidth}
    \includegraphics[width=1.0\textwidth]{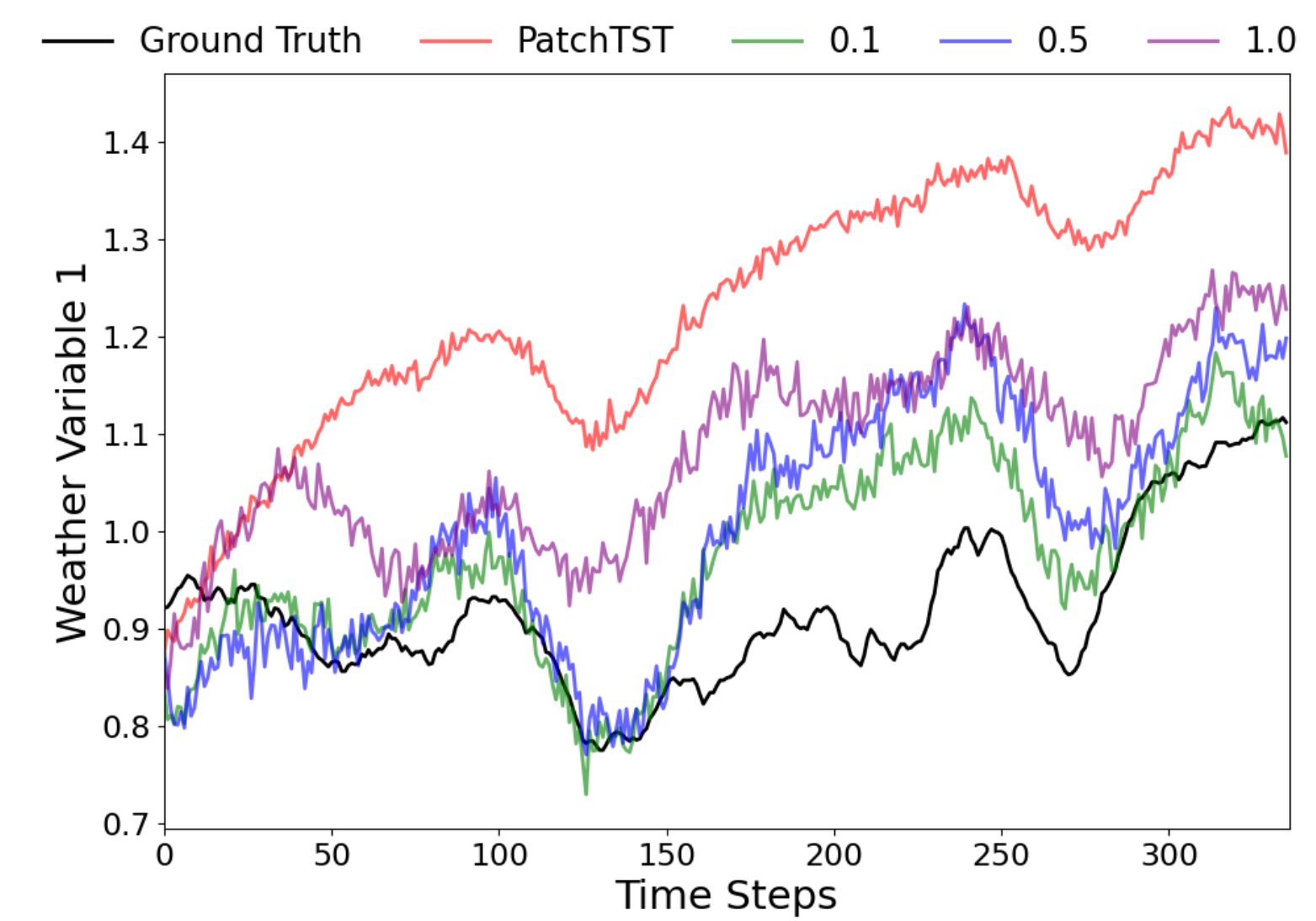}
\end{minipage}
\hfill    
\begin{minipage}{.495\textwidth}
    \includegraphics[width=1.0\textwidth]{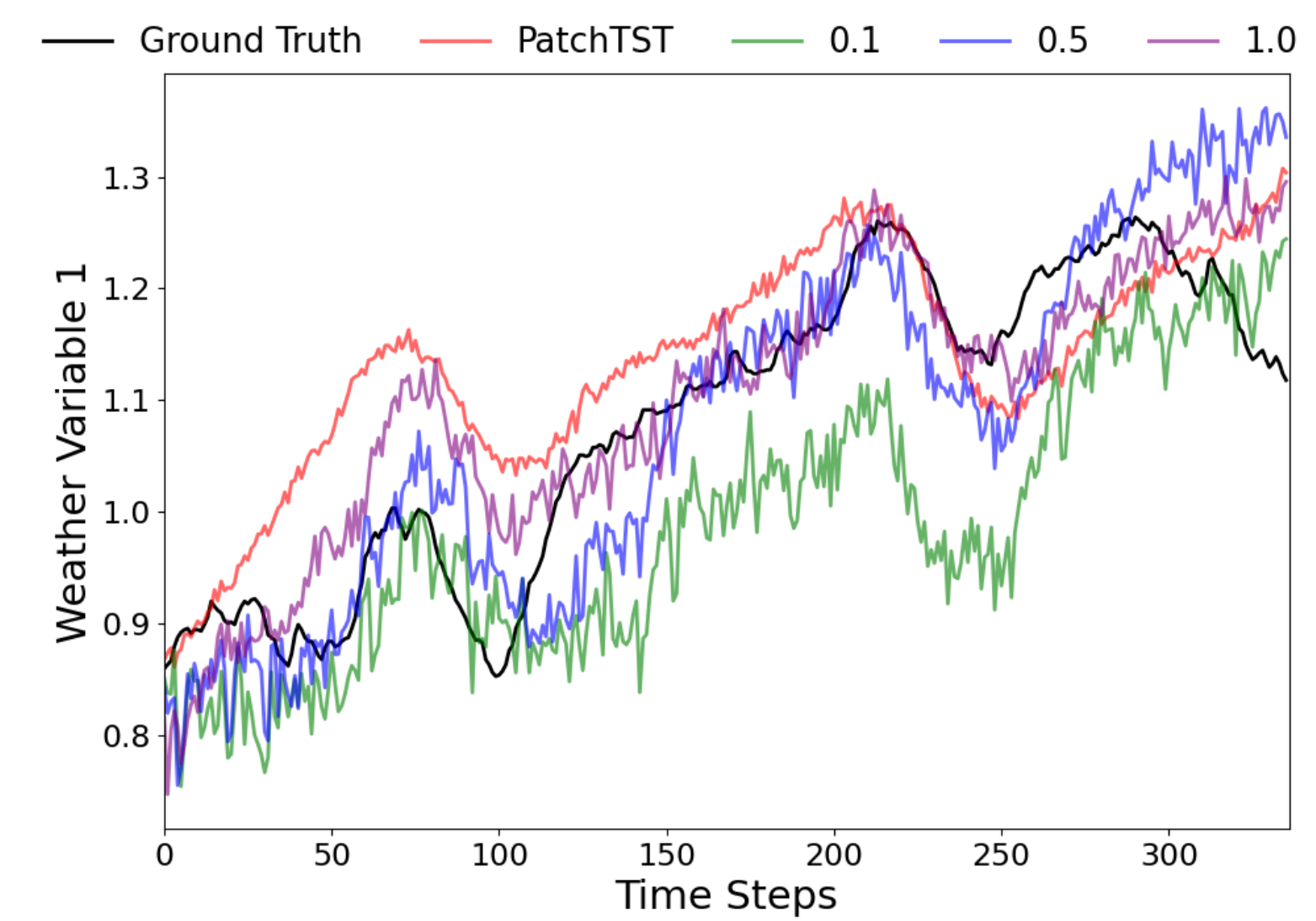}
\end{minipage}
\begin{minipage}{.495\textwidth}
    \includegraphics[width=1.0\textwidth]{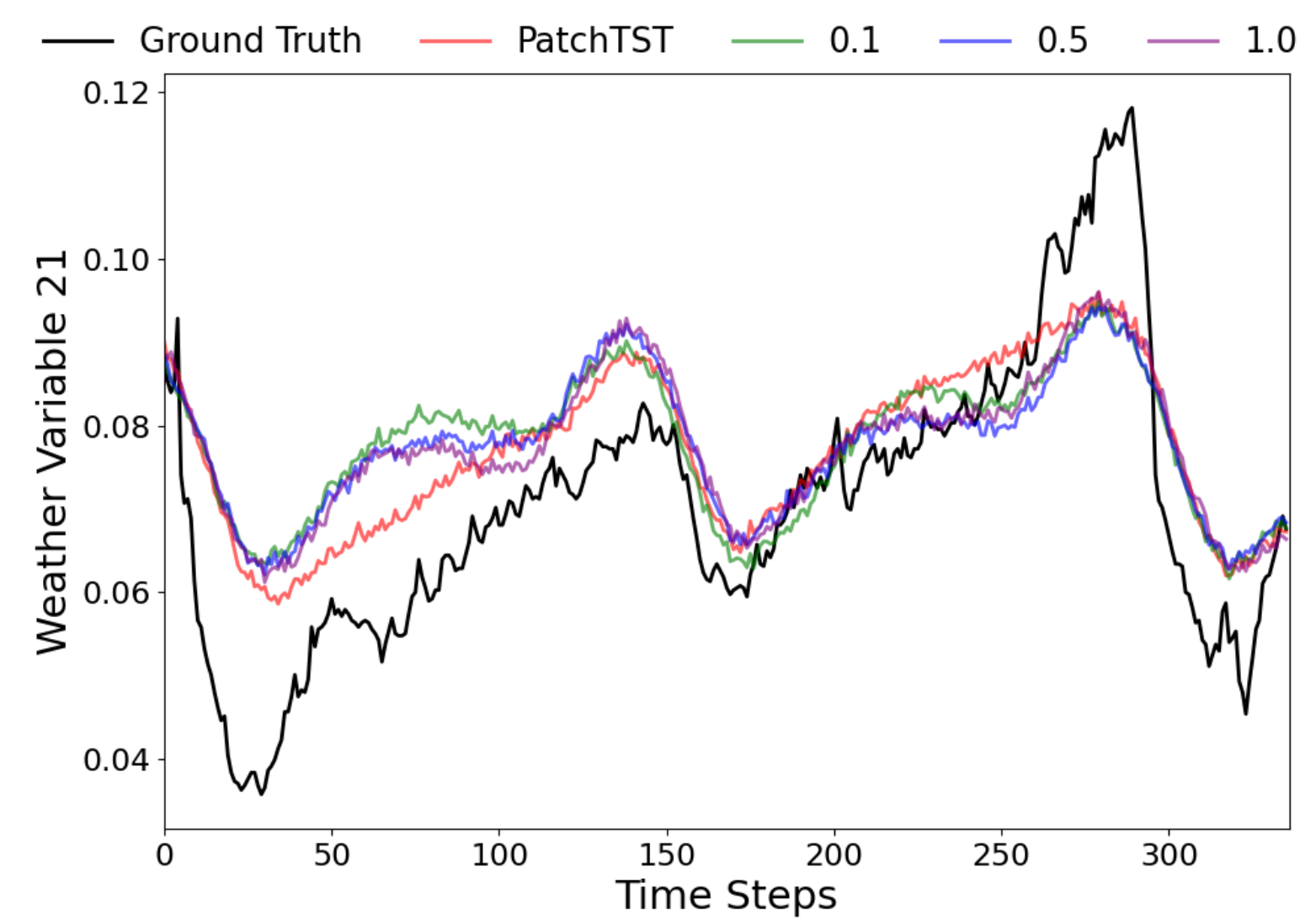}
\end{minipage}
\hfill    
\begin{minipage}{.495\textwidth}
    \includegraphics[width=1.0\textwidth]{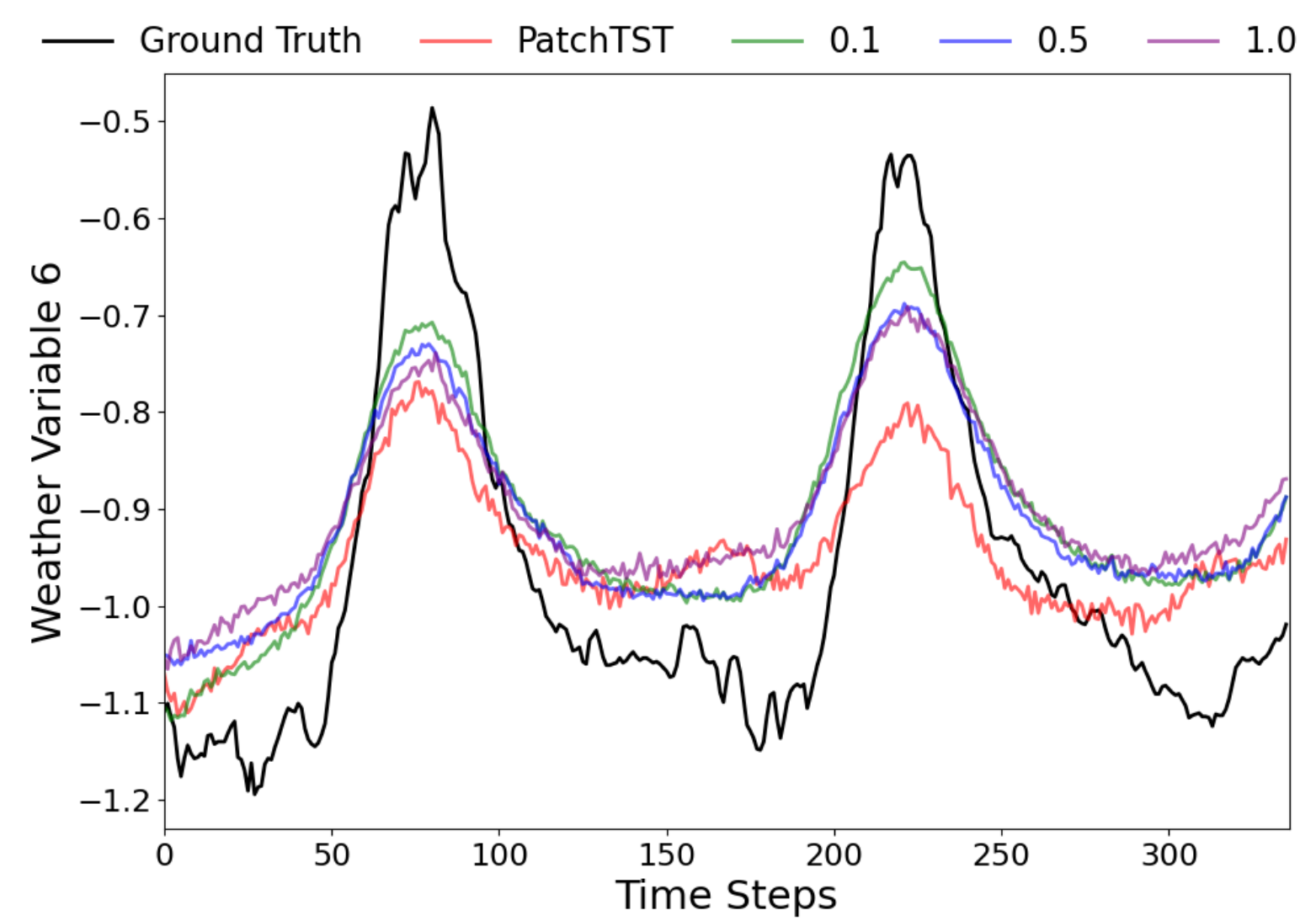}
\end{minipage}
\caption{We show forecasting results on the Weather dataset for \emph{Powerformer} with \fpl{} and PatchTST. The colored lines represent PatchTST and different \fpl{} decay constants. For these forecasts, the sequence length is 512 and the prediction length is 336.}
\end{figure}

\section{EXPERIMENTS}

\subsection{\label{sm:flipflop_exp}Flip Flop Experiment Details}

\begin{figure*}[bt]
    \centering
    \includegraphics[width=\linewidth]{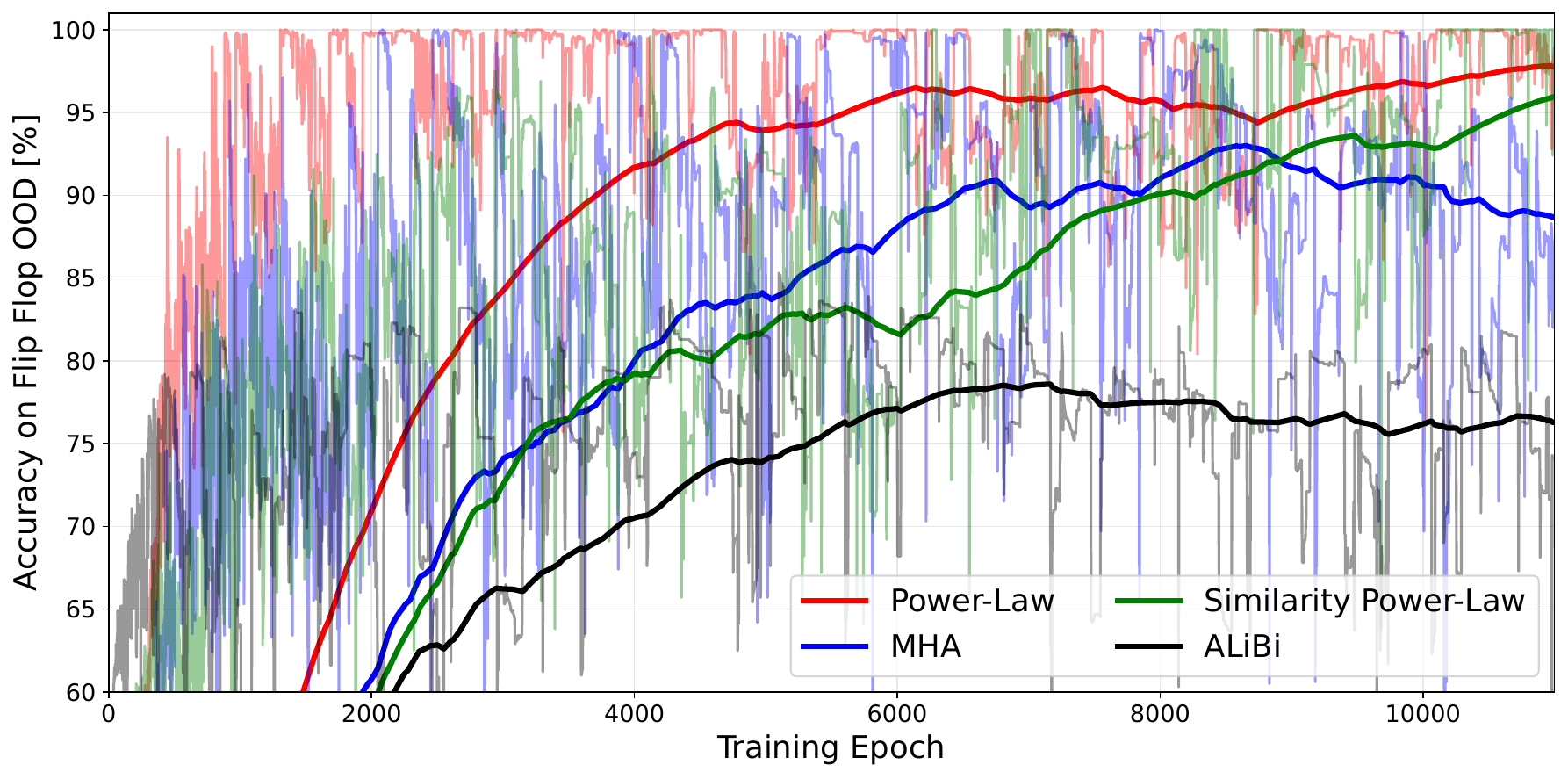}
    \caption{The flip-flip OOD evaluation for various recency biases and MHA. The transparent lines are the accuracies, while the solid lines are an exponential moving average.}
    \label{fig:flipflop_train}
\end{figure*}

We generate the flip-flop dataset by producing a string of 512 characters, alternating between instructional characters $c \in \{w, r, i\}$ and numbers $n \int \{0,0\}$~\citep{press2022train}.
The character commands dictate the model to do the following: write ($w$) the following number to memory (replacing the previous one), ignore ($i$) the following number and do not commit it to memory, and read ($r$) the number saved in memory.
The first character of the string is a $w$, the last character is $r$, and the character following $r$ is space holder `-' that is meant to be replaced by the predicted number.
During training, there can be multiple $r$ commands where the most recently written number is predicted.
We also generate and test the models on an out-of-distribution (OOD) dataset, where there are many ignore ($i$) commands between the last write and the last read commands.
When evaluating the OOD task, we only record results from the last read.

Due to the recurrent neural network (RNNs) architecture, these write-ignore-read capabilities are built in; however, this is not the case for Transformers~\citep{vaswani.transformer.2017}.
To test this, we use a 2-layer Transformer with embedding size 512, max sequence length of 512, absolute positional embeddings, and  4 attention heads.
In our experiments, we either varied or removed the causal mask.
We trained each model with a learning rate of $10^-5$ for 11000 epochs with a batch size of 32 and a weight decay scale of 0.1.

The OOD evaluation as a function of training, Fig.~\ref{fig:flipflop_train}, show that the power-law recency biases outperform.
In general, the power-law bias \fpl{} consistently outperforms regular MHA and other recency biases.
Interestingly, the ALiBi bias \fe{} does significantly worse than MHA even though it is introducing a recency bias.
The similarity power-law \fspl{} performs similarly to MHA, likely because this recency bias is relatively flat.

\subsection{Time-Series Experiment Details}
\label{sm:exp_details}

All experiments are evaluated 3 times with different initialization (\ref{sm:hyperparams}).
For each dataset, we select a single input sequence based on the best-performing input sequence length across all prediction lengths.
We choose a single input sequence length for each dataset because the strength of the correlations between time steps is independent of the prediction length, but the required input sequence length to capture the information necessary to forecast is unique for each dataset.
For each prediction length, we treat the attention mask as a hyperparameter and select the best mask and time constant pair.
We select attention masks for each prediction length because the forecasting timescale will likely emphasize short or long-term correlations for shorter or longer forecasting windows.
We make our selections based on the MSE error.
Section~\ref{sm:hyperparams} provides more details on the hyperparameter tuning and the hyperparameters used in this work.

\subsection{Powerformer}
\label{sm:experiments_powerformer}
We evaluate the weight power-law (\fpl) and similarity power-law (\fspl) filters on all 7 datasets.
For \fpl{} we evaluate $\alpha \in \{ 0.1, 0.25, 0.5, 0.75, 1.0 \}$ and for \fspl{} we evaluate $\alpha \in \{ 0.1, 0.5, 1.0, 2.0 \}$.
Similar to PatchTST~\citep{nie.patchtst.2023a} we evaluate the power-law filters with two input lengths: 336 and 512.

We evaluate both Butterworth filters $\left( \fbwomm \right.$ and $\left. \fbwtmm \right)$ on all the ETT datasets and the Weather datasets.
For each Butterworth filter, we evaluated $\alpha \in \{ 2, 5, 10, 15, 20 \}$ with an input length of 336.
Due to poor performance, we did not evaluate the Butterworth filters on the larger Electricity and Traffic datasets, nor with the longer 512 input length.

\subsection{Transformer}
\label{sm:experiments_transformer}
We train a standard encoder-decoder Transformer architecture both with and without RBCA.
Generic Transformer models often incorporate causality through masking.
To account for this, our Transformer benchmark uses masking for the decoder MHA but not for the encoder MHA nor the decoder cross-attention.
We investigate RBCA's effects by replacing the encoder and decoder self-attention, but not the decoder cross-attention as it is already causal.
Section~\ref{sm:transformer_architecture} provides more architectural details.

We train Transformer models, with and without RBCA, using the weight power-law (\fpl) filter on all 7 datasets.
We explore the following decay constants: $\alpha \in [0.1, 0.25, 0.5, 0.75, 1.0]$ and provide further experimental detail in sections~\ref{sec:experiments} and \ref{sm:exp_details}.

\subsection{Baseline Selection}
We compare \emph{Powerformer} against multiple Transformer-based state-of-the-art methods.
We populate Table~\ref{tab:performance} with values taken from the models' respective papers.
All models except TOTEM~\citep{talukder2024totem} and iTransformer~\citep{liu2024itransformer} treat the input length as a hyperparameter and report the best results.
PatchTST~\citep{nie.patchtst.2023a} provides results for two input sequence lengths (336 and 512).
In Table~\ref{tab:performance} we report the best result between these input sequence lengths for each dataset based on the MSE.
We employ the same input sequence length selection for \emph{Powerformer}.

\subsection{Computing Resources}
\label{sm:compute}
These experiments were run on NVIDIA A40 and NVIDIA TITAN RTX GPUs.
Compute time and resources are highly variable between datasets; the smallest (ETTh1) takes roughly 20 minutes on a single A40, while the largest (Traffic) takes roughly 15 hours on 4 A40s.

\section{TRANSFORMER RESULTS}
\label{sm:experiments_transformer_results}

Here, we present the full results from evaluating Transformer with RBCA.
The architecture is described in Appendix~\ref{sm:transformer_architecture} and the experimental details are provided in Appendices~\ref{sm:exp_details} and \ref{sm:experiments_transformer}.

Table~\ref{tab:transformer_powerLaw_full} provides the aggregate Transformer performance without RBCA and with RBCA for varying decay times ($\alpha$).
RBCA decisively outperforms MHA on 4 or the 7 datasets and ties with the number of best results on ETTh2.
Looking further into ETTh2 we see that when RBCA outperforms MHA it does so by much larger margins (16\%) than when it underperforms (2\%).
Moreover, we observe the same larger margins of overperformance across all of the datasets.
This indicates that when the natural pairwise distribution is found we can see significant improvements in MSE and MAE, but the effects of imposing the wrong pairwise distribution are much less significant.

\begin{figure*}[!ht]
    \centering
    \includegraphics[width=0.98\textwidth]{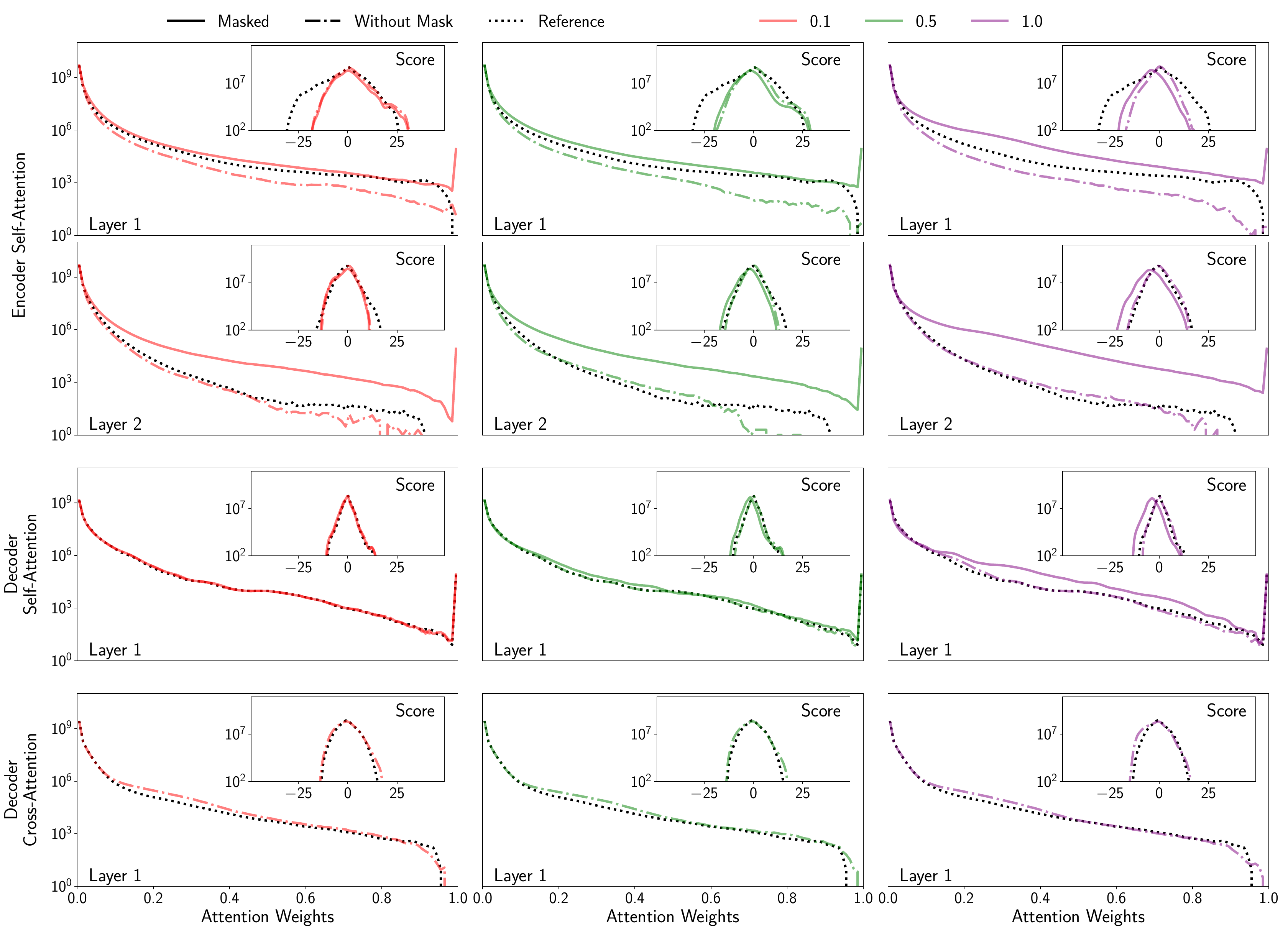}
    \caption{We present the Transformer attention score (inset) and weight distributions on the Weather dataset with a forecasting length of 96. The dotted line represents the reference MHA Transformer results, the dashed-dotted line represents RBCA with \fpl{} results before applying \maskCL, and the solid lines represent RBCA with \fpl{} results after applying \maskCL.}
    \label{fig:transformer_weather_96_attn_full}
\end{figure*}

\begin{figure*}[!ht]
    \centering
    \includegraphics[width=0.98\textwidth]{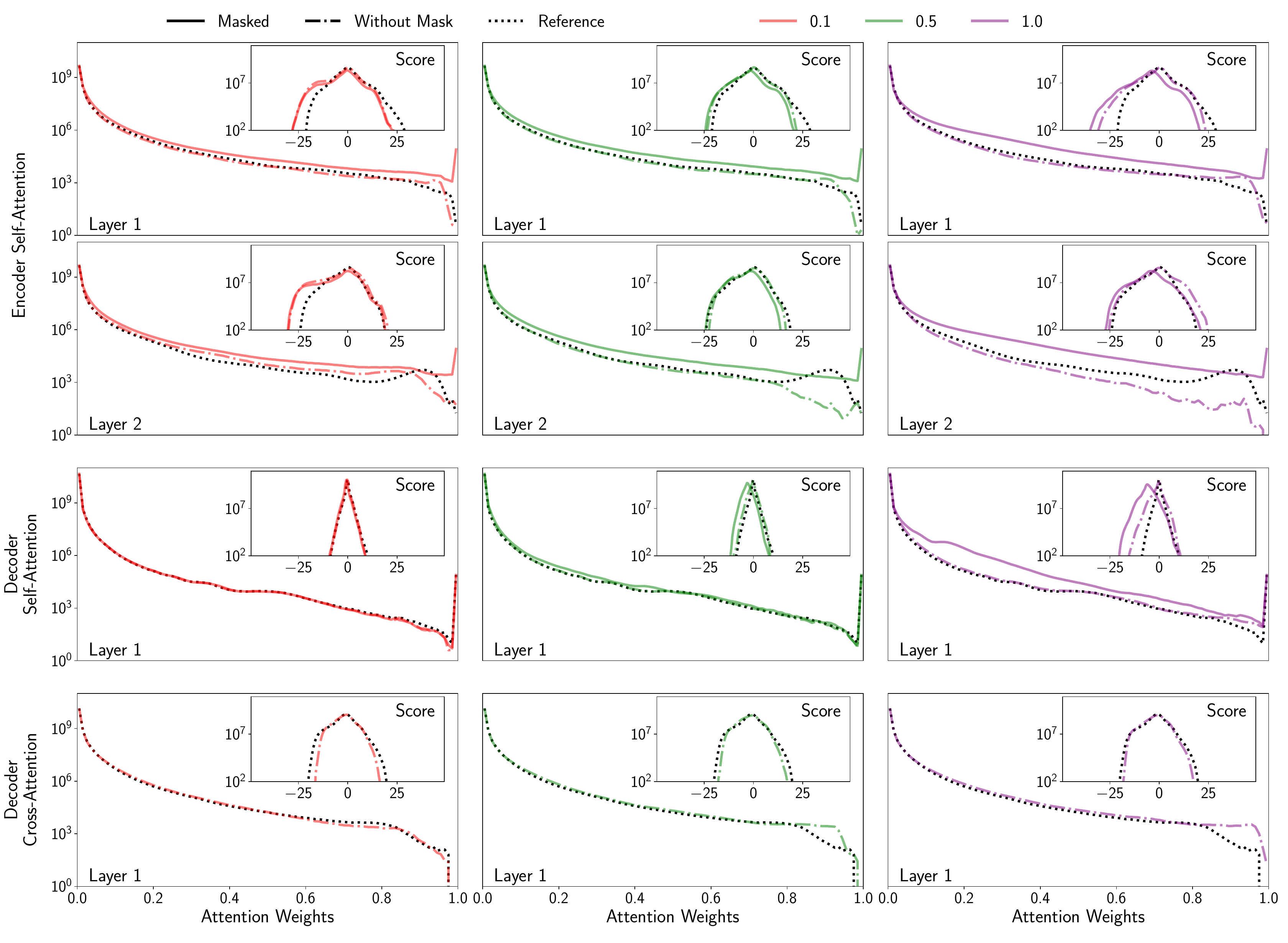}
    \caption{We present the Transformer attention score (inset) and weight distributions on the Weather dataset with a forecasting length of 720. The dotted line represents the reference MHA Transformer results, the dashed-dotted line represents RBCA with \fpl{} results before applying \maskCL, and the solid lines represent RBCA with \fpl{} results after applying \maskCL.}
    \label{fig:transformer_weather_720_attn_full}
\end{figure*}

\begin{figure*}[!ht]
    \centering
    \includegraphics[width=0.98\textwidth]{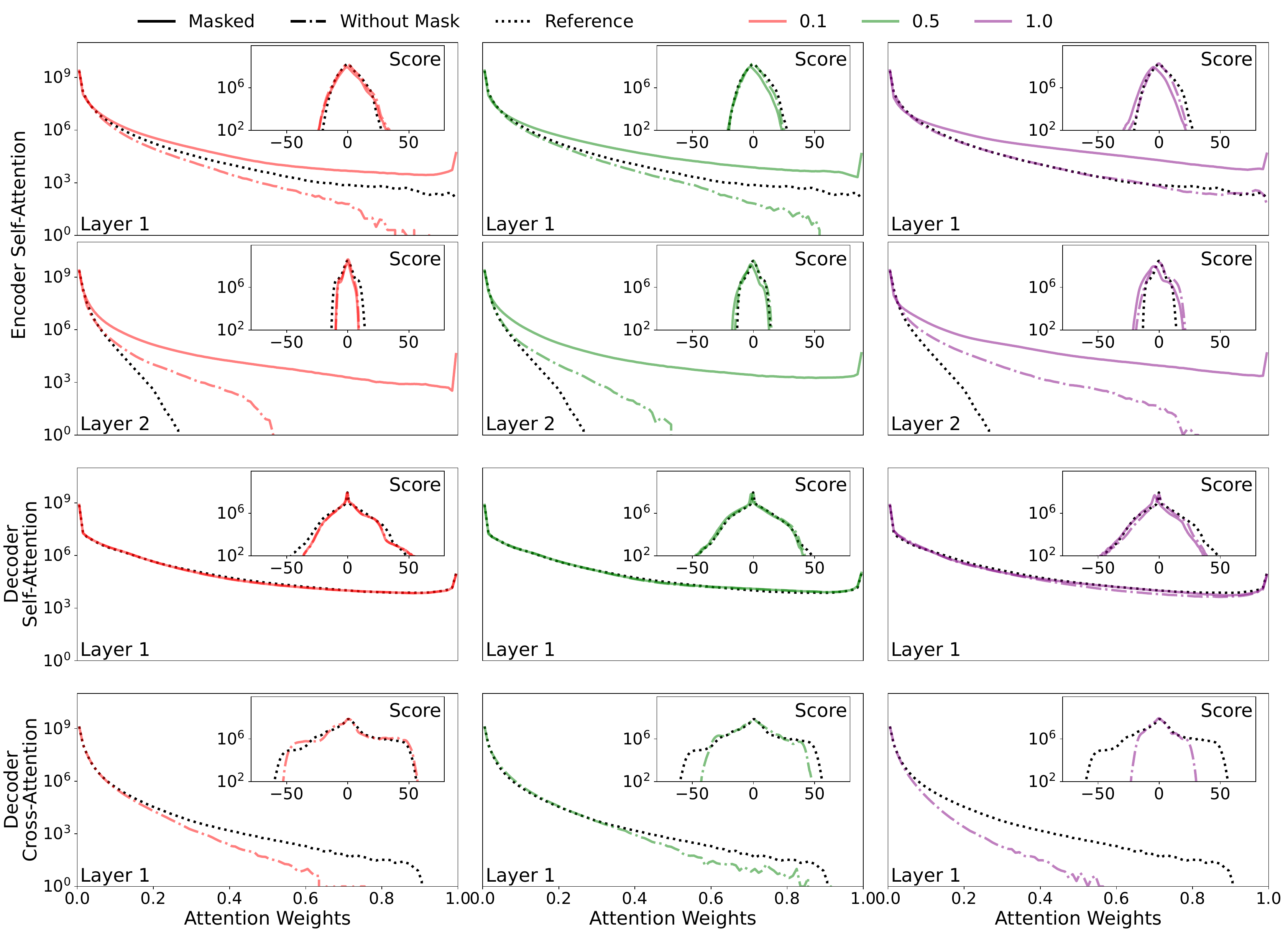}
    \caption{We present the Transformer attention score (inset) and weight distributions on the Electricity dataset with a forecasting length of 96. The dotted line represents the reference MHA Transformer results, the dashed-dotted line represents RBCA with \fpl{} results before applying \maskCL, and the solid lines represent RBCA with \fpl{} results after applying \maskCL.}
    \label{fig:transformer_electricity_96_attn_full}
\end{figure*}

\begin{figure*}[!ht]
    \centering
    \includegraphics[width=0.98\textwidth]{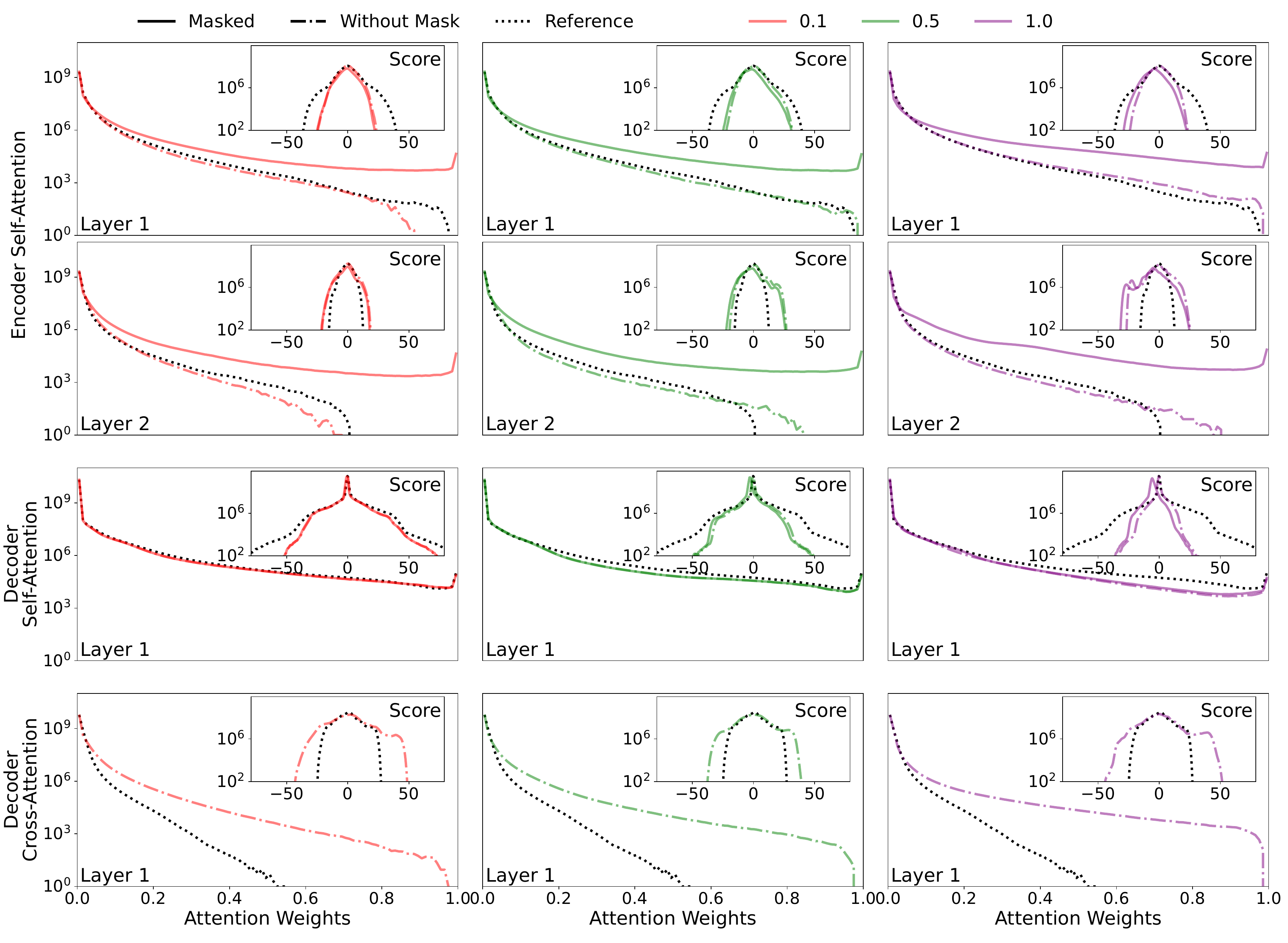}
    \caption{We present the Transformer attention score (inset) and weight distributions on the Electricity dataset with a forecasting length of 720. The dotted line represents the reference MHA Transformer results, the dashed-dotted line represents RBCA with \fpl{} results before applying \maskCL, and the solid lines represent RBCA with \fpl{} results after applying \maskCL.}
    \label{fig:transformer_electricity_720_attn_full}
\end{figure*}

In Figs.~\ref{fig:transformer_weather_96_attn_full}-\ref{fig:transformer_electricity_720_attn_full} we present the attention score and weight distributions with and without applying the local-causal mask.
The distributions can have significant deviations between datasets.
In general, the RBCA decoder self-attention shows the smallest deviation from the MHA base case.
We note that the MHA base case already has a causal mask.
The last RBCA encoder self-attention layer often shows the largest difference between RBCA and MHA distributions.
This makes sense since the MHA encoder attention is not causal.
We do not alter the decoder cross-attention, but we still observe shifts in the distribution, which differ between datasets.
Interestingly, we do not observe the same bimodal distribution in attention weights as we do for \emph{Powerformer}.
This may be due to the Transformer's encoder-decoder structure, whereas \emph{Powerformer} only has an encoder.

\begin{table*}[!ht]
	\centering
    \caption{We provide the forecasting MSE and MAE on the test sets for Transformer with MHA (base case) and for Transformer with RBCA and the weight power-law mask \fpl. The top row indicates the model or decay constant $(\alpha)$, the bold numbers are the best (lowest) performance, and the underlined numbers are the second best.}
    \label{tab:transformer_powerLaw_full}
    \vskip 0.1in
    \resizebox{\textwidth}{!}{
	\begin{tabular}{c|c|cc|cc|cc|cc|cc|cc}
	\toprule
	\multicolumn{2}{c|}{}  & \multicolumn{2}{c|}{Base Case} & \multicolumn{2}{c|}{0.1} & \multicolumn{2}{c|}{0.25} & \multicolumn{2}{c|}{0.5} & \multicolumn{2}{c|}{0.75} & \multicolumn{2}{c}{1} \\
	\midrule
	\multicolumn{2}{c|}{Metric} & MSE & MAE & MSE & MAE & MSE & MAE & MSE & MAE & MSE & MAE & MSE & MAE \\
	\midrule
	\multirow{4}{*}{\rotatebox[origin=c]{90}{\text{ETTh1}}}
		& 96 &  0.936 & 0.741 &  0.955 & 0.752 &  0.944 & 0.745 &  0.921 & 0.729 &  \underline{0.896} & \underline{0.711} &  \textbf{0.874} & \textbf{0.693}  \\
		& 192 &  0.988 & 0.769 &  0.916 & 0.759 &  0.951 & 0.789 &  \underline{0.887} & \underline{0.749} &  0.895 & \textbf{0.745} &  \textbf{0.876} & \underline{0.749}  \\
		& 336 &  1.061 & 0.815 &  1.067 & 0.835 &  1.063 & 0.814 &  1.054 & \underline{0.809} &  \textbf{1.038} & 0.836 &  \underline{1.044} & \textbf{0.802}  \\
		& 720 &  1.113 & 0.881 &  \textbf{1.047} & \textbf{0.842} &  \underline{1.061} & \underline{0.855} &  1.180 & 0.893 &  1.185 & 0.895 &  1.259 & 0.899  \\
	\midrule
	\multirow{4}{*}{\rotatebox[origin=c]{90}{\text{ETTh2}}}
		& 96 &  1.585 & 1.038 &  \textbf{1.344} & \textbf{0.918} &  \underline{1.349} & \underline{0.920} &  1.392 & 0.936 &  1.496 & 1.000 &  1.511 & 1.005  \\
		& 192 &  \textbf{1.889} & \textbf{1.131} &  \underline{1.899} & \underline{1.133} &  3.025 & 1.459 &  2.916 & 1.423 &  2.800 & 1.384 &  2.761 & 1.365  \\
		& 336 &  \textbf{2.668} & \textbf{1.384} &  2.875 & 1.393 &  2.872 & \underline{1.386} &  2.879 & 1.412 &  \underline{2.860} & 1.409 &  2.951 & 1.442  \\
		& 720 &  3.143 & 1.528 &  3.389 & 1.570 &  3.455 & 1.582 &  3.533 & 1.621 &  \underline{2.635} & \underline{1.331} &  \textbf{2.488} & \textbf{1.272}  \\
	\midrule
	\multirow{4}{*}{\rotatebox[origin=c]{90}{\text{ETTm1}}}
		& 96 &  0.560 & 0.545 &  0.566 & 0.533 &  0.592 & 0.545 &  0.563 & 0.541 &  \underline{0.535} & \underline{0.522} &  \textbf{0.487} & \textbf{0.492}  \\
		& 192 &  0.874 & 0.729 &  0.907 & 0.724 &  0.838 & 0.711 &  0.830 & 0.698 &  \underline{0.774} & \underline{0.686} &  \textbf{0.718} & \textbf{0.645}  \\
		& 336 &  0.960 & 0.786 &  0.948 & 0.786 &  \underline{0.938} & \underline{0.779} &  0.956 & 0.786 &  1.029 & 0.800 &  \textbf{0.888} & \textbf{0.752}  \\
		& 720 &  0.900 & 0.746 &  \textbf{0.864} & \textbf{0.710} &  \underline{0.886} & \underline{0.729} &  0.902 & 0.749 &  0.892 & \underline{0.729} &  1.005 & 0.788  \\
	\midrule
	\multirow{4}{*}{\rotatebox[origin=c]{90}{\text{ETTm2}}}
		& 96 &  \textbf{0.597} & \textbf{0.614} &  \underline{0.619} & \underline{0.628} &  0.632 & 0.636 &  0.660 & 0.653 &  0.697 & 0.674 &  0.700 & 0.638  \\
		& 192 &  \textbf{0.849} & \textbf{0.749} &  \underline{0.876} & \underline{0.761} &  0.893 & 0.769 &  0.927 & 0.787 &  0.970 & 0.808 &  1.011 & 0.827  \\
		& 336 &  \textbf{1.227} & \textbf{0.909} &  \underline{1.259} & \underline{0.920} &  1.271 & 0.925 &  1.295 & 0.936 &  1.320 & 0.947 &  1.333 & 0.955  \\
		& 720 &  \textbf{2.176} & \textbf{1.237} &  2.214 & \underline{1.241} &  2.213 & \underline{1.241} &  2.216 & 1.252 &  2.209 & 1.252 &  \underline{2.195} & 1.250  \\
	\midrule
	\multirow{4}{*}{\rotatebox[origin=c]{90}{\text{Weather}}}
		& 96 &  0.250 & 0.333 &  0.211 & \underline{0.297} &  \underline{0.209} & 0.302 &  \textbf{0.207} & \textbf{0.291} &  0.216 & 0.305 &  0.228 & 0.316  \\
		& 192 &  0.344 & 0.414 &  \underline{0.295} & \underline{0.373} &  \textbf{0.287} & \textbf{0.367} &  0.297 & 0.375 &  0.307 & 0.383 &  0.313 & 0.390  \\
		& 336 &  0.435 & 0.467 &  \underline{0.403} & \underline{0.446} &  0.416 & 0.455 &  0.420 & 0.458 &  0.410 & 0.452 &  \textbf{0.384} & \textbf{0.436}  \\
		& 720 &  0.463 & 0.489 &  \underline{0.420} & \underline{0.460} &  0.435 & 0.471 &  0.424 & 0.464 &  0.426 & 0.468 &  \textbf{0.403} & \textbf{0.454}  \\
	\midrule
	\multirow{4}{*}{\rotatebox[origin=c]{90}{\text{Electricity}}}
		& 96 &  \underline{0.265} & 0.360 &  0.271 & 0.364 &  \textbf{0.264} & \textbf{0.355} &  0.266 & \underline{0.359} &  0.272 & 0.367 &  0.280 & 0.370  \\
		& 192 &  0.284 & 0.378 &  0.278 & 0.372 &  0.286 & 0.378 &  \textbf{0.270} & \textbf{0.366} &  \underline{0.276} & \underline{0.369} &  0.291 & 0.386  \\
		& 336 &  \underline{0.280} & 0.371 &  0.289 & 0.381 &  0.290 & 0.381 &  0.288 & 0.381 &  0.281 & \underline{0.369} &  \textbf{0.277} & \textbf{0.366}  \\
		& 720 &  0.300 & 0.382 &  0.318 & 0.395 &  0.308 & 0.389 &  \underline{0.296} & \underline{0.378} &  0.302 & 0.383 &  \textbf{0.289} & \textbf{0.373}  \\
	\midrule
	\multirow{4}{*}{\rotatebox[origin=c]{90}{\text{Traffic}}}
		& 96 &  0.662 & 0.365 &  0.656 & 0.359 &  0.653 & \underline{0.357} &  \underline{0.652} & \textbf{0.355} &  0.669 & 0.364 &  \textbf{0.649} & 0.358  \\
		& 192 &  \textbf{0.655} & \textbf{0.350} &  0.675 & 0.368 &  0.669 & 0.361 &  0.669 & 0.365 &  \underline{0.660} & 0.356 &  \textbf{0.655} & \underline{0.355}  \\
		& 336 &  \textbf{0.651} & \textbf{0.350} &  \underline{0.652} & \underline{0.353} &  0.657 & 0.357 &  0.670 & 0.364 &  0.667 & 0.361 &  0.658 & 0.358  \\
		& 720 &  \textbf{0.660} & \textbf{0.354} &  0.669 & 0.363 &  0.676 & 0.360 &  \underline{0.667} & \underline{0.355} &  0.668 & 0.361 &  0.689 & 0.370  \\
	\bottomrule
	\end{tabular}}
\end{table*}

\FloatBarrier

\section{POWERFORMER RESULTS}
\label{sm:experiments_powerformer_results}

We present the full \emph{Powerformer} results on all the evaluated masks: \fpl, \fspl, \fbwo, and \fbwt.
Architecture details can be found in Appendix~\ref{sm:powerformer_architecture}, while Appendices~\ref{sm:exp_details} and \ref{sm:experiments_powerformer} outline the experimental details.

\subsection{Weight Power-Law Mask}
\label{sm:experiments_powerformer_simpowerLaw}
Table~\ref{tab:powerformer_powerLaw_results} presents the aggregate MSE and MAE results for weight power-law mask (\fpl) with varying time decays ($\alpha$), forecasting lengths, and input sequence lengths.
The 512 input sequence generally outperforms the shorter 336 input length.
As expected, the best performing $\alpha$ varies between datasets, but it also varies as a function of forecast length and input sequence length.
Given that some of the best-performing decay lengths are $\alpha=1$ with further experiments one may improve upon these values.
However, our \fspl{} mask includes faster-decaying weights, as shown in Fig.~\ref{fig:maskPL}.

\begin{figure*}[!htb]
    \centering
    \includegraphics[width=0.92\textwidth]{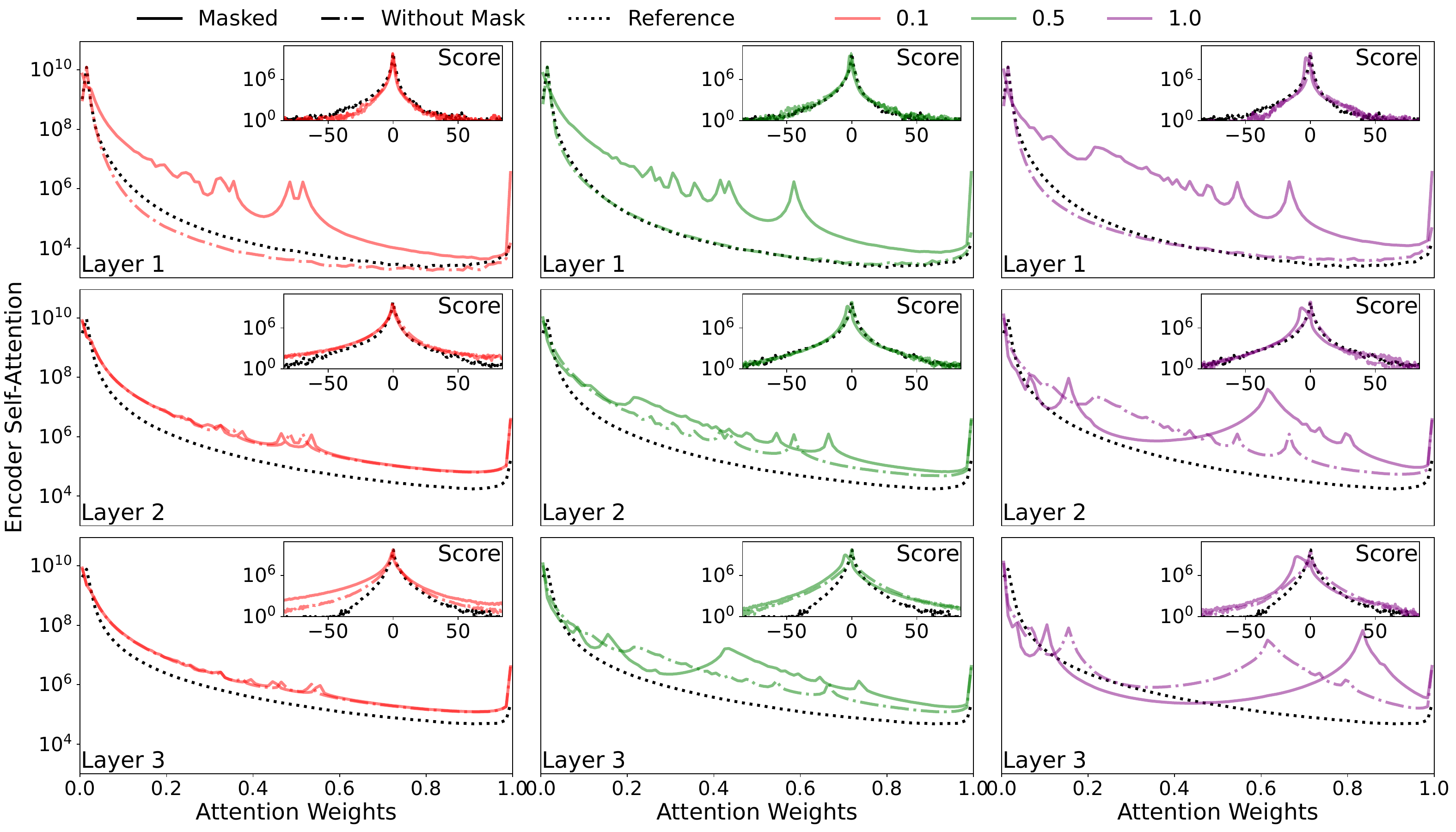}
    \caption{We present the \emph{Powerformer} attention score (inset) and weight distributions on the Weather dataset with a forecast and input length of 96 and 512, respectively. The dotted line represents the reference MHA results, the dashed-dotted line represents RBCA with \fpl{} results before applying \maskCL, and the solid lines represent RBCA with \fpl{} results after applying \maskCL.}
    \label{fig:powerformer_weather_96_powerLaw_attn_full}
\end{figure*}

\begin{figure*}[!htb]
    \centering
    \includegraphics[width=0.92\textwidth]{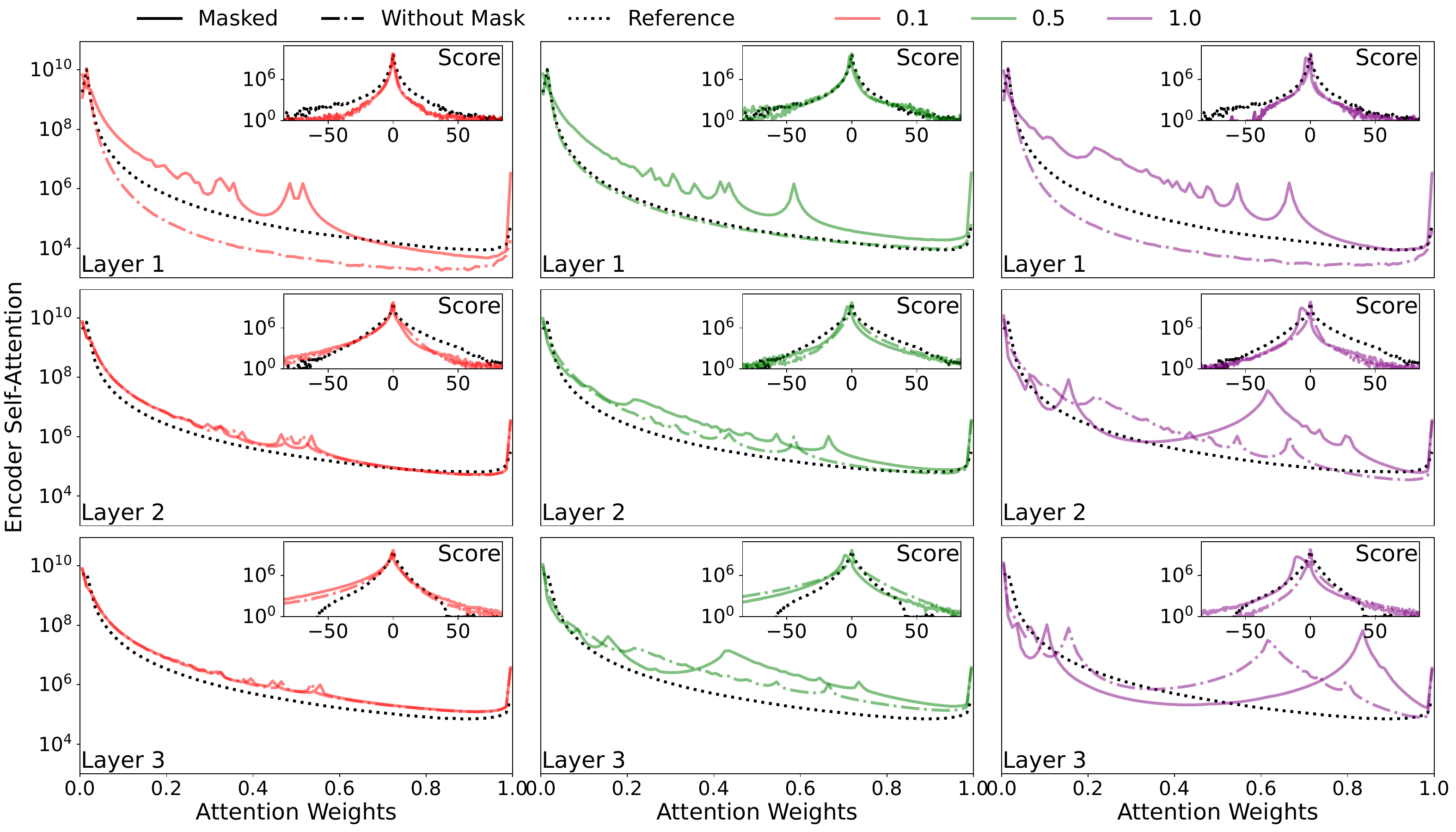}
    \caption{We present the \emph{Powerformer} attention score (inset) and weight distributions on the Weather dataset with a forecast and input length of 720 and 512, respectively. The dotted line represents the reference MHA results, the dashed-dotted line represents RBCA with \fpl{} results before applying \maskCL, and the solid lines represent RBCA with \fpl{} results after applying \maskCL.}
    \label{fig:powerformer_weather_720_powerLaw_attn_full}
\end{figure*}

\begin{figure*}[!htb]
    \centering
    \includegraphics[width=0.92\textwidth]{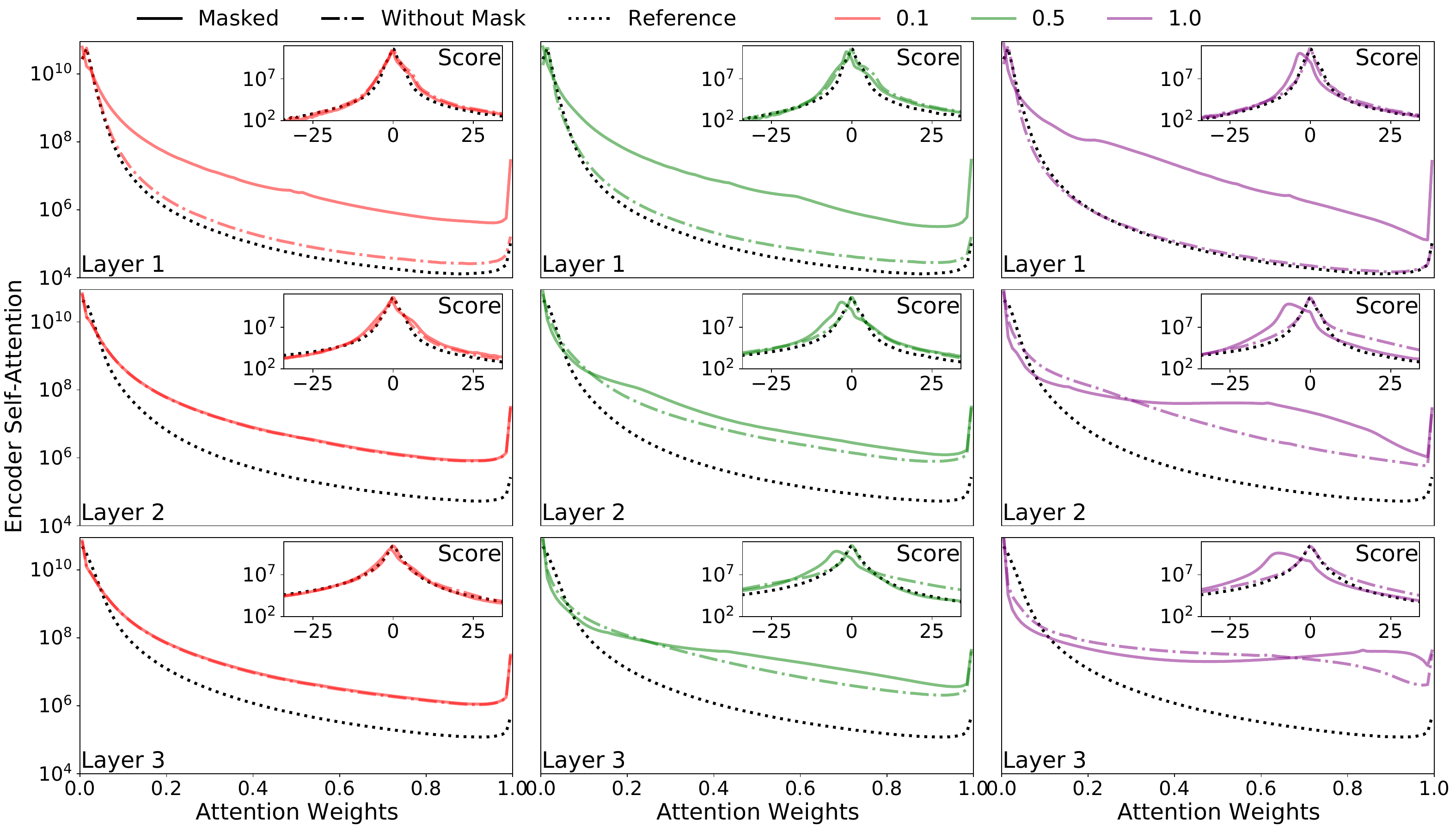}
    \caption{We present the \emph{Powerformer} attention score (inset) and weight distributions on the Electricity dataset with a forecast and input length of 96 and 512, respectively. The dotted line represents the reference MHA results, the dashed-dotted line represents RBCA with \fpl{} results before applying \maskCL, and the solid lines represent RBCA with \fpl{} results after applying \maskCL.}
    \label{fig:powerformer_electricity_96_powerLaw_attn_full}
\end{figure*}

\begin{figure*}[!htb]
    \centering
    \includegraphics[width=0.92\textwidth]{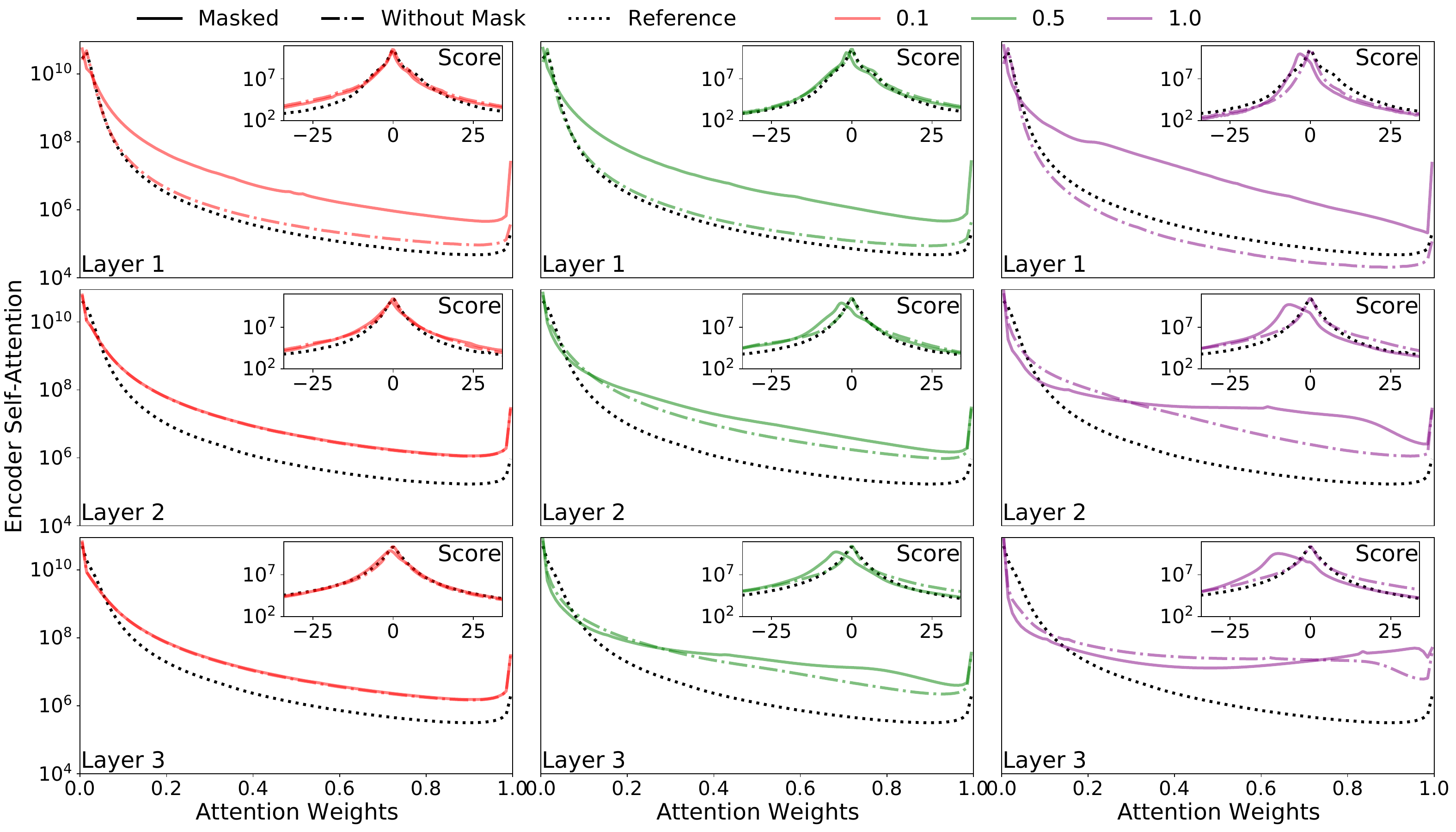}
    \caption{We present the \emph{Powerformer} attention score (inset) and weight distributions on the Electricity dataset with a forecast and input length of 720 and 512, respectively. The dotted line represents the reference MHA results, the dashed-dotted line represents RBCA with \fpl{} results before applying \maskCL, and the solid lines represent RBCA with \fpl{} results after applying \maskCL.}
    \label{fig:powerformer_electricity_720_powerLaw_attn_full}
\end{figure*}

We compare all datasets' attention score and weight distributions both with MHA and RBCA in Figs~\ref{fig:powerformer_weather_96_powerLaw_attn_full}-\ref{fig:powerformer_electricity_720_powerLaw_attn_full}.
We observe similar bimodal distributions for all datasets and note wider \weightCL{} distributions for ETTm1, ETTm2, and Weather.

\begingroup
\setlength{\tabcolsep}{7pt} 
\renewcommand{\arraystretch}{0.52} 
\begin{table*}[!ht]
    \centering
    \caption{We compare \emph{Powerformer's} performance with the weight power law mask \fpl{} on standard time-series datasets for varying decay lengths. The best results are bolded and second best are underlined.}
    \label{tab:powerformer_powerLaw_results}
    \vskip 0.1in
    \resizebox{\textwidth}{!}{
    \begin{tabular}{c|c|c|cc|cc|cc|cc|cc}
    	\toprule
    	\multicolumn{3}{c|}{Delay}  & \multicolumn{2}{c|}{0.1} & \multicolumn{2}{c|}{0.25} & \multicolumn{2}{c|}{0.5} & \multicolumn{2}{c|}{0.75} & \multicolumn{2}{c}{1} \\
    	\midrule
    	\multicolumn{3}{c|}{Metric} & MSE & MAE & MSE & MAE & MSE & MAE & MSE & MAE & MSE & MAE \\
    	\midrule
    	\multirow{8}{*}{\rotatebox[origin=c]{90}{\text{ETTh1}}}
    	& \multirow{4}{*}{\rotatebox[origin=c]{90}{\text{336}}}
    		 & 96 &  \underline{0.377} & \textbf{0.401} &  \underline{0.377} & \textbf{0.401} &  \underline{0.377} & \textbf{0.401} &  \textbf{0.376} & \textbf{0.401} &  \textbf{0.376} & \textbf{0.401} \\
    		 & & 192 &  \textbf{0.413} & \textbf{0.421} &  \textbf{0.413} & \textbf{0.421} &  \textbf{0.413} & \textbf{0.421} &  \textbf{0.413} & \textbf{0.421} &  \underline{0.414} & \textbf{0.421} \\
    		 & & 336 &  \textbf{0.424} & \textbf{0.430} &  \underline{0.425} & \textbf{0.430} &  \underline{0.425} & \textbf{0.430} &  \underline{0.425} & \textbf{0.430} &  \underline{0.425} & \textbf{0.430} \\
    		 & & 720 &  \textbf{0.437} & \textbf{0.455} &  \textbf{0.437} & \textbf{0.455} &  \textbf{0.437} & \textbf{0.455} &  \textbf{0.437} & \underline{0.456} &  \underline{0.438} & \underline{0.456} \\ \cmidrule{2-13}
    	& \multirow{4}{*}{\rotatebox[origin=c]{90}{\text{512}}}
    		 & 96 &  \textbf{0.369} & \textbf{0.399} &  \textbf{0.369} & \textbf{0.399} &  \underline{0.370} & \textbf{0.399} &  \underline{0.370} & \textbf{0.399} &  \underline{0.370} & \textbf{0.399} \\
    		 & & 192 &  0.404 & \underline{0.420} &  0.404 & \underline{0.420} &  \textbf{0.402} & \textbf{0.418} &  \underline{0.403} & \textbf{0.418} &  \underline{0.403} & \textbf{0.418} \\
    		 & & 336 &  \textbf{0.414} & \textbf{0.428} &  \textbf{0.414} & \textbf{0.428} &  \underline{0.415} & \underline{0.429} &  \underline{0.415} & 0.430 &  0.416 & 0.431 \\
    		 & & 720 &  \textbf{0.440} & \textbf{0.460} &  \textbf{0.440} & \textbf{0.460} &  \textbf{0.440} & \textbf{0.460} &  \underline{0.443} & \underline{0.462} &  \underline{0.443} & \textbf{0.460} \\
    	\midrule
    	\multirow{8}{*}{\rotatebox[origin=c]{90}{\text{ETTh2}}}
    	& \multirow{4}{*}{\rotatebox[origin=c]{90}{\text{336}}}
    		 & 96 &  \textbf{0.274} & \textbf{0.335} &  \textbf{0.274} & \underline{0.336} &  \underline{0.275} & \underline{0.336} &  \underline{0.275} & \textbf{0.335} &  \underline{0.275} & \textbf{0.335} \\
    		 & & 192 &  0.341 & \underline{0.381} &  \textbf{0.338} & \textbf{0.379} &  \underline{0.340} & \textbf{0.379} &  \underline{0.340} & \textbf{0.379} &  0.341 & \textbf{0.379} \\
    		 & & 336 &  \textbf{0.325} & \textbf{0.379} &  \underline{0.326} & \underline{0.380} &  0.331 & 0.384 &  0.333 & 0.383 &  0.334 & 0.383 \\
    		 & & 720 &  \underline{0.377} & \underline{0.420} &  \textbf{0.376} & \textbf{0.419} &  0.381 & 0.423 &  0.381 & 0.422 &  0.381 & 0.422 \\ \cmidrule{2-13}
    	& \multirow{4}{*}{\rotatebox[origin=c]{90}{\text{512}}}
    		 & 96 &  \textbf{0.274} & \textbf{0.337} &  \textbf{0.274} & \textbf{0.337} &  \textbf{0.274} & \textbf{0.337} &  \underline{0.275} & \textbf{0.337} &  \underline{0.275} & \underline{0.338} \\
    		 & & 192 &  \textbf{0.340} & \textbf{0.381} &  \textbf{0.340} & \textbf{0.381} &  \underline{0.341} & \underline{0.382} &  0.342 & \underline{0.382} &  0.342 & \underline{0.382} \\
    		 & & 336 &  \textbf{0.330} & \textbf{0.387} &  \underline{0.331} & \textbf{0.387} &  0.333 & \underline{0.388} &  0.334 & \underline{0.388} &  0.334 & \underline{0.388} \\
    		 & & 720 &  \textbf{0.383} & \underline{0.426} &  \underline{0.384} & \underline{0.426} &  \textbf{0.383} & \underline{0.426} &  \underline{0.384} & \underline{0.426} &  \textbf{0.383} & \textbf{0.425} \\
    	\midrule
    	\multirow{8}{*}{\rotatebox[origin=c]{90}{\text{ETTm1}}}
    	& \multirow{4}{*}{\rotatebox[origin=c]{90}{\text{336}}}
    		 & 96 &  \underline{0.292} & \underline{0.343} &  \textbf{0.290} & \textbf{0.342} &  \underline{0.292} & \underline{0.343} &  0.293 & 0.344 &  \underline{0.292} & 0.345 \\
    		 & & 192 &  0.333 & 0.371 &  0.334 & 0.372 &  \underline{0.332} & 0.371 &  \underline{0.332} & \underline{0.370} &  \textbf{0.330} & \textbf{0.369} \\
    		 & & 336 &  0.362 & 0.392 &  \underline{0.361} & 0.391 &  0.362 & \underline{0.390} &  \textbf{0.359} & \textbf{0.389} &  0.362 & \textbf{0.389} \\
    		 & & 720 &  \underline{0.413} & 0.423 &  \underline{0.413} & \underline{0.422} &  \underline{0.413} & \textbf{0.421} &  \underline{0.413} & \textbf{0.421} &  \textbf{0.412} & 0.423 \\ \cmidrule{2-13}
    	& \multirow{4}{*}{\rotatebox[origin=c]{90}{\text{512}}}
    		 & 96 &  \underline{0.291} & \textbf{0.345} &  \textbf{0.290} & \textbf{0.345} &  \textbf{0.290} & \textbf{0.345} &  \underline{0.291} & \underline{0.346} &  0.292 & \underline{0.346} \\
    		 & & 192 &  0.331 & \underline{0.370} &  \underline{0.330} & \underline{0.370} &  \textbf{0.329} & \textbf{0.368} &  0.332 & \underline{0.370} &  0.332 & 0.372 \\
    		 & & 336 &  \underline{0.362} & \underline{0.391} &  \underline{0.362} & \textbf{0.390} &  \textbf{0.361} & 0.392 &  \underline{0.362} & \textbf{0.390} &  0.363 & \underline{0.391} \\
    		 & & 720 &  0.417 & 0.428 &  \textbf{0.415} & \underline{0.425} &  0.417 & \textbf{0.423} &  0.420 & 0.426 &  \underline{0.416} & 0.430 \\
    	\midrule
    	\multirow{8}{*}{\rotatebox[origin=c]{90}{\text{ETTm2}}}
    	& \multirow{4}{*}{\rotatebox[origin=c]{90}{\text{336}}}
    		 & 96 &  \textbf{0.162} & \textbf{0.252} &  0.165 & 0.255 &  \underline{0.163} & \underline{0.253} &  0.164 & 0.254 &  0.164 & 0.254 \\
    		 & & 192 &  \textbf{0.222} & \textbf{0.292} &  \textbf{0.222} & \underline{0.293} &  \textbf{0.222} & \underline{0.293} &  \textbf{0.222} & \underline{0.293} &  \textbf{0.222} & \underline{0.293} \\
    		 & & 336 &  \textbf{0.277} & \textbf{0.329} &  \textbf{0.277} & \textbf{0.329} &  \textbf{0.277} & \textbf{0.329} &  \textbf{0.277} & \textbf{0.329} &  \underline{0.278} & \underline{0.330} \\
    		 & & 720 &  0.363 & \underline{0.381} &  0.363 & \underline{0.381} &  \underline{0.362} & 0.382 &  \textbf{0.359} & \textbf{0.380} &  \textbf{0.359} & 0.384 \\ \cmidrule{2-13}
    	& \multirow{4}{*}{\rotatebox[origin=c]{90}{\text{512}}}
    		 & 96 &  \underline{0.165} & \textbf{0.255} &  \underline{0.165} & \textbf{0.255} &  \underline{0.165} & \underline{0.256} &  \underline{0.165} & 0.257 &  \textbf{0.164} & \textbf{0.255} \\
    		 & & 192 &  \underline{0.222} & \underline{0.295} &  0.224 & 0.297 &  0.224 & 0.296 &  0.223 & 0.296 &  \textbf{0.221} & \textbf{0.294} \\
    		 & & 336 &  0.274 & \underline{0.329} &  0.274 & \underline{0.329} &  0.274 & \underline{0.329} &  \textbf{0.272} & \textbf{0.327} &  \underline{0.273} & \textbf{0.327} \\
    		 & & 720 &  \underline{0.358} & 0.383 &  \underline{0.358} & 0.383 &  \underline{0.358} & \underline{0.382} &  \underline{0.358} & 0.383 &  \textbf{0.356} & \textbf{0.381} \\
    	\midrule
    	\multirow{8}{*}{\rotatebox[origin=c]{90}{\text{Weather}}}
    	& \multirow{4}{*}{\rotatebox[origin=c]{90}{\text{336}}}
    		 & 96 &  \textbf{0.151} & \underline{0.201} &  \textbf{0.151} & \underline{0.201} &  \textbf{0.151} & \textbf{0.200} &  \underline{0.152} & \textbf{0.200} &  0.154 & 0.202 \\
    		 & & 192 &  \textbf{0.196} & 0.243 &  \textbf{0.196} & \underline{0.242} &  \textbf{0.196} & \textbf{0.241} &  \textbf{0.196} & \textbf{0.241} &  \underline{0.197} & \underline{0.242} \\
    		 & & 336 &  \underline{0.248} & 0.283 &  \textbf{0.247} & 0.283 &  \textbf{0.247} & \textbf{0.281} &  \textbf{0.247} & \underline{0.282} &  \textbf{0.247} & \textbf{0.281} \\
    		 & & 720 &  \underline{0.318} & \underline{0.334} &  \underline{0.318} & \underline{0.334} &  \underline{0.318} & \underline{0.334} &  \textbf{0.317} & \textbf{0.333} &  \textbf{0.317} & \textbf{0.333} \\ \cmidrule{2-13}
    	& \multirow{4}{*}{\rotatebox[origin=c]{90}{\text{512}}}
    		 & 96 &  \underline{0.148} & \underline{0.198} &  \textbf{0.147} & \textbf{0.197} &  \textbf{0.147} & 0.199 &  \underline{0.148} & 0.199 &  \underline{0.148} & 0.199 \\
    		 & & 192 &  \underline{0.193} & \underline{0.241} &  \underline{0.193} & \underline{0.241} &  \underline{0.193} & \underline{0.241} &  \textbf{0.191} & \textbf{0.239} &  \underline{0.193} & \underline{0.241} \\
    		 & & 336 &  \textbf{0.243} & 0.281 &  \underline{0.244} & 0.281 &  \textbf{0.243} & \underline{0.280} &  \textbf{0.243} & \textbf{0.279} &  \underline{0.244} & \underline{0.280} \\
    		 & & 720 &  \textbf{0.311} & \underline{0.330} &  \textbf{0.311} & \underline{0.330} &  0.314 & 0.331 &  \underline{0.312} & \underline{0.330} &  \textbf{0.311} & \textbf{0.329} \\
    	\midrule
    	\multirow{8}{*}{\rotatebox[origin=c]{90}{\text{Electricity}}}
    	& \multirow{4}{*}{\rotatebox[origin=c]{90}{\text{336}}}
    		 & 96 &  \underline{0.131} & \textbf{0.224} &  \underline{0.131} & \textbf{0.224} &  \textbf{0.130} & \textbf{0.224} &  \underline{0.131} & \textbf{0.224} &  \underline{0.131} & \textbf{0.224} \\
    		 & & 192 &  \textbf{0.147} & \underline{0.241} &  \textbf{0.147} & \textbf{0.240} &  \textbf{0.147} & \textbf{0.240} &  \textbf{0.147} & \underline{0.241} &  \underline{0.148} & \underline{0.241} \\
    		 & & 336 &  \textbf{0.164} & 0.259 &  \textbf{0.164} & \textbf{0.257} &  \textbf{0.164} & 0.259 &  \textbf{0.164} & \underline{0.258} &  \textbf{0.164} & 0.259 \\
    		 & & 720 &  \textbf{0.201} & \textbf{0.291} &  \textbf{0.201} & \underline{0.292} &  \underline{0.202} & \underline{0.292} &  \textbf{0.201} & \textbf{0.291} &  \underline{0.202} & 0.293 \\ \cmidrule{2-13}
    	& \multirow{4}{*}{\rotatebox[origin=c]{90}{\text{512}}}
    		 & 96 &  \underline{0.130} & \underline{0.224} &  \textbf{0.129} & \underline{0.224} &  \textbf{0.129} & \underline{0.224} &  \textbf{0.129} & \textbf{0.223} &  \textbf{0.129} & \textbf{0.223} \\
    		 & & 192 &  \underline{0.147} & \underline{0.241} &  \textbf{0.146} & \underline{0.241} &  \underline{0.147} & \textbf{0.240} &  \textbf{0.146} & \textbf{0.240} &  \underline{0.147} & \underline{0.241} \\
    		 & & 336 &  \underline{0.163} & \underline{0.258} &  \textbf{0.162} & \textbf{0.257} &  \textbf{0.162} & \underline{0.258} &  \textbf{0.162} & \textbf{0.257} &  \textbf{0.162} & \underline{0.258} \\
    		 & & 720 &  \underline{0.199} & \underline{0.290} &  \textbf{0.198} & \underline{0.290} &  \textbf{0.198} & \underline{0.290} &  \textbf{0.198} & \textbf{0.289} &  \textbf{0.198} & \underline{0.290} \\
    	\midrule
    	\multirow{8}{*}{\rotatebox[origin=c]{90}{\text{Traffic}}}
    	& \multirow{4}{*}{\rotatebox[origin=c]{90}{\text{336}}}
    		 & 96 &  \underline{0.371} & \textbf{0.253} &  \textbf{0.370} & \textbf{0.253} &  0.372 & \underline{0.254} &  0.372 & \underline{0.254} &  0.373 & 0.255 \\
    		 & & 192 &  \textbf{0.388} & \textbf{0.260} &  \textbf{0.388} & \textbf{0.260} &  \underline{0.390} & \underline{0.261} &  \underline{0.390} & \underline{0.261} &  \underline{0.390} & \underline{0.261} \\
    		 & & 336 &  \textbf{0.400} & \underline{0.267} &  \textbf{0.400} & \textbf{0.266} &  0.404 & 0.269 &  \underline{0.402} & 0.268 &  0.403 & 0.268 \\
    		 & & 720 &  \underline{0.435} & \underline{0.287} &  \textbf{0.434} & \textbf{0.286} &  \underline{0.435} & \textbf{0.286} &  \textbf{0.434} & \textbf{0.286} &  \textbf{0.434} & \textbf{0.286} \\ \cmidrule{2-13}
    	& \multirow{4}{*}{\rotatebox[origin=c]{90}{\text{512}}}
    		 & 96 &  \textbf{0.365} & \textbf{0.252} &  \textbf{0.365} & \textbf{0.252} &  \underline{0.367} & \underline{0.253} &  \underline{0.367} & \underline{0.253} &  0.394 & 0.282 \\
    		 & & 192 &  \textbf{0.382} & \textbf{0.258} &  0.391 & 0.269 &  \underline{0.383} & \underline{0.259} &  \textbf{0.382} & 0.260 &  0.407 & 0.286 \\
    		 & & 336 &  0.393 & \textbf{0.265} &  \textbf{0.391} & \textbf{0.265} &  \underline{0.392} & \textbf{0.265} &  0.393 & \textbf{0.265} &  0.394 & \textbf{0.265} \\
    		 & & 720 &  0.435 & 0.289 &  0.432 & 0.288 &  0.432 & \underline{0.287} &  \underline{0.431} & \underline{0.287} &  \textbf{0.430} & \textbf{0.286} \\
    	\bottomrule
    \end{tabular}}
\end{table*}
\endgroup

\FloatBarrier

\FloatBarrier

\subsection{Similarity Power-Law Mask}
\label{sm:experiments_powerformer_simPowerLaw}

Table~\ref{tab:powerformer_simPowerLaw_results} presents the aggregate MSE and MAE results for the similarity power-law mask (\fspl) with varying time decays ($\alpha$), forecast lengths, and input sequence lengths.
The 512 input sequence generally outperforms the shorter 336 input length.
Compared to \fpl, there are more instances in which a 336 input sequence length outperforms.
This is likely due to the much faster decay of \fspl{} compared to \fpl.
As expected, the best performing $\alpha$ varies between datasets.
However, there is much less variation as a function of forecast length and input sequence length when compared to \fpl.

\begin{figure*}[!htb]
    \centering
    \includegraphics[width=0.92\textwidth]{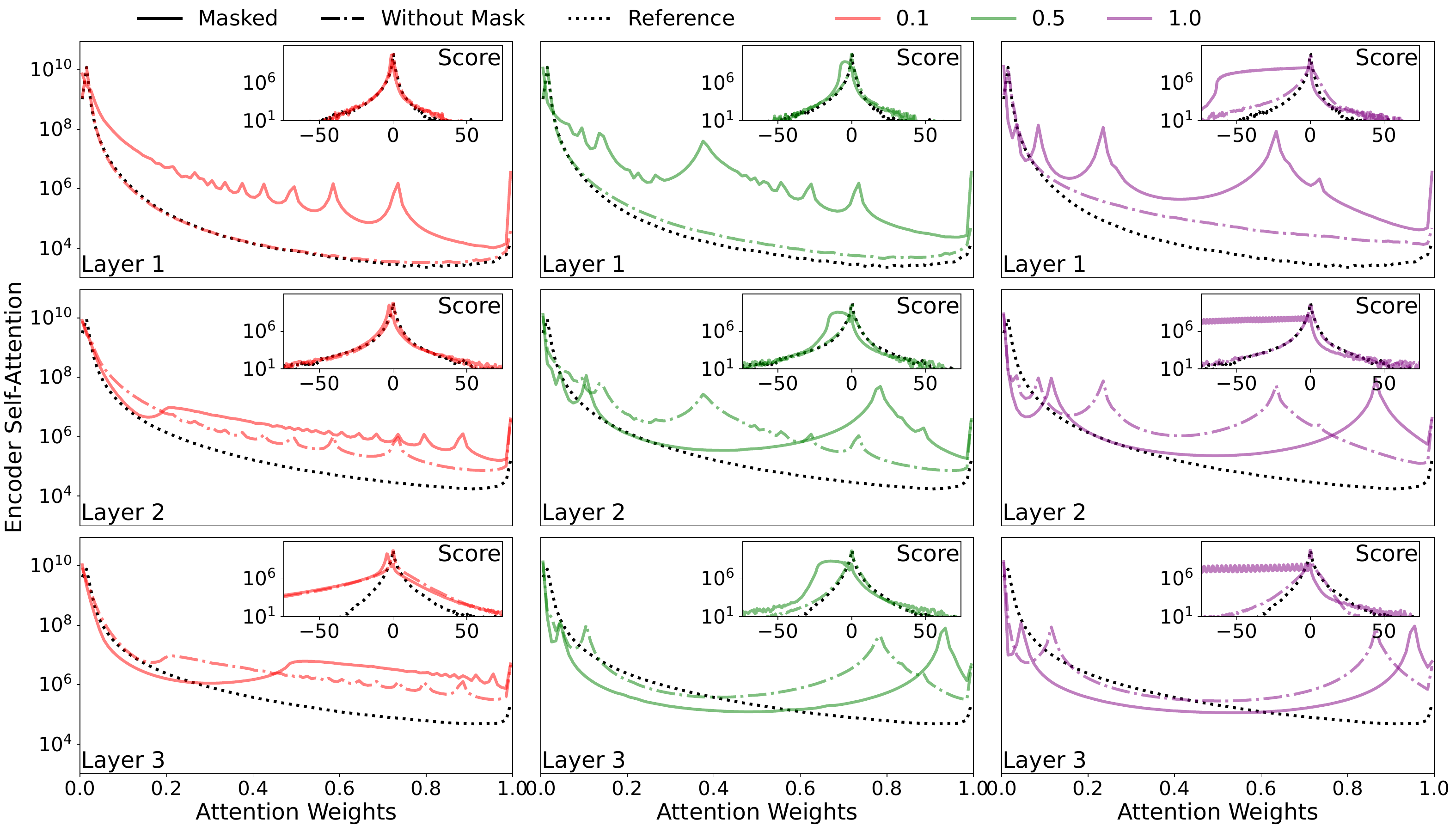}
    \caption{We present \emph{Powerformer's} attention score and weight distributions on the Weather dataset with a forecast and input length of 96 and 512, respectively. The dotted line represents the reference MHA results, the dashed-dotted line represents RBCA with \fspl{} results before applying \maskCL, and the solid lines represent RBCA with \fspl{} results after applying \maskCL.}
    \label{fig:powerformer_weather_96_simPowerLaw_attn_full}
\end{figure*}

\begin{figure*}[!htb]
    \centering
    \includegraphics[width=0.92\textwidth]{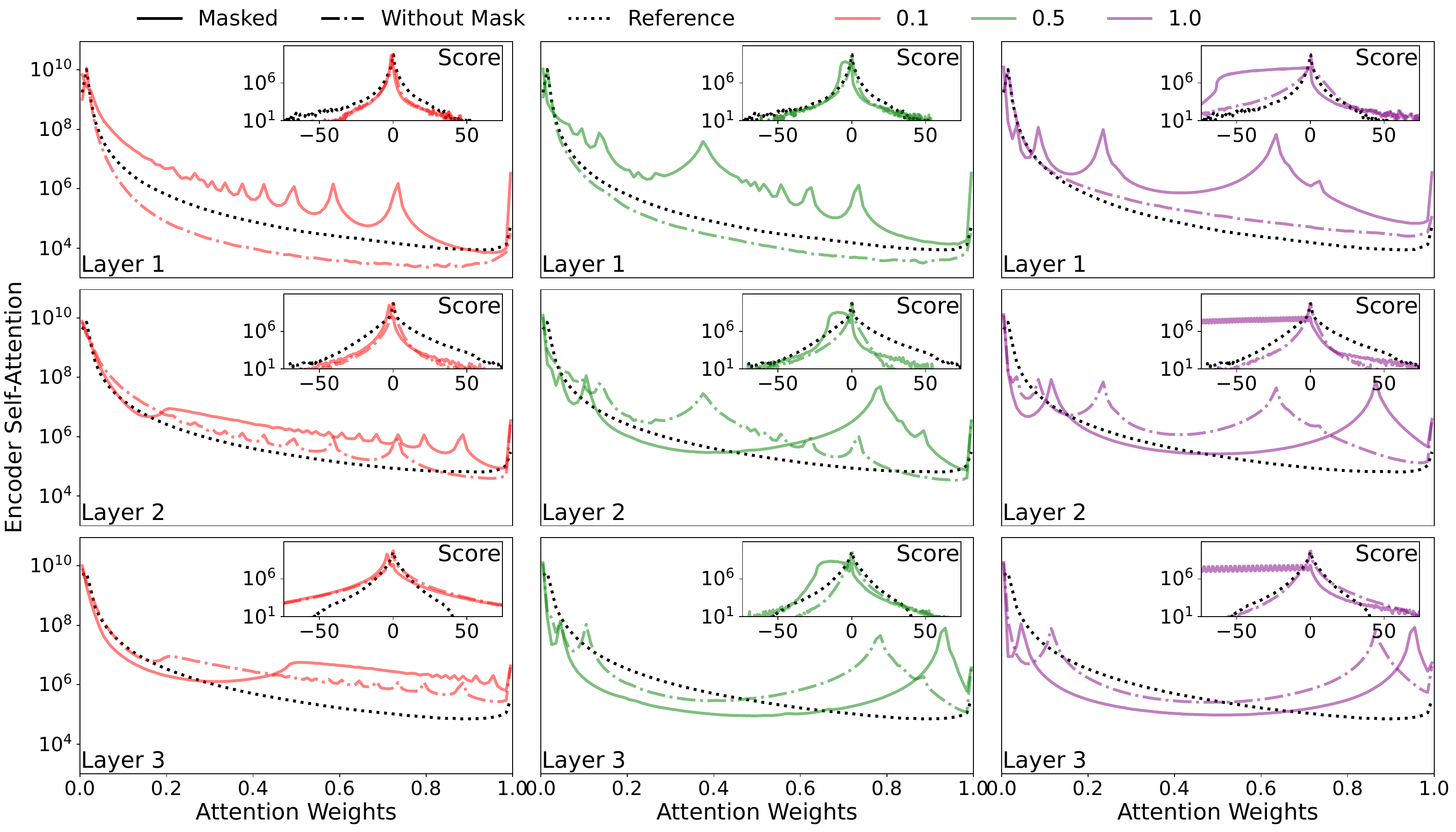}
    \caption{We present \emph{Powerformer's} attention score and weight distributions on the Weather dataset with a forecast and input length of 720 and 512, respectively. The dotted line represents the reference MHA results, the dashed-dotted line represents RBCA with \fspl{} results before applying \maskCL, and the solid lines represent RBCA with \fspl{} results after applying \maskCL.}
    \label{fig:powerformer_weather_720_simPowerLaw_attn_full}
\end{figure*}

\begin{figure*}[!htb]
    \centering
    \includegraphics[width=0.92\textwidth]{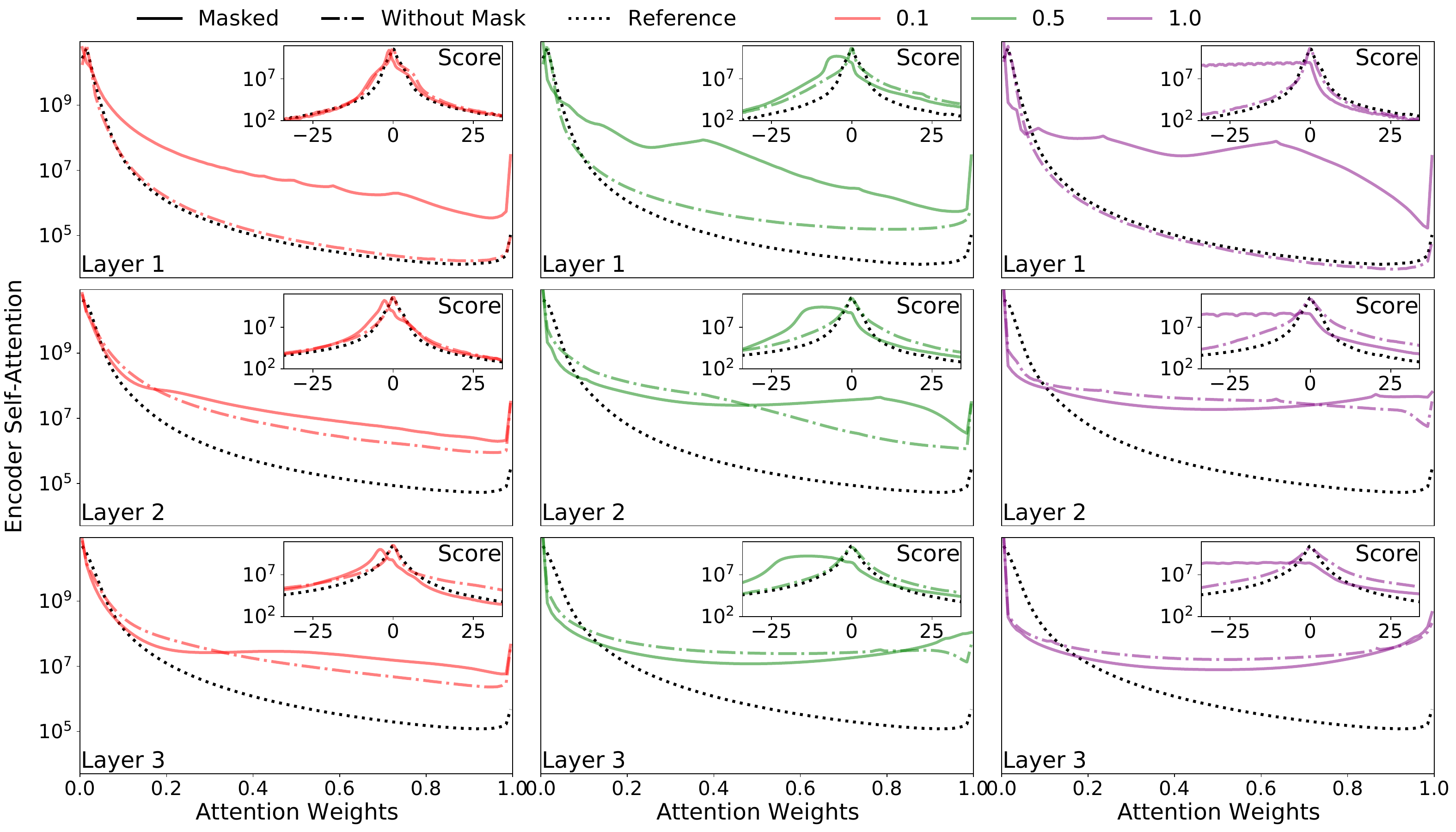}
    \caption{We present \emph{Powerformer's} attention score and weight distributions on the Electricity dataset with a forecast and input length of 96 and 512, respectively. The dotted line represents the reference MHA results, the dashed-dotted line represents RBCA with \fspl{} results before applying \maskCL, and the solid lines represent RBCA with \fspl{} results after applying \maskCL.}
    \label{fig:powerformer_electricity_96_simPowerLaw_attn_full}
\end{figure*}

\begin{figure*}[!htb]
    \centering
    \includegraphics[width=0.92\textwidth]{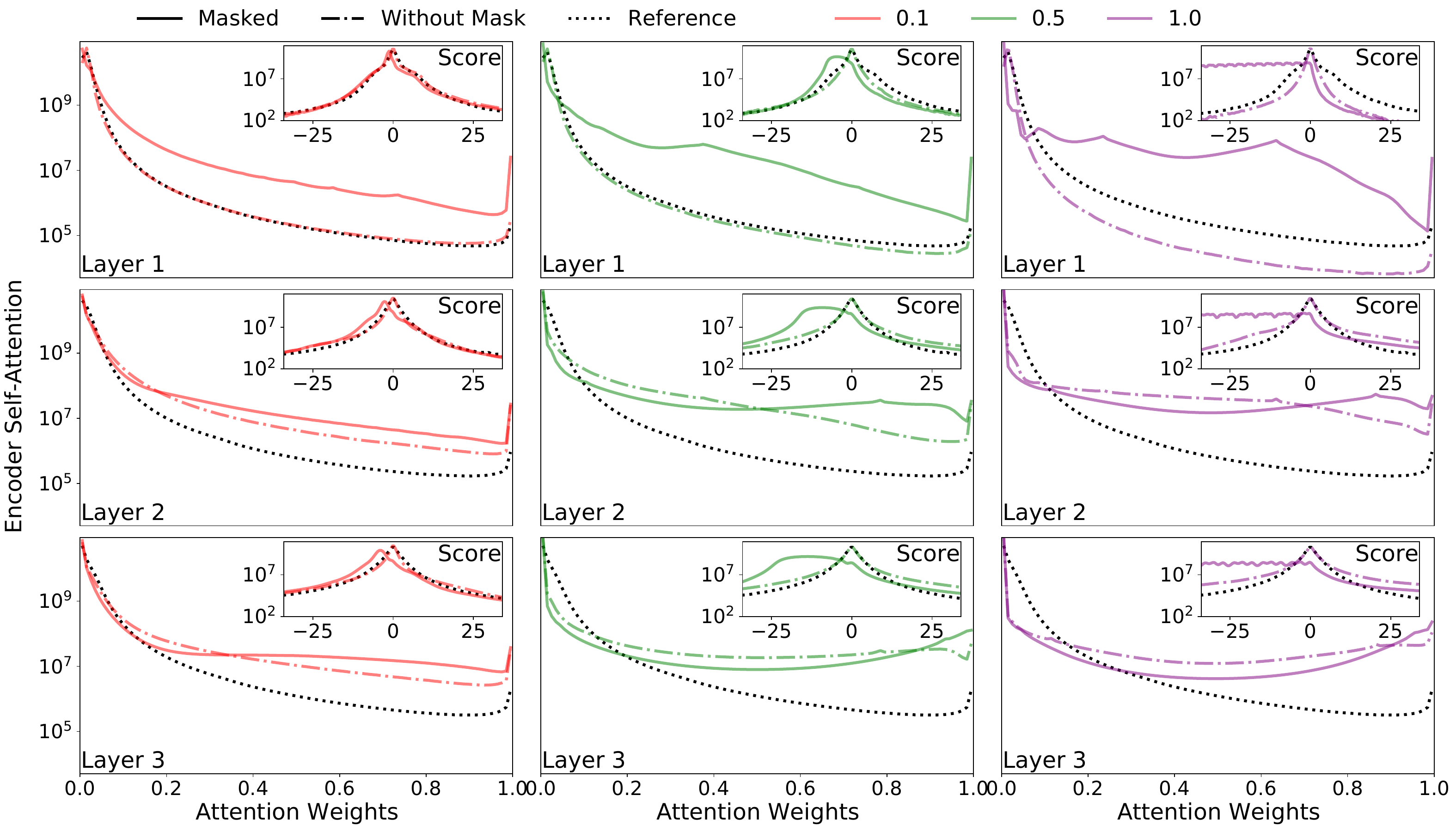}
    \caption{We present \emph{Powerformer's} attention score and weight distributions on the Electricity dataset with a forecast and input length of 720 and 512, respectively. The dotted line represents the reference MHA results, the dashed-dotted line represents RBCA with \fspl{} results before applying \maskCL, and the solid lines represent RBCA with \fspl{} results after applying \maskCL.}
    \label{fig:powerformer_electricity_720_simPowerLaw_attn_full}
\end{figure*}

We compare all datasets' attention score and weight distributions for recency biased attention in Figs~\ref{fig:powerformer_weather_96_simPowerLaw_attn_full}-\ref{fig:powerformer_electricity_720_simPowerLaw_attn_full}.
Due to the faster decay of \fspl{} we observe stronger bimodal distributions for all datasets than for \fpl, with wider \weightCL{} distributions for ETTm1, ETTm2, and Weather.

\begingroup
\setlength{\tabcolsep}{6pt} 
\renewcommand{\arraystretch}{0.62} 

\begin{table*}[!ht]
\centering
    \caption{We compare \emph{Powerformer's} performance with the similarity power-law mask \fspl{} on standard time-series datasets for varying decay lengths. The best results are bolded and the second best are underlined.}
    \label{tab:powerformer_simPowerLaw_results}
    \def\arraystretch{0.57}%
    \resizebox{0.85\textwidth}{!}{
    \begin{tabular}{c|c|c|cc|cc|cc|cc}
	\toprule
	\multicolumn{3}{c|}{Delay}  & \multicolumn{2}{c|}{0.1} & \multicolumn{2}{c|}{0.5} & \multicolumn{2}{c|}{1} & \multicolumn{2}{c}{2} \\
	\midrule
	\multicolumn{3}{c|}{Metric} & MSE & MAE & MSE & MAE & MSE & MAE & MSE & MAE \\
	\midrule
	\multirow{8}{*}{\rotatebox[origin=c]{90}{\text{ETTh1}}}
	& \multirow{4}{*}{\rotatebox[origin=c]{90}{\text{336}}}
		 & 96 &  \textbf{0.377} & \textbf{0.401} &  \underline{0.378} & \underline{0.402} &  \underline{0.378} & \underline{0.402} &  0.379 & 0.403 \\
		 & & 192 &  \textbf{0.413} & \textbf{0.421} &  \underline{0.414} & \textbf{0.421} &  0.415 & \underline{0.422} &  0.415 & \underline{0.422} \\
		 & & 336 &  \textbf{0.424} & \textbf{0.429} &  \underline{0.425} & \textbf{0.429} &  \underline{0.425} & \textbf{0.429} &  \underline{0.425} & \textbf{0.429} \\
		 & & 720 &  \textbf{0.439} & \textbf{0.457} &  \underline{0.440} & \textbf{0.457} &  0.443 & \underline{0.458} &  0.443 & \underline{0.458} \\ \cmidrule{2-11}
	& \multirow{4}{*}{\rotatebox[origin=c]{90}{\text{512}}}
		 & 96 &  \textbf{0.370} & \textbf{0.400} &  \underline{0.371} & \textbf{0.400} &  \underline{0.371} & \textbf{0.400} &  \underline{0.371} & \textbf{0.400} \\
		 & & 192 &  \textbf{0.404} & \textbf{0.420} &  \underline{0.405} & \textbf{0.420} &  \underline{0.405} & \underline{0.421} &  \underline{0.405} & \textbf{0.420} \\
		 & & 336 &  0.418 & 0.433 &  \underline{0.417} & \underline{0.429} &  \textbf{0.415} & \textbf{0.428} &  \textbf{0.415} & \textbf{0.428} \\
		 & & 720 &  \textbf{0.439} & \textbf{0.459} &  \underline{0.441} & \underline{0.460} &  0.442 & 0.461 &  0.442 & 0.461 \\
	\midrule
	\multirow{8}{*}{\rotatebox[origin=c]{90}{\text{ETTh2}}}
	& \multirow{4}{*}{\rotatebox[origin=c]{90}{\text{336}}}
		 & 96 &  \textbf{0.274} & \textbf{0.336} &  \underline{0.275} & \textbf{0.336} &  \underline{0.275} & \textbf{0.336} &  \underline{0.275} & \textbf{0.336} \\
		 & & 192 &  \textbf{0.338} & \textbf{0.379} &  \underline{0.340} & \textbf{0.379} &  \textbf{0.338} & \textbf{0.379} &  \textbf{0.338} & \textbf{0.379} \\
		 & & 336 &  \underline{0.330} & \textbf{0.384} &  0.333 & \underline{0.385} &  \textbf{0.329} & \underline{0.385} &  \textbf{0.329} & \underline{0.385} \\
		 & & 720 &  \underline{0.380} & \textbf{0.422} &  0.381 & \textbf{0.422} &  \textbf{0.379} & \textbf{0.422} &  \textbf{0.379} & \textbf{0.422} \\ \cmidrule{2-11}
	& \multirow{4}{*}{\rotatebox[origin=c]{90}{\text{512}}}
		 & 96 &  \textbf{0.274} & \textbf{0.337} &  0.276 & \underline{0.338} &  \underline{0.275} & \underline{0.338} &  \underline{0.275} & \underline{0.338} \\
		 & & 192 &  \textbf{0.340} & \textbf{0.381} &  0.342 & \underline{0.382} &  \underline{0.341} & \textbf{0.381} &  \underline{0.341} & \textbf{0.381} \\
		 & & 336 &  0.332 & 0.388 &  0.333 & \underline{0.387} &  \underline{0.331} & \underline{0.387} &  \textbf{0.330} & \textbf{0.386} \\
		 & & 720 &  0.386 & \underline{0.428} &  \underline{0.381} & \textbf{0.424} &  \underline{0.381} & \textbf{0.424} &  \textbf{0.380} & \textbf{0.424} \\
	\midrule
	\multirow{8}{*}{\rotatebox[origin=c]{90}{\text{ETTm1}}}
	& \multirow{4}{*}{\rotatebox[origin=c]{90}{\text{336}}}
		 & 96 &  0.295 & 0.345 &  \textbf{0.292} & \textbf{0.343} &  \underline{0.293} & \underline{0.344} &  0.294 & 0.345 \\
		 & & 192 &  0.336 & 0.372 &  \textbf{0.329} & \textbf{0.369} &  \underline{0.333} & \underline{0.371} &  0.337 & 0.373 \\
		 & & 336 &  0.363 & \underline{0.392} &  \textbf{0.360} & \textbf{0.390} &  0.363 & 0.393 &  \underline{0.362} & \underline{0.392} \\
		 & & 720 &  \underline{0.413} & \textbf{0.423} &  0.419 & 0.425 &  0.425 & 0.427 &  \textbf{0.412} & \underline{0.424} \\ \cmidrule{2-11}
	& \multirow{4}{*}{\rotatebox[origin=c]{90}{\text{512}}}
		 & 96 &  \underline{0.291} & \textbf{0.345} &  \textbf{0.289} & \textbf{0.345} &  0.293 & \underline{0.346} &  \underline{0.291} & \textbf{0.345} \\
		 & & 192 &  \textbf{0.333} & \textbf{0.372} &  \underline{0.334} & \underline{0.374} &  \underline{0.334} & \underline{0.374} &  \textbf{0.333} & \underline{0.374} \\
		 & & 336 &  \textbf{0.362} & \textbf{0.391} &  \underline{0.365} & \textbf{0.391} &  \textbf{0.362} & \underline{0.396} &  \textbf{0.362} & \underline{0.396} \\
		 & & 720 &  \textbf{0.419} & 0.427 &  0.426 & \underline{0.418} &  \underline{0.425} & \underline{0.418} &  \underline{0.425} & \textbf{0.417} \\
	\midrule
	\multirow{8}{*}{\rotatebox[origin=c]{90}{\text{ETTm2}}}
	& \multirow{4}{*}{\rotatebox[origin=c]{90}{\text{336}}}
		 & 96 &  \textbf{0.163} & \textbf{0.253} &  \underline{0.164} & \underline{0.254} &  0.165 & 0.256 &  0.165 & \underline{0.254} \\
		 & & 192 &  \textbf{0.222} & \textbf{0.293} &  \textbf{0.222} & \underline{0.294} &  \underline{0.223} & \underline{0.294} &  \underline{0.223} & \underline{0.294} \\
		 & & 336 &  \textbf{0.277} & \textbf{0.329} &  \underline{0.278} & \underline{0.330} &  0.279 & \underline{0.330} &  \underline{0.278} & \textbf{0.329} \\
		 & & 720 &  0.363 & \underline{0.381} &  \textbf{0.358} & 0.383 &  \textbf{0.358} & \textbf{0.380} &  \underline{0.361} & 0.384 \\ \cmidrule{2-11}
	& \multirow{4}{*}{\rotatebox[origin=c]{90}{\text{512}}}
		 & 96 &  \textbf{0.164} & \textbf{0.255} &  \textbf{0.164} & \textbf{0.255} &  \underline{0.165} & \underline{0.256} &  \underline{0.165} & \underline{0.256} \\
		 & & 192 &  \textbf{0.222} & \textbf{0.295} &  \underline{0.223} & \underline{0.296} &  0.224 & \underline{0.296} &  0.224 & \underline{0.296} \\
		 & & 336 &  \textbf{0.274} & \textbf{0.329} &  \underline{0.275} & \textbf{0.329} &  \underline{0.275} & \textbf{0.329} &  \underline{0.275} & \textbf{0.329} \\
		 & & 720 &  \underline{0.354} & \textbf{0.379} &  \underline{0.354} & \underline{0.380} &  \textbf{0.353} & \underline{0.380} &  \textbf{0.353} & \textbf{0.379} \\
	\midrule
	\multirow{8}{*}{\rotatebox[origin=c]{90}{\text{Weather}}}
	& \multirow{4}{*}{\rotatebox[origin=c]{90}{\text{336}}}
		 & 96 &  \textbf{0.151} & \textbf{0.199} &  \underline{0.154} & \underline{0.202} &  \underline{0.154} & \underline{0.202} &  0.156 & 0.204 \\
		 & & 192 &  \textbf{0.195} & \underline{0.243} &  \underline{0.197} & \textbf{0.242} &  0.198 & 0.244 &  0.199 & 0.244 \\
		 & & 336 &  \textbf{0.247} & \underline{0.283} &  \underline{0.248} & \textbf{0.282} &  0.249 & \textbf{0.282} &  \underline{0.248} & \underline{0.283} \\
		 & & 720 &  \underline{0.318} & \underline{0.334} &  \textbf{0.317} & \textbf{0.332} &  0.319 & 0.335 &  0.319 & \underline{0.334} \\ \cmidrule{2-11}
	& \multirow{4}{*}{\rotatebox[origin=c]{90}{\text{512}}}
		 & 96 &  \textbf{0.149} & \textbf{0.199} &  \underline{0.150} & \underline{0.200} &  0.152 & 0.202 &  0.154 & 0.204 \\
		 & & 192 &  \textbf{0.192} & \textbf{0.240} &  \underline{0.194} & \underline{0.242} &  0.196 & 0.244 &  0.198 & 0.244 \\
		 & & 336 &  \textbf{0.243} & \textbf{0.280} &  \underline{0.244} & \textbf{0.280} &  0.246 & 0.283 &  0.246 & \underline{0.281} \\
		 & & 720 &  \textbf{0.310} & \textbf{0.329} &  \underline{0.311} & \underline{0.330} &  0.312 & 0.331 &  0.313 & 0.331 \\
	\midrule
	\multirow{8}{*}{\rotatebox[origin=c]{90}{\text{Electricity}}}
	& \multirow{4}{*}{\rotatebox[origin=c]{90}{\text{336}}}
		 & 96 &  \textbf{0.131} & \textbf{0.224} &  \textbf{0.131} & \underline{0.225} &  \textbf{0.131} & \underline{0.225} &  \textbf{0.131} & \underline{0.225} \\
		 & & 192 &  \underline{0.147} & \textbf{0.240} &  0.148 & \underline{0.242} &  \textbf{0.146} & \textbf{0.240} &  \textbf{0.146} & \textbf{0.240} \\
		 & & 336 &  0.164 & \textbf{0.258} &  \underline{0.163} & \textbf{0.258} &  \textbf{0.162} & \textbf{0.258} &  \underline{0.163} & \textbf{0.258} \\
		 & & 720 &  \textbf{0.201} & \textbf{0.291} &  \underline{0.202} & \underline{0.292} &  \underline{0.202} & 0.293 &  0.204 & 0.295 \\ \cmidrule{2-11}
	& \multirow{4}{*}{\rotatebox[origin=c]{90}{\text{512}}}
		 & 96 &  \textbf{0.129} & \textbf{0.224} &  \underline{0.130} & \textbf{0.224} &  \textbf{0.129} & \textbf{0.224} &  \textbf{0.129} & \textbf{0.224} \\
		 & & 192 &  \underline{0.146} & \textbf{0.240} &  0.147 & \underline{0.241} &  \textbf{0.145} & \textbf{0.240} &  \textbf{0.145} & \textbf{0.240} \\
		 & & 336 &  \textbf{0.162} & \textbf{0.257} &  \underline{0.163} & \underline{0.259} &  \underline{0.163} & \underline{0.259} &  \underline{0.163} & \underline{0.259} \\
		 & & 720 &  \textbf{0.198} & \textbf{0.290} &  \underline{0.199} & \underline{0.291} &  0.205 & 0.295 &  0.207 & 0.296 \\
	\midrule
	\multirow{8}{*}{\rotatebox[origin=c]{90}{\text{Traffic}}}
	& \multirow{4}{*}{\rotatebox[origin=c]{90}{\text{336}}}
		 & 96 &  \textbf{0.372} & \textbf{0.254} &  \underline{0.374} & 0.256 &  0.375 & \underline{0.255} &  0.376 & \underline{0.255} \\
		 & & 192 &  \textbf{0.390} & \textbf{0.261} &  \underline{0.391} & \underline{0.262} &  \underline{0.391} & \underline{0.262} &  0.392 & \underline{0.262} \\
		 & & 336 &  \textbf{0.402} & \textbf{0.268} &  \underline{0.403} & \underline{0.269} &  \underline{0.403} & \textbf{0.268} &  0.404 & \underline{0.269} \\
		 & & 720 &  \textbf{0.435} & \textbf{0.286} &  \underline{0.436} & \underline{0.287} &  \textbf{0.435} & \underline{0.287} &  \underline{0.436} & \underline{0.287} \\ \cmidrule{2-11}
	& \multirow{4}{*}{\rotatebox[origin=c]{90}{\text{512}}}
		 & 96 &  \textbf{0.368} & \textbf{0.254} &  \underline{0.394} & \underline{0.281} &  \underline{0.394} & \underline{0.281} &  0.395 & 0.282 \\
		 & & 192 &  \textbf{0.383} & \textbf{0.260} &  \underline{0.407} & \underline{0.286} &  \underline{0.407} & \underline{0.286} &  \underline{0.407} & 0.287 \\
		 & & 336 &  \textbf{0.393} & \textbf{0.266} &  \underline{0.395} & \textbf{0.266} &  \underline{0.395} & \textbf{0.266} &  \underline{0.395} & \textbf{0.266} \\
		 & & 720 &  \underline{0.432} & \underline{0.288} &  0.433 & \underline{0.288} &  0.434 & \underline{0.288} &  \textbf{0.431} & \textbf{0.287} \\
	\bottomrule
	\end{tabular}}
\end{table*}
\endgroup

\FloatBarrier

\FloatBarrier

\subsection{Butterworth Filter: Order 1}
\label{sm:experiments_powerformer_butter1}

Here we evaluate the Butterworth order 1 mask (\fbwo{}) on the ETT and Weather datasets with a 336 look-back window.
The \fbwo{} decay is not as steep as \fpl{} but has a more gradual decay than \fbwt, as shown in Figs.~\ref{fig:maskBF} and \ref{fig:maskPL}.
This mask offers a middle ground between the step function like \fbwt{} and the faster decaying \fpl.
Table~\ref{tab:powerformer_butter1_results} presents the results for \emph{Powerformer} with RBCA and \fbwo.
We observe that \fbwo{} underperforms  \fpl, \fspl{} and \fbwt.
The poor performance further supports our hypothesis that the functional form of $f(t)$ is important when imposing a locality bias.

\begin{table*}[h]
    \centering
    \caption{We compare \emph{Powerformer's} performance with RBCA and the order 1 Butterworth Filter mask \fbwo{} on standard time-series datasets for varying decay lengths. The best results are bolded and second best are underlined.}
    \label{tab:powerformer_butter1_results}
    \vskip 0.1in
    \begin{tabular}{c|c|cc|cc|cc|cc|cc}
	\toprule
	\multicolumn{2}{c|}{Delay}  & \multicolumn{2}{c|}{2} & \multicolumn{2}{c|}{5} & \multicolumn{2}{c|}{10} & \multicolumn{2}{c|}{15} & \multicolumn{2}{c}{20} \\
	\midrule
	\multicolumn{2}{c|}{Metric} & MSE & MAE & MSE & MAE & MSE & MAE & MSE & MAE & MSE & MAE \\
	\midrule
	\multirow{4}{*}{\rotatebox[origin=c]{90}{\text{ETTh1}}}
         & 96 &  \underline{0.378} & \underline{0.402} &  \textbf{0.377} & \underline{0.402} &  \textbf{0.377} & \underline{0.402} &  \textbf{0.377} & \textbf{0.401} &  \textbf{0.377} & \textbf{0.401} \\
	   & 192 &  \underline{0.415} & \underline{0.422} &  \textbf{0.414} & \textbf{0.421} &  \textbf{0.414} & \textbf{0.421} &  \textbf{0.414} & \textbf{0.421} &  \textbf{0.414} & \textbf{0.421} \\
		 & 336 &  \underline{0.425} & \textbf{0.429} &  \underline{0.425} & \textbf{0.429} &  \underline{0.425} & \textbf{0.429} &  \underline{0.425} & \textbf{0.429} &  \textbf{0.424} & \textbf{0.429} \\
		 & 720 &  0.442 & 0.459 &  \textbf{0.439} & \underline{0.458} &  \textbf{0.439} & \textbf{0.457} &  \underline{0.440} & \underline{0.458} &  \textbf{0.439} & \underline{0.458} \\ 
	\midrule
	\multirow{4}{*}{\rotatebox[origin=c]{90}{\text{ETTh2}}}
		 & 96 &  \underline{0.275} & \textbf{0.336} &  \underline{0.275} & \textbf{0.336} &  \textbf{0.274} & \textbf{0.336} &  \textbf{0.274} & \textbf{0.336} &  \textbf{0.274} & \textbf{0.336} \\
		 & 192 &  \textbf{0.338} & \textbf{0.379} &  \textbf{0.338} & \textbf{0.379} &  \underline{0.339} & \textbf{0.379} &  \underline{0.339} & \textbf{0.379} &  \underline{0.339} & \textbf{0.379} \\
		 & 336 &  \underline{0.331} & \underline{0.386} &  \underline{0.331} & \underline{0.386} &  \underline{0.331} & \textbf{0.385} &  \underline{0.331} & \textbf{0.385} &  \textbf{0.330} & \textbf{0.385} \\
		 & 720 &  \textbf{0.380} & \underline{0.422} &  \textbf{0.380} & \textbf{0.421} &  \underline{0.382} & \underline{0.422} &  \underline{0.382} & 0.424 &  \textbf{0.380} & \underline{0.422} \\ 
	\midrule
	\multirow{4}{*}{\rotatebox[origin=c]{90}{\text{ETTm1}}}
		 & 96 &  \textbf{0.291} & \textbf{0.343} &  \underline{0.293} & \textbf{0.343} &  0.296 & \textbf{0.343} &  0.295 & \underline{0.345} &  0.296 & 0.346 \\
		 & 192 &  \textbf{0.329} & \textbf{0.369} &  \underline{0.333} & \underline{0.371} &  0.337 & 0.372 &  0.338 & 0.373 &  0.336 & 0.373 \\
		 & 336 &  \underline{0.360} & \textbf{0.389} &  \textbf{0.359} & \underline{0.390} &  0.369 & 0.396 &  0.363 & 0.393 &  0.365 & 0.393 \\
		 & 720 &  \textbf{0.414} & 0.426 &  0.425 & 0.429 &  0.418 & \underline{0.425} &  0.417 & \textbf{0.424} &  \underline{0.416} & \underline{0.425} \\ 
	\midrule
	\multirow{4}{*}{\rotatebox[origin=c]{90}{\text{ETTm2}}}
		 & 96 &  \underline{0.164} & \textbf{0.254} &  \textbf{0.163} & \textbf{0.254} &  0.165 & \underline{0.255} &  0.166 & 0.256 &  0.166 & \underline{0.255} \\
		 & 192 &  \underline{0.223} & \underline{0.294} &  \underline{0.223} & \underline{0.294} &  \textbf{0.222} & \textbf{0.293} &  \textbf{0.222} & \textbf{0.293} &  \textbf{0.222} & \textbf{0.293} \\
		 & 336 &  0.279 & 0.330 &  0.278 & 0.330 &  \textbf{0.275} & \textbf{0.327} &  \underline{0.277} & 0.331 &  \underline{0.277} & \underline{0.329} \\
		 & 720 &  0.358 & \underline{0.381} &  \underline{0.353} & \underline{0.381} &  \textbf{0.351} & \underline{0.381} &  0.359 & \textbf{0.380} &  0.359 & \textbf{0.380} \\ 
	\midrule
	\multirow{4}{*}{\rotatebox[origin=c]{90}{\text{Weather}}}
		 & 96 &  0.155 & 0.204 &  0.154 & \underline{0.203} &  \underline{0.153} & \textbf{0.202} &  \underline{0.153} & \underline{0.203} &  \textbf{0.152} & \textbf{0.202} \\
		 & 192 &  0.199 & \textbf{0.244} &  \underline{0.198} & \textbf{0.244} &  \underline{0.198} & \textbf{0.244} &  \underline{0.198} & \underline{0.245} &  \textbf{0.197} & \underline{0.245} \\
		 & 336 &  \underline{0.250} & \underline{0.283} &  \textbf{0.248} & \textbf{0.282} &  \underline{0.250} & 0.284 &  \textbf{0.248} & \underline{0.283} &  \textbf{0.248} & \underline{0.283} \\
		 & 720 &  \underline{0.320} & \underline{0.334} &  \textbf{0.319} & \textbf{0.333} &  0.325 & 0.338 &  0.322 & 0.336 &  \textbf{0.319} & 0.335 \\
        \bottomrule
    \end{tabular}
\end{table*}

\FloatBarrier

\subsection{Butterworth Filter: Order 2}
\label{sm:experiments_powerformer_butter2}

Here we evaluate the Butterworth order 2 mask (\fbwt) on the ETT and Weather datasets with a 336 look-back window.
The \fbwt{} decay is analogous to the step function, but has a smooth decay to 0 at the end of the window, as shown in Figs.~\ref{fig:maskBF}.
We use \fbwt{} as an improved step function to avoid the sharp cutoff.
Table~\ref{tab:powerformer_butter2_results} presents the results for \emph{Powerformer} with RBCA and \fbwt.
We observe that \fbwt{} underperforms \fpl{} and \fspl, but outperforms \fbwo.
The poor performance further supports our hypothesis that the functional form of $f(t)$ is important when imposing a locality bias.

\begin{table*}[!ht]
    \centering
    \caption{We compare \emph{Powerformer's} performance with RBCA and the Butterworth filter mask \fbwt{} on standard time-series datasets for varying decay lengths. The best results are bolded and second best are underlined.}
    \label{tab:powerformer_butter2_results}
    \vskip 0.1in
    \begin{tabular}{c|c|cc|cc|cc|cc|cc}
	\toprule
	\multicolumn{2}{c|}{Delay}  & \multicolumn{2}{c|}{2} & \multicolumn{2}{c|}{5} & \multicolumn{2}{c|}{10} & \multicolumn{2}{c|}{15} & \multicolumn{2}{c}{20} \\
	\midrule
	\multicolumn{2}{c|}{Metric} & MSE & MAE & MSE & MAE & MSE & MAE & MSE & MAE & MSE & MAE \\
	\midrule
	\multirow{4}{*}{\rotatebox[origin=c]{90}{\text{ETTh1}}}
		 & 96 &  \underline{0.378} & \underline{0.402} &  \textbf{0.377} & \underline{0.402} &  \textbf{0.377} & \textbf{0.401} &  \textbf{0.377} & \textbf{0.401} &  \textbf{0.377} & \textbf{0.401} \\
		 & 192 &  \textbf{0.414} & \underline{0.422} &  \textbf{0.414} & \textbf{0.421} &  \textbf{0.414} & \textbf{0.421} &  \textbf{0.414} & \textbf{0.421} &  \textbf{0.414} & \textbf{0.421} \\
		 & 336 &  \underline{0.425} & \textbf{0.429} &  \underline{0.425} & \textbf{0.429} &  \underline{0.425} & \textbf{0.429} &  \textbf{0.424} & \textbf{0.429} &  \textbf{0.424} & \textbf{0.429} \\
		 & 720 &  \underline{0.442} & 0.459 &  \textbf{0.439} & \textbf{0.457} &  \textbf{0.439} & \textbf{0.457} &  \textbf{0.439} & \underline{0.458} &  \textbf{0.439} & \underline{0.458} \\ 
	\midrule
	\multirow{4}{*}{\rotatebox[origin=c]{90}{\text{ETTh2}}}
		 & 96 &  \underline{0.275} & \textbf{0.336} &  \underline{0.275} & \textbf{0.336} &  \textbf{0.274} & \textbf{0.336} &  \textbf{0.274} & \textbf{0.336} &  \textbf{0.274} & \textbf{0.336} \\
		 & 192 &  \textbf{0.338} & \textbf{0.379} &  \textbf{0.338} & \textbf{0.379} &  \textbf{0.338} & \textbf{0.379} &  \underline{0.339} & \textbf{0.379} &  \underline{0.339} & \textbf{0.379} \\
		 & 336 &  \underline{0.331} & 0.386 &  \underline{0.331} & 0.386 &  \underline{0.331} & \underline{0.385} &  \textbf{0.330} & \underline{0.385} &  \textbf{0.330} & \textbf{0.384} \\
		 & 720 &  \textbf{0.380} & \underline{0.422} &  \textbf{0.380} & \textbf{0.421} &  \underline{0.381} & \textbf{0.421} &  \underline{0.381} & 0.423 &  \textbf{0.380} & \underline{0.422} \\ 
	\midrule
	\multirow{4}{*}{\rotatebox[origin=c]{90}{\text{ETTm1}}}
		 & 96 &  \textbf{0.290} & \textbf{0.343} &  \underline{0.293} & \underline{0.344} &  0.295 & 0.345 &  0.295 & 0.346 &  0.297 & \underline{0.344} \\
		 & 192 &  \textbf{0.329} & \textbf{0.370} &  \underline{0.333} & \underline{0.371} &  0.339 & 0.373 &  0.337 & 0.373 &  0.336 & 0.373 \\
		 & 336 &  \textbf{0.361} & \underline{0.391} &  \textbf{0.361} & \underline{0.391} &  \underline{0.362} & 0.393 &  0.365 & 0.395 &  0.363 & \textbf{0.390} \\
		 & 720 &  \textbf{0.413} & 0.425 &  0.422 & 0.428 &  \underline{0.415} & 0.423 &  0.417 & \underline{0.422} &  \textbf{0.413} & \textbf{0.421} \\ 
	\midrule
	\multirow{4}{*}{\rotatebox[origin=c]{90}{\text{ETTm2}}}
		 & 96 &  \underline{0.164} & \underline{0.254} &  \underline{0.164} & \underline{0.254} &  0.165 & \textbf{0.253} &  \textbf{0.163} & \textbf{0.253} &  0.165 & 0.255 \\
		 & 192 &  \underline{0.223} & \underline{0.294} &  \underline{0.223} & \underline{0.294} &  \textbf{0.222} & \textbf{0.293} &  \textbf{0.222} & \textbf{0.293} &  \textbf{0.222} & \textbf{0.293} \\
		 & 336 &  0.279 & \underline{0.330} &  0.278 & \underline{0.330} &  \textbf{0.276} & 0.331 &  \textbf{0.276} & \textbf{0.329} &  \underline{0.277} & \textbf{0.329} \\
		 & 720 &  0.358 & \underline{0.381} &  \underline{0.354} & \underline{0.381} &  \textbf{0.353} & 0.382 &  0.359 & \textbf{0.380} &  0.360 & \textbf{0.380} \\ 
	\midrule
	\multirow{4}{*}{\rotatebox[origin=c]{90}{\text{Weather}}}
		 & 96 &  0.156 & 0.204 &  0.154 & \underline{0.203} &  0.154 & \underline{0.203} &  \underline{0.153} & \textbf{0.202} &  \textbf{0.152} & \underline{0.203} \\
		 & 192 &  0.199 & \textbf{0.244} &  0.198 & \textbf{0.244} &  0.198 & \textbf{0.244} &  \underline{0.197} & \textbf{0.244} &  \textbf{0.196} & \textbf{0.244} \\
		 & 336 &  0.250 & \underline{0.283} &  \textbf{0.248} & \textbf{0.282} &  \underline{0.249} & 0.284 &  \textbf{0.248} & \underline{0.283} &  \textbf{0.248} & 0.284 \\
		 & 720 &  0.320 & \textbf{0.334} &  \textbf{0.318} & \textbf{0.334} &  0.326 & 0.339 &  0.321 & 0.336 &  \underline{0.319} & \underline{0.335} \\ 
	\bottomrule
	\end{tabular}
\end{table*}

\FloatBarrier

\section{ABLATION STUDIES}
\label{sm:ablation}

Here, we perform ablation experiments on \emph{Powerformer}.
To separate the effects of the causal mask and the recency bias, we train three \emph{Powerformer} variants: with MHA, with the causal mask only, and with RBCA.
We train and evaluate these variants on the ETT datasets and the Weather dataset for the 96 and 336 prediction lengths.
Table~\ref{tab:ablation} shows the corresponding results.

\begin{table}
  \centering
    \captionof{table}{Ablation results using vanilla non-causal MHA, causal MHA, and RBCA.}
    \begin{adjustbox}{max width=0.5\textwidth}
        \centering
        \renewcommand{\arraystretch}{1.5} 
        \begin{tabular}{c c | c c | c c | c c}
            \multicolumn{2}{c|}{} & \multicolumn{2}{c|}{MHA} & \multicolumn{2}{c|}{Causal MHA} & \multicolumn{2}{c}{RBCA} \\ \midrule
            \multicolumn{2}{c|}{Metric} & MSE & MAE & MSE & MAE & MSE & MAE \\ \midrule
            \multirow{2}{*}{\rotatebox[origin=c]{90}{\text{ETTh1}}}
                & 96 & \underline{0.370} & \underline{0.400} & \textbf{0.361} & \textbf{0.390} & \textbf{0.361} & \textbf{0.390} \\
                & 336 & 0.422 & 0.440 & \underline{0.413} & \underline{0.429} & \textbf{0.406} & \textbf{0.420} \\ \midrule
            \multirow{2}{*}{\rotatebox[origin=c]{90}{\text{ETTh2}}}
                & 96 & 0.274 & \underline{0.336} & \underline{0.270} & \textbf{0.334} & \textbf{0.269} & \textbf{0.334} \\
                & 336 & \underline{0.331} & \textbf{0.380} & \textbf{0.323} & \textbf{0.380} & \textbf{0.323} & \textbf{0.380} \\ \midrule
            \multirow{2}{*}{\rotatebox[origin=c]{90}{\text{ETTm1}}}
                & 96 & \underline{0.290} & \textbf{0.342} & 0.298 & \underline{0.344} & \textbf{0.285} & \textbf{0.342} \\
                & 336 & \underline{0.366} & 0.392 & 0.370 & \textbf{0.384} & \textbf{0.357} & \underline{0.385} \\ \midrule
            \multirow{2}{*}{\rotatebox[origin=c]{90}{\text{ETTm2}}}
                & 96 & \underline{0.165} & 0.255 & \textbf{0.163} & \underline{0.253} & \textbf{0.163} & \textbf{0.251} \\
                & 336 & 0.278 & 0.329 & \textbf{0.272} & \textbf{0.324} & \underline{0.273} & \underline{0.326} \\ \midrule
            \multirow{2}{*}{\rotatebox[origin=c]{90}{\text{Weather}}}
                & 96 & \underline{0.149} & \underline{0.198} & \textbf{0.147} & \textbf{0.197} & \textbf{0.147} & \textbf{0.197} \\
                & 336 & \underline{0.245} & 0.282 & \textbf{0.243} & \underline{0.281} & \textbf{0.243} & \textbf{0.279} \\ \bottomrule
        \end{tabular}
        \label{tab:ablation}
    \end{adjustbox}
\end{table}

For both models, we examine the effects of different masks (\fpl, \fspl, and \fbwn) and various time decays $\alpha$.
We do not perform the same amount of experiments for each model and mask pairing due to poor performance.
In the remainder of this section, we provide detailed tables of the MSE and MAE scores, as well as plots of the attention scores and weights.
In the last subsection (\ref{sm:learnable_decay_times}), we investigate the effects of making $\alpha$ a learnable parameter.
\FloatBarrier

\subsection{Learnable Mask Time Decays}
\label{sm:learnable_decay_times}

Although we treat the decay time $\alpha$ as a hyperparameter, one may consider learning the optimal decay time by making $\alpha$ learnable.
We observe that $\alpha$ monotonically decreases and in some cases flips sign, changing our mask from a power-law decay to a power-law growth.
To combat this continuous decrease we applied a learning rate scheduler to limit how far $\alpha$ can change from its initialization.
To fairly compare with our hyperparameter results, we initialize $\alpha$ at the same times used in our hyperparameter tuning.
We present the aggregate MSE and MAE results in Table~\ref{tab:powerformer_learnable_time}.

We observe very minor differences between the best hyperparameter results and the best results from learning $\alpha$.
During training, we still observe a monotonically decreasing $\alpha$ that slows down only because of the learning rate scheduler.
This makes sense since the training algorithm gains access to more information as $\alpha$ decreases and facilitates overfitting.
This behavior further supports our claim that \maskCL{} and $\alpha$ act as regularizers.

\begingroup
\renewcommand{\arraystretch}{0.62} 
\begin{table*}[!ht]
    \centering
    \caption{We train \emph{Powerformer} with \fpl{} and learnable time constant and present the results here. The base cases are the best results from \emph{Powerformer} with constant decay weights: Table~\ref{tab:powerformer_powerLaw_results}. The remaining columns designate the initialized time decay.}
    \label{tab:powerformer_learnable_time}
    \vskip 0.1in
    \resizebox{\textwidth}{!}{
    \begin{tabular}{c|c|c|cc|cc|cc|cc|cc|cc}
    \toprule
    \multicolumn{3}{c|}{Delay} & \multicolumn{2}{c|}{Base Case} & \multicolumn{2}{c|}{0.1} & \multicolumn{2}{c|}{0.25} & \multicolumn{2}{c|}{0.5} & \multicolumn{2}{c|}{0.75} & \multicolumn{2}{c}{1} \\
    \midrule
    \multicolumn{3}{c|}{Metric} & MSE & MAE & MSE & MAE & MSE & MAE & MSE & MAE & MSE & MAE & MSE & MAE \\
    \midrule
    \multirow{8}{*}{\rotatebox[origin=c]{90}{\text{ETTh1}}}
    & \multirow{4}{*}{\rotatebox[origin=c]{90}{\text{336}}}
         & 96 &  \textbf{0.376} & \textbf{0.401} & \underline{0.377} & \textbf{0.401} &  \underline{0.377} & \textbf{0.401} &  \underline{0.377} & \textbf{0.401} &  \textbf{0.376} & \textbf{0.401} &  \textbf{0.376} & \textbf{0.401} \\
         & & 192 &  \textbf{0.413} & \textbf{0.421} & \textbf{0.413} & \textbf{0.421} &  \textbf{0.413} & \textbf{0.421} &  \textbf{0.413} & \textbf{0.421} &  \textbf{0.413} & \textbf{0.421} &  \textbf{0.413} & \textbf{0.421} \\
         & & 336 & \textbf{0.424} & \textbf{0.430} & \underline{0.425} & \textbf{0.430} &  \underline{0.425} & \textbf{0.430} &  \underline{0.425} & \textbf{0.430} &  \underline{0.425} & \textbf{0.430} &  \underline{0.425} & \textbf{0.430} \\
         & & 720 & \textbf{0.437} & \textbf{0.455} & \textbf{0.437} & \textbf{0.455} &  \textbf{0.437} & \textbf{0.455} &  \textbf{0.437} & \textbf{0.455} &  \textbf{0.437} & \underline{0.456} &  \underline{0.438} & \underline{0.456} \\ \cmidrule{2-15}
    & \multirow{4}{*}{\rotatebox[origin=c]{90}{\text{512}}}
         & 96 & \textbf{0.369} & \textbf{0.399} & \textbf{0.369} & \textbf{0.399} &  \textbf{0.369} & \textbf{0.399} &  \underline{0.370} & \textbf{0.399} &  \underline{0.370} & \textbf{0.399} &  \underline{0.370} & \textbf{0.399} \\
         & & 192 & \textbf{0.402} & \textbf{0.418} &  0.404 & \underline{0.420} &  0.404 & \underline{0.420} &  \textbf{0.402} & \textbf{0.418} &  \underline{0.403} & \textbf{0.418} &  \underline{0.403} & \textbf{0.418} \\
         & & 336 & \textbf{0.414} & \textbf{0.428} & \textbf{0.414} & \textbf{0.428} &  \textbf{0.414} & \textbf{0.428} &  \underline{0.415} & \underline{0.429} &  \underline{0.415} & 0.430 &  0.416 & 0.431 \\
         & & 720 &  \textbf{0.440} & \textbf{0.460} & \textbf{0.440} & \textbf{0.460} &  \textbf{0.440} & \textbf{0.460} &  \textbf{0.440} & \textbf{0.460} &  \underline{0.443} & \underline{0.462} &  \underline{0.443} & \textbf{0.460} \\
    \midrule
    \multirow{8}{*}{\rotatebox[origin=c]{90}{\text{ETTh2}}}
    & \multirow{4}{*}{\rotatebox[origin=c]{90}{\text{336}}}
         & 96 &  \textbf{0.274} & \textbf{0.335} & \textbf{0.274} & \textbf{0.335} &  \textbf{0.274} & \textbf{0.335} &  \underline{0.275} & \underline{0.336} &  \underline{0.275} & \textbf{0.335} &  \underline{0.275} & \textbf{0.335} \\
         & & 192 & \textbf{0.338} & \textbf{0.379} & 0.341 & \underline{0.381} &  \textbf{0.338} & \textbf{0.379} &  \underline{0.340} & \textbf{0.379} &  \underline{0.340} & \textbf{0.379} &  0.341 & \textbf{0.379} \\
         & & 336 & \textbf{0.325} & \textbf{0.379} & \textbf{0.325} & \textbf{0.379} &  \underline{0.326} & \underline{0.380} &  0.331 & 0.384 &  0.333 & 0.383 &  0.334 & 0.383 \\
         & & 720 &  \textbf{0.376} & \textbf{0.419} & \underline{0.377} & \underline{0.420} &  0.379 & 0.421 &  0.381 & 0.423 &  0.381 & 0.422 &  0.381 & 0.422 \\ \cmidrule{2-15}
    & \multirow{4}{*}{\rotatebox[origin=c]{90}{\text{512}}}
         & 96 &  \textbf{0.274} & \textbf{0.337} & \textbf{0.274} & \textbf{0.337} & \textbf{0.274} & \textbf{0.337} &  \textbf{0.274} & \textbf{0.337} &  \underline{0.275} & \textbf{0.337} &  \underline{0.275} & \underline{0.338} \\
         & & 192 & \textbf{0.340} & \textbf{0.381} & \textbf{0.340} & \textbf{0.381} &  \textbf{0.340} & \textbf{0.381} &  \underline{0.341} & \underline{0.382} &  0.342 & \underline{0.382} &  0.342 & \underline{0.382} \\
         & & 336 &  \textbf{0.330} & \textbf{0.387} & \textbf{0.330} & \textbf{0.387} &  \underline{0.331} & \textbf{0.387} &  0.333 & \underline{0.388} &  0.334 & \underline{0.388} &  0.334 & \underline{0.388} \\
         & & 720 &  \textbf{0.383} & \textbf{0.425} & \textbf{0.383} & \underline{0.426} &  \underline{0.384} & \underline{0.426} &  \textbf{0.383} & \underline{0.426} &  \underline{0.384} & \underline{0.426} &  \textbf{0.383} & \textbf{0.425} \\
    \midrule
    \multirow{8}{*}{\rotatebox[origin=c]{90}{\text{ETTm1}}}
    & \multirow{4}{*}{\rotatebox[origin=c]{90}{\text{336}}}
         & 96 &  \textbf{0.290} & \textbf{0.342} & \underline{0.292} & \underline{0.343} &  \textbf{0.290} & \textbf{0.342} &  \underline{0.292} & \underline{0.343} &  0.293 & 0.344 &  \underline{0.292} & 0.345 \\
         & & 192 &  \textbf{0.330} & \textbf{0.369} & \underline{0.331} & 0.371 &  0.334 & 0.372 &  0.332 & 0.371 &  0.332 & \underline{0.370} &  \textbf{0.330} & \textbf{0.369} \\
         & & 336 &  \textbf{0.359} & \textbf{0.389} & 0.362 & 0.392 &  \underline{0.361} & 0.391 &  0.362 & \underline{0.390} &  \textbf{0.359} & \textbf{0.389} &  \underline{0.361} & \underline{0.390} \\
         & & 720 & \textbf{0.412} & \textbf{0.421} & \underline{0.413} & 0.423 &  \underline{0.413} & \underline{0.422} &  \underline{0.413} & \textbf{0.421} &  \underline{0.413} & \textbf{0.421} &  \textbf{0.412} & 0.423 \\ \cmidrule{2-15}
    & \multirow{4}{*}{\rotatebox[origin=c]{90}{\text{512}}}
         & 96 &  \textbf{0.290} & \textbf{0.345} & \underline{0.291} & \textbf{0.345} &  \textbf{0.290} & \textbf{0.345} &  \textbf{0.290} & \textbf{0.345} &  \underline{0.291} & \underline{0.346} &  0.292 & \underline{0.346} \\
         & & 192 &  \textbf{0.329} & \textbf{0.368} &  0.331 & \underline{0.370} &  \underline{0.330} & \underline{0.370} &  \textbf{0.329} & \textbf{0.368} &  0.332 & \underline{0.370} &  0.332 & 0.372 \\
         & & 336 &  \textbf{0.361} &  \textbf{0.390} & \underline{0.362} & \underline{0.391} &  \underline{0.362} & \textbf{0.390} &  \textbf{0.361} & 0.392 &  \underline{0.362} & \textbf{0.390} &  0.363 & \textbf{0.390} \\
         & & 720 &  \textbf{0.415} & \textbf{0.423} & 0.417 & 0.428 &  \textbf{0.415} & \underline{0.425} &  0.417 & \textbf{0.423} &  0.420 & 0.426 &  \underline{0.416} & 0.431 \\
    \midrule
    \multirow{8}{*}{\rotatebox[origin=c]{90}{\text{ETTm2}}}
    & \multirow{4}{*}{\rotatebox[origin=c]{90}{\text{336}}}
         & 96 &  \textbf{0.162} & \textbf{0.252} & \textbf{0.162} & \textbf{0.252} &  0.165 & 0.255 &  0.166 & 0.256 &  \underline{0.164} & \underline{0.254} &  \underline{0.164} & \underline{0.254} \\
         & & 192 &  \textbf{0.222} & \textbf{0.292} & \textbf{0.222} & \textbf{0.292} &  \textbf{0.222} & \underline{0.293} &  \textbf{0.222} & \underline{0.293} &  \textbf{0.222} & \underline{0.293} &  \textbf{0.222} & \underline{0.293} \\
         & & 336 &  \textbf{0.277} & \textbf{0.329} &  \textbf{0.277} & \textbf{0.329} &  \textbf{0.277} & \textbf{0.329} &  \textbf{0.277} & \textbf{0.329} &  \textbf{0.277} & \textbf{0.329} &  \underline{0.278} & \underline{0.330} \\
         & & 720 &  \textbf{0.359} & \textbf{0.380} &  0.363 & \underline{0.381} &  0.363 & \underline{0.381} &  \underline{0.362} & 0.382 &  \textbf{0.359} & \textbf{0.380} &  \textbf{0.359} & 0.384 \\ \cmidrule{2-15}
    & \multirow{4}{*}{\rotatebox[origin=c]{90}{\text{512}}}
         & 96 & \textbf{0.164} & \textbf{0.255} & \underline{0.165} & \textbf{0.255} &  \underline{0.165} & \textbf{0.255} &  \underline{0.165} & \underline{0.256} &  \underline{0.165} & 0.257 &  \textbf{0.164} & \textbf{0.255} \\
         & & 192 &  \textbf{0.221} & \textbf{0.294} & \underline{0.222} & \underline{0.295} &  0.224 & 0.297 &  0.224 & 0.296 &  0.223 & 0.296 &  \textbf{0.221} & \textbf{0.294} \\
         & & 336 &  \textbf{0.272} & \textbf{0.327} &  0.274 & \underline{0.329} &  0.274 & \underline{0.329} &  0.274 & \underline{0.329} &  \textbf{0.272} & \textbf{0.327} &  \underline{0.273} & \textbf{0.327} \\
         & & 720 & \textbf{0.356} & \textbf{0.381} & \underline{0.358} & 0.383 &  \underline{0.358} & 0.383 &  \underline{0.358} & \underline{0.382} &  \underline{0.358} & 0.383 &  \textbf{0.356} & \textbf{0.381} \\
    \midrule
    \multirow{8}{*}{\rotatebox[origin=c]{90}{\text{Weather}}}
    & \multirow{4}{*}{\rotatebox[origin=c]{90}{\text{336}}}
         & 96 &  \textbf{0.151} & \textbf{0.200} &  \textbf{0.151} & \underline{0.201} &  \textbf{0.151} & \textbf{0.200} &  \textbf{0.151} & \textbf{0.200} &  \underline{0.153} & \underline{0.201} &  0.154 & 0.202 \\
         & & 192 &  \underline{0.196} & \textbf{0.241} &  \underline{0.196} & 0.244 &  \underline{0.196} & 0.243 &  \textbf{0.195} & \underline{0.242} &  \underline{0.196} & \textbf{0.241} &  0.197 & \underline{0.242} \\
         & & 336 &  \underline{0.247} & \textbf{0.281} & 0.248 & 0.283 &  \underline{0.247} & 0.284 &  \underline{0.247} & \underline{0.282} &  \textbf{0.246} & \textbf{0.281} &  0.248 & \underline{0.282} \\
         & & 720 &  \textbf{0.317} & \underline{0.333} & \underline{0.318} & 0.334 &  \underline{0.318} & 0.334 &  \underline{0.318} & 0.334 &  \underline{0.318} & \underline{0.333} &  \textbf{0.317} & \textbf{0.332} \\ \cmidrule{2-15}
    & \multirow{4}{*}{\rotatebox[origin=c]{90}{\text{512}}}
         & 96 &  \textbf{0.147} & \textbf{0.197} & \underline{0.148} & \underline{0.198} &  \underline{0.148} & 0.199 &  \textbf{0.147} & \underline{0.198} &  \underline{0.148} & \underline{0.198} &  0.149 & 0.199 \\
         & & 192 &  \textbf{0.191} & \textbf{0.239} & \underline{0.193} & 0.241 &  0.194 & 0.242 &  \underline{0.193} & \underline{0.240} &  \textbf{0.191} & \textbf{0.239} &  \underline{0.193} & 0.241 \\
         & & 336 &  \textbf{0.243} & \textbf{0.279} & \underline{0.244} & 0.281 &  \textbf{0.243} & 0.281 &  \textbf{0.243} & \underline{0.280} &  \textbf{0.243} & \textbf{0.279} &  \underline{0.244} & \underline{0.280} \\
         & & 720 & \textbf{0.311} & \underline{0.329} & \textbf{0.311} & 0.330 &  \textbf{0.311} & 0.330 &  \underline{0.314} & 0.331 &  \textbf{0.311} & \textbf{0.328} &  \textbf{0.311} & \underline{0.329} \\
    \midrule
    \multirow{8}{*}{\rotatebox[origin=c]{90}{\text{Electricity}}}
    & \multirow{4}{*}{\rotatebox[origin=c]{90}{\text{336}}}
         & 96 &  \textbf{0.130} & \textbf{0.224} & \textbf{0.130} & \textbf{0.224} &  \underline{0.131} & \textbf{0.224} &  \underline{0.131} & \textbf{0.224} &  \underline{0.131} & \textbf{0.224} &  \underline{0.131} & \textbf{0.224} \\
         & & 192 &  \textbf{0.147} & \textbf{0.240} &  \underline{0.148} & \underline{0.241} &  \textbf{0.147} & \textbf{0.240} &  \textbf{0.147} & \textbf{0.240} &  \underline{0.148} & \textbf{0.240} &  \textbf{0.147} & \underline{0.241} \\
         & & 336 &  \textbf{0.164} & \textbf{0.257} &  \textbf{0.164} & \underline{0.258} &  \textbf{0.164} & \underline{0.258} &  \textbf{0.164} & \underline{0.258} &  \textbf{0.164} & 0.259 &  \textbf{0.164} & \underline{0.258} \\
         & & 720 &  \underline{0.201} & \textbf{0.291} &  \underline{0.201} & \underline{0.292} &  \textbf{0.200} & \textbf{0.291} &  \underline{0.201} & \underline{0.292} &  0.202 & \underline{0.292} &  0.202 & 0.293 \\ \cmidrule{2-15}
    & \multirow{4}{*}{\rotatebox[origin=c]{90}{\text{512}}}
         & 96 &  \textbf{0.129} & \textbf{0.223} & \underline{0.130} & \textbf{0.223} &  \textbf{0.129} & \underline{0.224} &  \textbf{0.129} & \underline{0.224} &  \textbf{0.129} & \textbf{0.223} &  \underline{0.130} & \textbf{0.223} \\
         & & 192 &  \textbf{0.146} & \textbf{0.240} &  \underline{0.147} & \underline{0.241} &  \textbf{0.146} & \textbf{0.240} &  \textbf{0.146} & \textbf{0.240} &  \textbf{0.146} & \textbf{0.240} &  \underline{0.147} & \underline{0.241} \\
         & & 336 &  \textbf{0.162} & \textbf{0.257} &  \textbf{0.162} & \underline{0.258} &  \textbf{0.162} & \underline{0.258} &  \textbf{0.162} & \underline{0.258} &  \textbf{0.162} & \textbf{0.257} &  \textbf{0.162} & \underline{0.258} \\
         & & 720 &  \textbf{0.198} & \textbf{0.289} &  \underline{0.199} & \underline{0.290} &  \textbf{0.198} & \underline{0.290} &  \underline{0.199} & \underline{0.290} &  0.200 & 0.291 &  0.200 & 0.291 \\
    \midrule
    \multirow{8}{*}{\rotatebox[origin=c]{90}{\text{Traffic}}}
    & \multirow{4}{*}{\rotatebox[origin=c]{90}{\text{336}}}
         & 96 &  \textbf{0.370} & \textbf{0.253} &  \textbf{0.370} & \textbf{0.253} &  \textbf{0.370} & \textbf{0.253} &  \underline{0.372} & \underline{0.254} &  0.373 & 0.255 &  0.374 & 0.256 \\
         & & 192 &  \textbf{0.388} & \textbf{0.260} &  \textbf{0.388} & \underline{0.261} &  \textbf{0.388} & \textbf{0.260} &  \underline{0.390} & \underline{0.261} &  \underline{0.390} & 0.262 &  \underline{0.390} & 0.262 \\
         & & 336 &  \textbf{0.400} & \textbf{0.266} &  \textbf{0.400} & \underline{0.267} &  \textbf{0.400} & \textbf{0.266} &  \underline{0.404} & 0.269 &  \underline{0.404} & 0.270 &  0.405 & 0.270 \\
         & & 720 &  \underline{0.434} & \textbf{0.286} &  \underline{0.435} & \textbf{0.286} &  \textbf{0.433} & \textbf{0.286} &  \underline{0.435} & \underline{0.287} &  0.437 & 0.289 &  0.436 & \underline{0.287} \\ \cmidrule{2-15}
    & \multirow{4}{*}{\rotatebox[origin=c]{90}{\text{512}}}
         & 96 &  \textbf{0.365} & \underline{0.252} & \textbf{0.365} & \textbf{0.251} &  \underline{0.366} & \underline{0.252} &  \underline{0.366} & 0.253 &  0.367 & 0.254 &  0.394 & 0.282 \\
         & & 192 & \textbf{0.382} & \textbf{0.258} &  0.390 & 0.269 &  \underline{0.383} & 0.261 &  \underline{0.383} & 0.261 &  \textbf{0.382} & \underline{0.260} &  0.407 & 0.287 \\
         & & 336 &  \textbf{0.391} & \underline{0.265} &  \textbf{0.391} & \textbf{0.264} &  \underline{0.392} & \underline{0.265} &  \underline{0.392} & \underline{0.265} &  0.393 & \underline{0.265} &  0.394 & \underline{0.265} \\
         & & 720 &  \textbf{0.430} & \underline{0.286} &  0.433 & 0.288 &  \underline{0.431} & 0.288 &  0.432 & 0.287 &  \textbf{0.430} & \textbf{0.285} &  \textbf{0.430} & \underline{0.286} \\
    \bottomrule
    \end{tabular}}
\end{table*}
\endgroup
\FloatBarrier

\end{document}